\def\year{2018}\relax
\documentclass[letterpaper]{article} 
\usepackage{aaai18}  
\usepackage{times}   
\usepackage{helvet}  
\usepackage{courier} 
\usepackage{url}     
\usepackage{graphicx}
\usepackage{subcaption}
\graphicspath{{images/}}
\usepackage{chngcntr}
\usepackage{etoc}

\usepackage{booktabs}       
\usepackage{microtype}      
\usepackage{nicefrac}       
\usepackage[usenames, dvipsnames]{color}

\usepackage{amsfonts}        
\usepackage{amsmath,amssymb} 
\usepackage{mathtools}

\usepackage{algorithm}
\usepackage[noend]{algpseudocode}

\begin{document}

\title{Beyond Sparsity: Tree Regularization of Deep Models for Interpretability
}

\author{
Mike Wu\textsuperscript{1},
Michael C. Hughes\textsuperscript{2},
Sonali Parbhoo\textsuperscript{3},
\\
 {\bf \Large
 Maurizio Zazzi\textsuperscript{4},
Volker Roth\textsuperscript{3},
\and Finale Doshi-Velez\textsuperscript{2}
 }
 \\
\textsuperscript{1}{Stanford University},
wumike@cs.stanford.edu
\\
\textsuperscript{2}{Harvard University SEAS},
mike@michaelchughes.com, finale@seas.harvard.edu
\\
\textsuperscript{3}{University of Basel},
\{sonali.parbhoo,volker.roth\}@unibas.ch
\\
\textsuperscript{4}{University of Siena},
maurizio.zazzi@unisi.it
}

\copyrighttext{
A version of this work will appear in AAAI 2018 (\url{https://aaai.org/Conferences/AAAI-18/}). This paper includes an extended appendix with supplementary material.
}

\maketitle

\setcounter{secnumdepth}{2}
\pagenumbering{arabic}
\setcounter{page}{1}
\pagestyle{plain}

\begin{abstract}
The lack of interpretability remains a key barrier to the adoption of
deep models in many applications. In this work, we explicitly regularize
deep models so human users might step through the process behind their
predictions in little time.  Specifically, we train deep time-series
models so their class-probability predictions have high accuracy while
being closely modeled by decision trees with few nodes.
Using intuitive toy examples as well as medical tasks for treating sepsis and HIV, we demonstrate that this new tree regularization yields
models that are easier for humans to simulate than
simpler L1 or L2 penalties without sacrificing predictive power.
\end{abstract}

\noindent
\section{Introduction}
Deep models have become the de-facto approach for prediction in a
variety of applications such as image classification
(e.g. \cite{krizhevsky2012imagenet}) and machine translation
(e.g. \cite{bahdanau2014neural,sutskever2014sequence}).  However, many
practitioners are reluctant to adopt deep models because their
predictions are difficult to interpret.  In this work, we seek a
specific form of interpretability known as \emph{human-simulability}.
A human-simulatable model is one in which a human user can ``take in
input data together with the parameters of the model and in reasonable
time step through every calculation required to produce a
prediction'' \cite{lipton2016interpretability}.  For example, small
decision trees with only a few nodes are easy for
humans to simulate and thus understand and trust.
In contrast, even
simple deep models like multi-layer perceptrons with a few dozen
units can have far too many parameters and connections for a human to
easily step through. Deep models for sequences are even more
challenging. Of course, decision trees with too many nodes are also hard to simulate.
Our key research question is: can we create deep models that are well-approximated
by compact, human-simulatable models?

The question of creating accurate yet human-simulatable models is an
important one, because in many domains simulatability is
paramount. For example, despite advances in deep learning for clinical
decision support
(e.g. \cite{miotto2016deep,choi2016doctor,che2015deep}), the clinical
community remains skeptical of machine learning
systems \cite{chen2017machine}.  Simulatability allows clinicians to
audit predictions easily.  They can manually inspect changes to
outputs under slightly-perturbed inputs, check substeps against their
expert knowledge, and identify when predictions are made due to
systemic bias in the data rather than real causes.  Similar needs for
simulatability exist in many decision-critical domains such as
disaster response or recidivism prediction.

To address this need for interpretability, a number of works have been
developed to assist in the interpretation of already-trained models.
\citeauthor{craven1996extracting}
\shortcite{craven1996extracting}
train decision trees that mimic the predictions of a fixed, pretrained neural network, but do not train the network itself to be simpler.
Other post-hoc interpretations typically
typically evaluate the sensitivity of predictions to local perturbations of inputs or the input gradient
\cite{ribeiro2016should,selvaraju2016grad,adler2016auditing,lundberg2016unexpected,erhan2009visualizing}.
In parallel,
research efforts have emphasized that simple lists of (perhaps
locally) important features are not sufficient:
\citeauthor{singh2016programs}
\shortcite{singh2016programs}
provide explanations in the form of programs;
\citeauthor{lakkaraju2016interpretable}
\shortcite{lakkaraju2016interpretable}
learn decision sets and show benefits over other rule-based methods.

These techniques focus on understanding already learned models, rather
than finding models that are more interpretable.  However, it is
well-known that deep models often have multiple optima of
similar predictive accuracy \cite{Goodfellow-et-al-2016}, and thus one
might hope to find more interpretable models with equal
predictive accuracy.  However, the field of \emph{optimizing} deep
models for interpretability remains
nascent.  \citeauthor{ross2017right}
\shortcite{ross2017right} penalize input sensitivity to features marked as less relevant.  \citeauthor{lei2016rationalizing}
\shortcite{lei2016rationalizing}
train deep models that make predictions from text and simultaneously
highlight contiguous subsets of words, called a ``rationale,'' to
justify each prediction.  While both works optimize their deep models
to expose relevant features, lists of features are not sufficient
to \emph{simulate} the prediction.

\paragraph{Contributions.}
In this work, we take steps toward \emph{optimizing} deep models for
human-simulatability via a new model complexity penalty function we call \emph{tree regularization}. Tree regularization favors models whose decision boundaries can be well-approximated by
small decision-trees, thus penalizing models that would require many calculations to simulate predictions.
We first demonstrate how this technique can be used to train
simple multi-layer perceptrons to have tree-like decision boundaries.
We then focus on time-series applications and show that gated recurrent unit (GRU) models trained with strong tree-regularization reach a high-accuracy-at-low-complexity sweet spot that is not possible with any strength of L1 or L2 regularization. Prediction quality can be further boosted by training new hybrid models -- GRU-HMMs -- which explain the residuals of interpretable discrete HMMs via tree-regularized GRUs.  We further show that the approximate decision trees for our tree-regularized deep models are useful for human simulation and interpretability.  We demonstrate our approach on a speech recognition task and two medical treatment prediction tasks for patients with sepsis in the intensive
care unit (ICU) and for patients with human immunodeficiency virus (HIV).
Throughout, we also show that standalone decision trees as a baseline are noticeably less accurate than our tree-regularized deep models.
We have released an open-source Python toolbox to allow others to experiment with tree regularization
\footnote{\scriptsize \url{https://github.com/dtak/tree-regularization-public}}.

\paragraph{Related work.}
While there is little work (as mentioned above) on optimizing models
for interpretability, there are some related threads.  The first
is \emph{model compression}, which trains smaller models that perform
similarly to large, black-box models
(e.g. \cite{buciluǎ2006model,hinton2015distilling,balan2015bayesian,han2015learning}).
Other efforts specifically train very sparse networks via L1 penalties
\cite{zhang2016l1} or even \emph{binary} neural
networks \cite{tang2017train,rastegari2016xnor} with the goal of
faster computation.  Edge and node regularization is commonly used to
improve prediction accuracy
\cite{drucker1992improving,ochiai2017automatic}, and recently \citeauthor{hu2016harnessingLogic}
\shortcite{hu2016harnessingLogic} improve prediction accuracy by training neural networks so that predictions match a small list of known
domain-specific first-order logic rules.  Sometimes, these
regularizations---which all smooth or simplify decision
boundaries---can have the effect of also improving interpretability.
However, there is no guarantee that these regularizations will improve
interpretability; we emphasize that specifically \emph{training} deep
models to have easily-simulatable decision boundaries is (to our
knowledge) novel.

\section{Background and Notation}
We consider supervised learning tasks given datasets of $N$ labeled
examples, where each example (indexed by $n$) has an input feature
vectors $x_n$ and a target output vector $y_n$.  We shall assume the
targets $y_n$ are binary, though it is simple to extend to other
types. When modeling time series, each example sequence $n$ contains
$T_n$ timesteps indexed by $t$ which each have a feature vector
$x_{nt}$ and an output $y_{nt}$. Formally, we write: $x_n =
[x_{n1} \ldots x_{nT_n}]$ and $y_n = [y_{n1} \ldots y_{nT_n}]$. Each
value $y_{nt}$ could be prediction about the next timestep (e.g. the
character at time $t+1$) or some other task-related annotation
(e.g. if the patient became septic at time $t$).

\paragraph{Simple neural networks.}
A multi-layer perceptron (MLP) makes predictions $\hat{y}_n$ of the target
$y_n$ via a function $\hat{y}_n(x_n, W)$, where the vector $W$
represents all parameters of the network.  Given a data set $\{
(x_n,y_n) \}$, our goal is to learn the parameters $W$ to minimize the
objective
\begin{align}
\min_{W} \lambda \Psi(W) + \sum_{n=1}^N \mbox{loss}( y_n , \hat{y}_n( x_n , W ) )
\label{eqn:orig_loss}
\end{align}
For binary targets $y_n$, the logistic loss (binary cross entropy) is
an effective choice.
The regularization term $\Psi(W)$ can represent L1 or L2 penalties
(e.g. \cite{drucker1992improving,Goodfellow-et-al-2016,ochiai2017automatic})
or our new regularization.

\paragraph{Recurrent Neural Networks with Gated Recurrent Units.}
A recurrent neural network (RNN) takes as input an arbitrary length
sequence $x_n = [x_{n1} \ldots x_{nT_n}]$ and produces a ``hidden
state'' sequence $h_n = [h_{n1} \ldots h_{nT_n}]$ of the same length
as the input. Each hidden state vector at timestep $t$ represents a
location in a (possibly low-dimensional) ``state space'' with $K$
dimensions: $h_{nt} \in \mathbb{R}^K$. RNNs perform
sequential \emph{nonlinear} embedding of the form $h_{nt} = f(x_{nt},
h_{nt-1})$ in hope that the state space location $h_{nt}$ is a useful
summary statistic for making predictions of the target $y_{nt}$ at
timestep $t$.
%

Many different variants of the transition function architecture $f$ have been proposed to solve the challenge of capturing long-term dependencies.
In this paper, we use gated recurrent units (GRUs)  \cite{cho2014gru}, which are simpler than other alternatives such as long short-term memory units (LSTMs) \cite{hochreiter1997long}. While GRUs are convenient, any differentiable RNN architecture is compatible with our new tree-regularization approach.

\begin{figure}[!t]
    \centering
    \includegraphics[width=.5\linewidth]{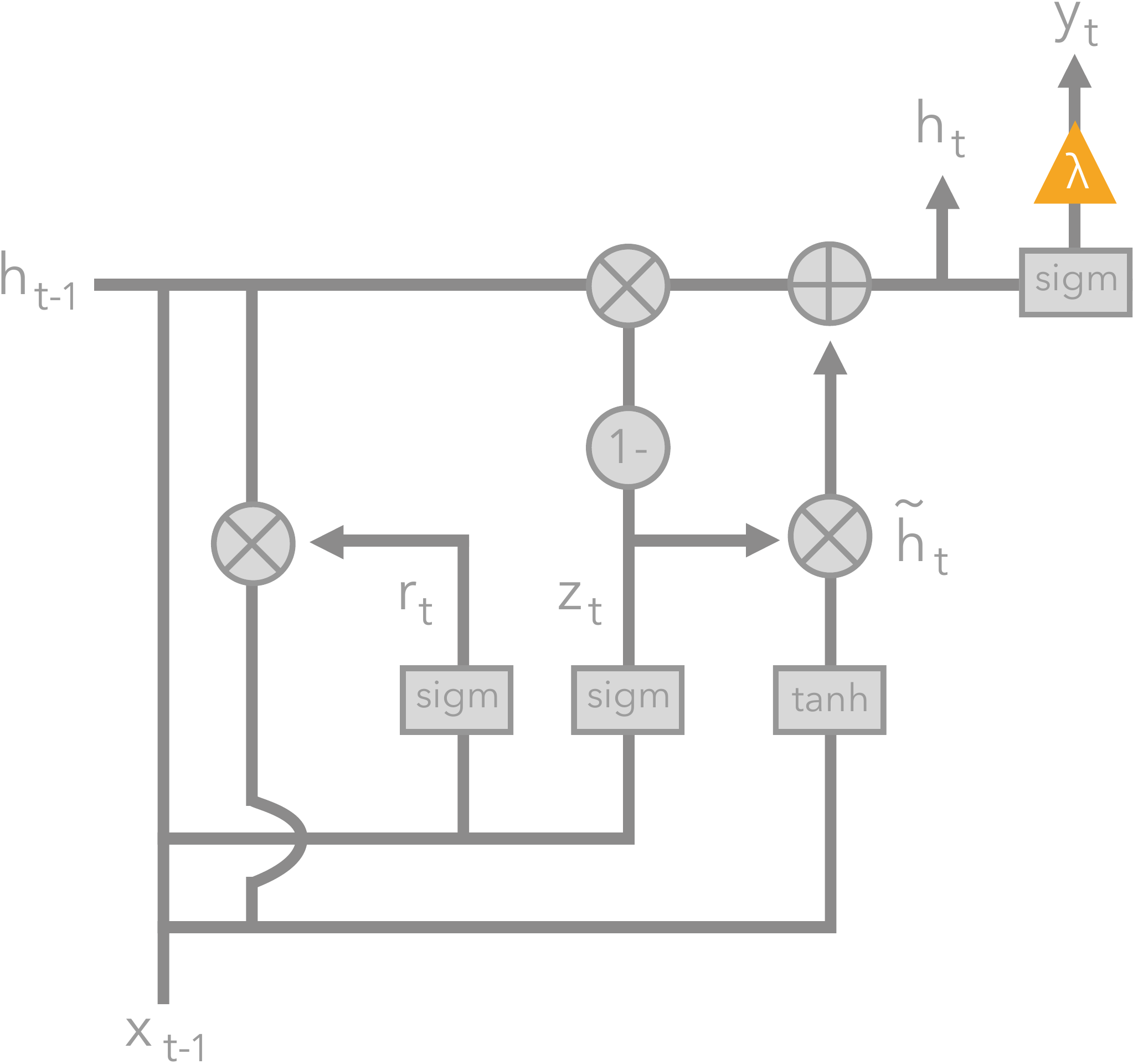}
    \caption{Diagram of gated recurrent unit (GRU) used for each timestep our neural time-series model. The orange triangle indicates the predicted output $\hat{y}_t$ at time $t$.}
    \label{fig:what-we-regularize}
\end{figure}

Below we describe the evolution of a single
GRU sequence, dropping the sequence index $n$ for readability.
The
GRU transition function $f$ produces the state vector $h_{t} =
[h_{t1} \ldots h_{tK}]$
from a previous state $h_{t-1}$ and an input vector $x_t$,
via the following feed-forward architecture:
\begin{align}
\textup{output state}: h_{tk} &= (1 - z_{tk}) h_{t-1,k} + z_{t,k}\tilde{h}_{tk}
\\ \notag
\textup{candidate state}:
\tilde{h}_{tk} &= \textup{tanh}( V_k^{h} x_t + U_k^{h}
    (r_{t} \odot h_{t-1}) )
\\ \notag
\textup{update gate}:
z_{tk} &= \sigma( V_k^{z} x_t + U_k^z h_{t-1} )
\\ \notag
\textup{reset gate}:
r_{tk} &= \sigma(V_k^r x_t + U_k^r h_{t-1})
\end{align}
The internal network nodes include candidate state gates $\tilde{h}$,
update gates $z$ and reset gates $r$ which have the same cardinalty as the state vector $h$.
Reset gates allow the network to forget past state vectors when set near zero via the logistic sigmoid nonlinearity $\sigma(\cdot)$.
Update gates allow the network to either pass along the previous state vector unchanged or use the new candidate state vector instead.
This architecture is diagrammed in Figure~\ref{fig:what-we-regularize}.

The predicted probability of the binary label $y_{t}$ for time
$t$ is a sigmoid transformation of the state at time $t$:
\begin{align}
\hat{y}_{t} = \sigma(w^T h_t)
\end{align}
Here, weight vector $w \in \mathbb{R}^K$ represents the parameters of
this output layer.
We denote the parameters for the entire GRU-RNN model as $W = (w, U, V)$, concatenating all component parameters.
We can train GRU-RNN time-series models (hereafter often just called GRUs) via the following loss minimization
objective:
\begin{align}
\min_{W} \lambda \Psi(W) + \sum_{n=1}^N \sum_{n=1}^{T_n} \mbox{loss}( y_{nt}, \hat{y}_{nt}(x_n, W))
\label{eqn:gru_loss}
\end{align}
where again $\Psi(W)$ defines a regularization cost.

\section{Tree Regularization for Deep Models}

We now propose a novel \emph{tree regularization} function $\Omega(W)$ for the
parameters of a differentiable model which attempts to penalize
models whose predictions are not easily \emph{simulatable}. Of course,
it is difficult to measure ``simulatability'' directly for an
arbitrary network, so we take inspiration from decision trees. Our
chosen method has two stages: first, find a single binary decision tree which
accurately reproduces the network's thresholded binary predictions
$\hat{y}_{n}$ given input $x_n$. Second, measure the complexity of
this decision tree as the output of $\Omega(W)$.  We
measure complexity as the \emph{average decision path length}---the
average number of decision nodes that must be touched to make a
prediction for an input example $x_n$.  We compute the \emph{average}
with respect to some designated reference dataset of example inputs $D
= \{x_n\}$ from the training set.  While many ways to measure
complexity exist, we find average path length is most relevant to our
notion of \emph{simulatability}. Remember that for us, human simulation requires stepping through every calculation required to make a prediction. Average path length exactly counts the number of true-or-false boolean calculations needed to make an average prediction, assuming the model is a decision tree. Total number of nodes could be used as a metric, but might penalize more accurate trees that have short paths for most examples but need more involved logic for few outliers.

Our true-average-path-length cost function $\Omega(W)$ is detailed in
Alg.~\ref{alg:true_tree_regularization}.  It requires two
subroutines, \textsc{TrainTree}
and \textsc{PathLength}. \textsc{TrainTree} trains a binary decision tree to
accurately reproduce the provided labeled examples
$\{x_n, \hat{y}_n \}$. We use the \texttt{DecisionTree} module
distributed in Python's scikit-learn \cite{scikit-learn} with post-pruning to simplify the tree. These trees can give probabilistic predictions at each leaf.
(Complete decision-tree training details are in the supplement.)
Next, \textsc{PathLength} counts how many nodes are needed to make a
specific input to an output node in the provided decision tree.  In
our evaluations, we will apply our average-decision-tree-path-length
regularization, or simply ``tree regularization,'' to several neural models.

\begin{algorithm}[!t]
\caption{Average-Path-Length Cost Function}
\begin{algorithmic}[1]
\Require{
\Statex $\hat{y}(\cdot, W)$ : binary prediction function, with parameters $W$
\Statex $D = \{ x_n \}_{n=1}^N$ : reference dataset with $N$ examples
}
\Function{$\Omega$}{$W$}
\State $\mbox{tree} \gets \textsc{TrainTree}( \{ x_n, \hat{y}(x_n, W) \})$
\State \Return $\frac{1}{N} \sum_{n} \textsc{PathLength}(\mbox{tree}, x_n)$
\EndFunction
\end{algorithmic}
\label{alg:true_tree_regularization}
\end{algorithm}

\noindent Alg.~\ref{alg:true_tree_regularization} defines our average-path-length cost
function $\Omega(W)$, which can be plugged into the abstract regularization term $\Psi(W)$ in the objectives in equations~\ref{eqn:orig_loss} and~\ref{eqn:gru_loss}.

\paragraph{Making the Decision-Tree Loss Differentiable}
Training decision trees is not differentiable, and thus $\Omega(W)$ as defined in Alg.~\ref{alg:true_tree_regularization} is not differentiable with
respect to the network parameters $W$ (unlike standard regularizers
such as the L1 or L2 norm).  While one could resort to derivative-free
optimization techniques \cite{audet2016blackbox}, gradient descent has
been an extremely fast and robust way of training
networks \cite{Goodfellow-et-al-2016}.

A key technical contribution of our work is introducing and training
a \emph{surrogate} regularization function
$\hat{\Omega}(W): \mbox{supp}(W) \rightarrow \mathbb{R}_+$ to map each
candidate neural model parameter vector $W$ to an \emph{estimate} of
the average-path-length.  Our approximate function $\hat{\Omega}$ is
implemented as a standalone multi-layer perceptron network and is
thus \emph{differentiable}.  Let vector $\xi$ of size $k$ denote the
parameters of this chosen MLP approximator.  We can train
$\hat{\Omega}$ to be a good estimator by minimizing a squared error
loss function:
\begin{align}
\min_{\xi} \textstyle \sum_{j=1}^J (\Omega( W_j ) - \hat{\Omega}( W_j, \xi ) )^2 + \epsilon || \xi ||_{2}^2
\label{eqn:surrogate_loss}
\end{align}
where $W_j$ are the \emph{entire} set of parameters for our model, $\epsilon > 0$ is a regularization strength, and
we assume we have a dataset of $J$ known parameter vectors and their
associated true path-lengths: $\{W_j, \Omega(W_j) \}_{j=1}^J$. This
dataset can be assembled using the candidate $W$ vectors obtained
while training our target neural model $\hat{y}(\cdot, W)$, as well as
by evaluating $\Omega(W)$ for randomly generated $W$.  Importantly,
one can train the surrogate function $\hat{\Omega}$ in parallel with
our network.  In the supplement, we show evidence that our
surrogate predictor $\hat{\Omega}(\cdot)$ tracks the true average path
length as we train the target predictor $\hat{y}(\cdot, W)$.

\paragraph{Training the Surrogate Loss}
Even moderately-sized GRUs can have parameter vectors $W$ with
thousands of dimensions.  Our labeled dataset for surrogate training
-- $\{ W_j, \Omega(W_j) \}_{j=1}^J$---will only have one $W_j$ example
from each target network training iteration.  Thus, in early
iterations, we will have only few examples from which to learn a good
surrogate function $\hat{\Omega}(W)$.  We resolve this challenge via
\emph{augmenting} our training set with additional examples: We randomly
sample weight vectors $W$ and calculate the true average path length
$\Omega(W)$, and we also perform several random restarts on the
unregularized GRU and use those weights in our training set.

A second challenge occurs later in training: as the model parameters
$W$ shift away from their initial values, those early parameters may
not be as relevant in characterizing the current decision function of
the GRU. To address this, for each epoch, we use examples only from
the past $E$ epochs (in addition to augmentation), where in practice,
$E$ is empirically chosen.  Using examples from a fixed window of
epochs also speeds up training. The supplement
shows a comparison of the importance of these heuristics for efficient and accurate training---empirically, data augmentation for
stabilizing surrogate training allows us to scale to GRUs with 100s of
nodes. GRUs of this size are sufficient for many real problems, such
as those we encounter in healthcare domains.

Typically, we use $J=50$ labeled pairs for surrogate
training for toy datasets and $J=100$ for real world datasets.
Optimization of our surrogate objective is done via
gradient descent. We use Autograd to compute gradients of the loss in
Eq.~\eqref{eqn:surrogate_loss} with respect to $\xi$, then use Adam
to compute descent directions with step sizes set to 0.01 for toy datasets and 0.001 for real world datasets.

\section{Tree-Regularized MLPs: A Demonstration}
\begin{figure}[!h]
\centering
    \begin{subfigure}[b]{\linewidth}
        \centering
        \includegraphics[width=.35\linewidth]{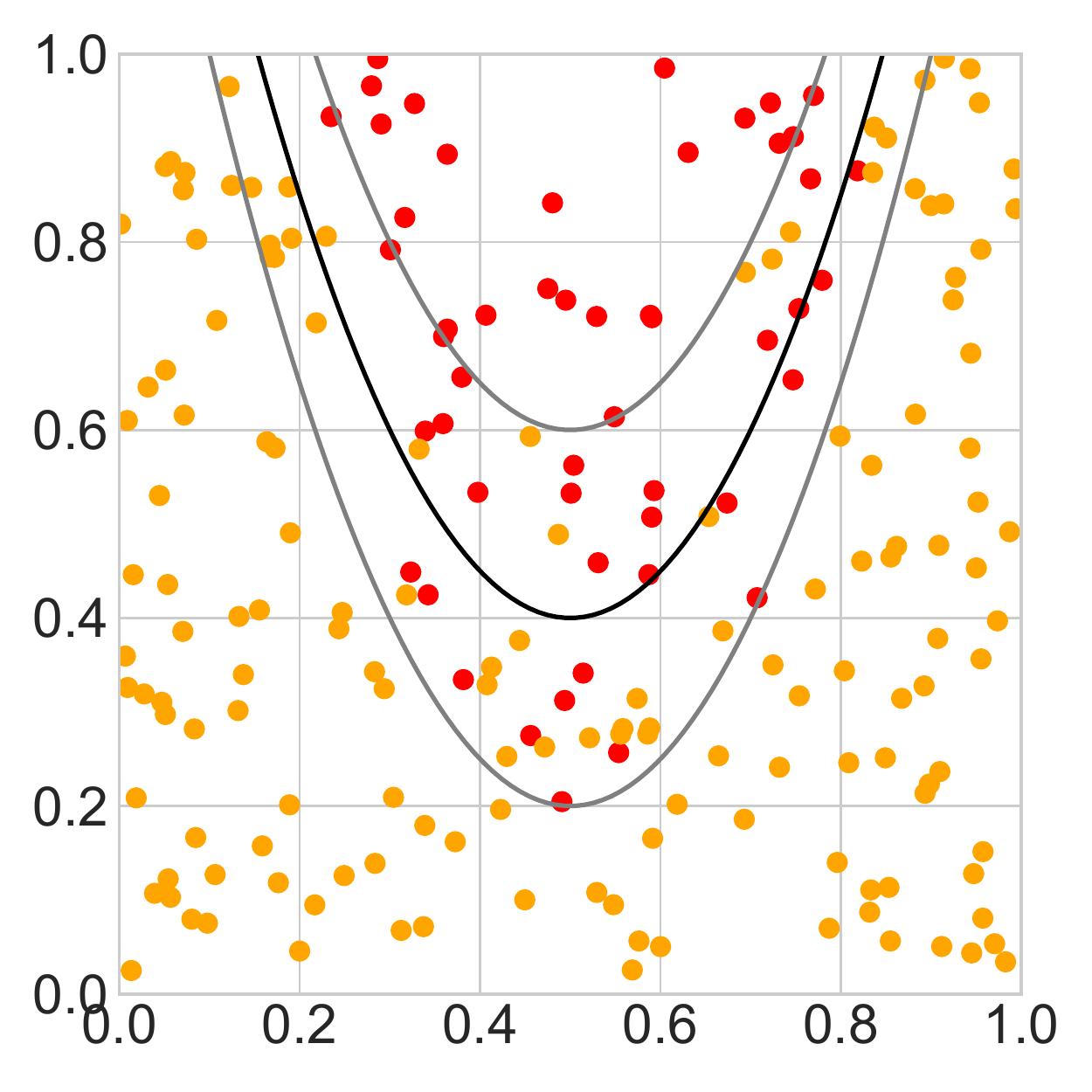}
        \caption{Training Data and Binary Class Labels for 2D Parabola}
    \end{subfigure}
    \begin{subfigure}[b]{\linewidth}
        \includegraphics[width=\linewidth]{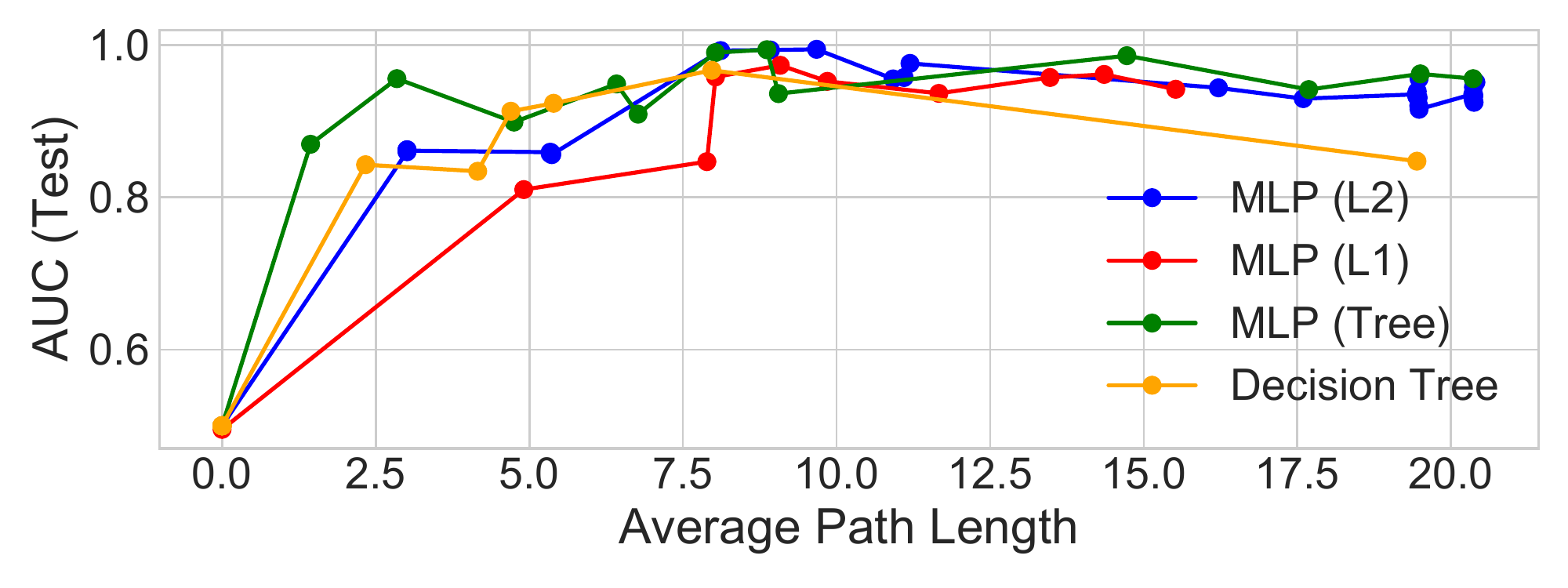}
        \caption{Prediction quality and complexity as reg. strength $\lambda$ varies}
    \end{subfigure}
    \begin{subfigure}[b]{\linewidth}
        \includegraphics[width=\linewidth]{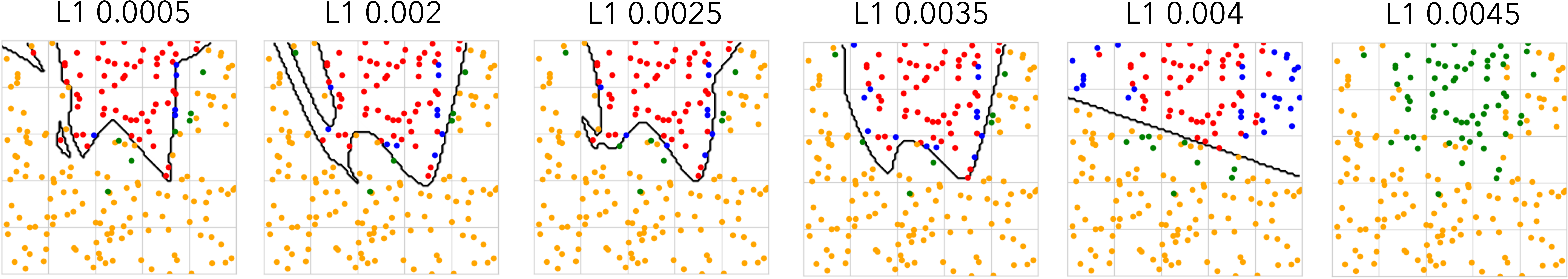}
        \caption{Decision Boundaries with L1 regularization\\}
        \label{fig:decision_function:l1}
    \end{subfigure}
    \begin{subfigure}[b]{\linewidth}
        \includegraphics[width=\linewidth]{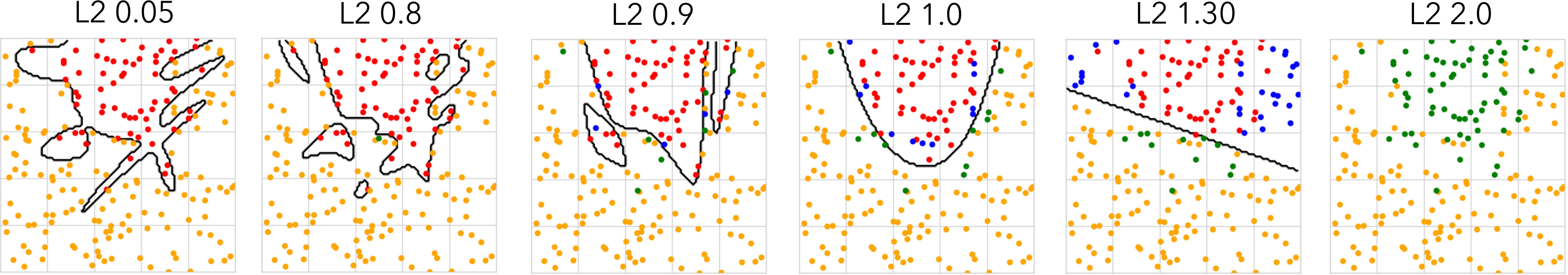}
        \caption{Decision Boundaries with L2 regularization\\}
        \label{fig:decision_function:l2}
    \end{subfigure}
    \begin{subfigure}[b]{\linewidth}
        \includegraphics[width=\linewidth]{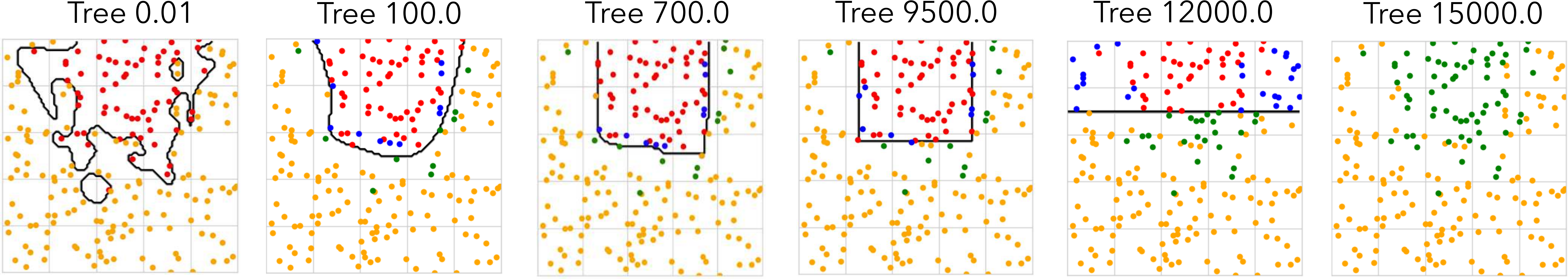}
        \caption{Decision Boundaries Tree regularization}
        \label{fig:decision_function:tree}
    \end{subfigure}
    \caption{
    \emph{2D Parabola task}: \emph{(a)} Each training data point in 2D space, overlaid with true parabolic class boundary.
    \emph{(b)}:
    Each method's prediction quality (AUC) and complexity (path length) metrics, across range of regularization strength $\lambda$.
    In the small path length regime between 0 and 5, tree regularization produces models with higher AUC than L1 or L2.
    \emph{(c-e)}: Decision boundaries (black lines) have qualitatively different shapes
    for different regularization schemes, as regularization strength $\lambda$ increases. We color predictions as true positive (red), true negative (yellow), false negative (green), and false positive (blue).}
    \label{fig:parabola}
\end{figure}

While time-series models are the main focus of this work, we first
demonstrate tree regularization on a simple binary classification task
to build intuition.  We call this task the 2D Parabola problem,
because as Fig.~\ref{fig:parabola}(a) shows, the training data
consists of 2D input points whose two-class decision boundary is
roughly shaped like a parabola.  The true decision function is defined
by $y = 5*(x-0.5)^{2} + 0.4$. We sampled 500 input points $x_n$
uniformly within the unit square $[0,1] \times [0,1]$ and labeled
those above the decision function as positive.  To make it easy for
models to overfit, we flipped 10\% of the points in a region near the
boundary.  A random 30\% were held out for testing.

For the classifier $\hat{y}$, we train a 3-layer MLP with 100 first layer nodes, 100 second layer nodes, and 10 third layer nodes.
This MLP is intentionally overly expressive to encourage overfitting and expose
the impact of different forms of regularization:
our proposed tree regularization $\Psi(W)
= \hat{\Omega}(W)$ and two baselines: an L2 penalty on the weights
$\Psi(W) = ||W||_2$, and an L1 penalty on the weights $\Psi(W) =
||W||_1$.  For each regularization function, we train models at many
different regularization strengths $\lambda$ chosen to explore the
full range of decision boundary complexities possible under each
technique.

For our tree regularization, we model our surrogate
$\hat{\Omega}(W)$ with a 1-hidden layer MLP with 25 units.
We find this simple architecture works well, but certainly more complex MLPs could could be used on more complex problems.
The
objective in equation~\ref{eqn:orig_loss} was optimized via Adam
gradient descent \cite{kingma2014adam} using a batch size of 100 and a
learning rate of 1e-3 for 250 epochs, and hyperparameters were set via
cross validation using grid search (see supplement for full
experimental details).

Fig.~\ref{fig:parabola} (b) shows the each trained model as a single
point in a 2D fitness space: the x-axis measures model complexity via
our average-path-length metric, and the y-axis measures AUC prediction
performance.  These results show that simple L1 or L2 regularization
does \emph{not} produce models with both small node count and good
predictions at \emph{any} value of the regularization strength
$\lambda$. As expected, large $\lambda$ values for L1 and L2 only
produce far-too-simple linear decision boundaries with poor
accuracies.  In contrast, our proposed tree regularization directly
optimizes the MLP to have simple tree-like boundaries at high
$\lambda$ values which can still yield good predictions.

The lower panes of Fig.~\ref{fig:parabola} shows these boundaries.
Our tree regularization is uniquely able to create axis-aligned
functions, because decision trees prefer functions that are
axis-aligned splits. These axis-aligned functions require very few
nodes but are more effective than L1 and L2 counterparts. The L1
boundary is more sharp, whereas the L2 is more round.


\section{Tree-Regularized Time-Series Models}
We now evaluate our tree-regularization approach on time-series
models. We focus on GRU-RNN models, with some later experiments on new hybrid GRU-HMM models.  As with the MLP,
each regularization technique (tree, L2, L1) can be applied to the
output node of the GRU across a range of strength parameters $\lambda$.
Importantly, Algorithm ~\ref{alg:true_tree_regularization} can compute the average-decision-tree-path-length for any fixed deep model given its parameters, and can hence be used to measure decision boundary complexity under any regularization, including L1 or L2.
This means that when training any model, we can track both the predictive performance (as measured by area-under-the-ROC-curve (AUC); higher values mean better predictions), as well as the complexity of the decision tree required to explain each model (as measured by our average path length metric; lower values mean more interpretable models).
We also show results for a baseline standalone decision tree classifier without any associated deep model, sweeping a range of parameters controlling leaf size to explore how this baseline trades off path length and prediction quality.
Further details of our experimental protocol are in the supplement, as well as more extensive results with additional baselines.

\subsection{Tasks}
\paragraph{Synthetic Task: Signal-and-noise HMM}
We generated a toy dataset of $N=100$ sequences, each with $T=50$
timesteps. Each timestep has a data vector $x_{nt}$ of 14 binary features and
a single binary output label $y_{nt}$.  The data comes from two separate HMM
processes. First, a ``signal'' HMM generates the first 7 data
dimensions from 5 well-separated states.  Second, an independent
``noise'' HMM generates the remaining 7 data dimensions from a
different set of 5 states.  Each timestep's output
label $y_{nt}$ is produced by a rule involving \emph{both} the signal
data and the signal hidden state: the target is 1 at timestep $t$ only
if both the first signal state is active and the first observation is
turned on.  We deliberately designed the generation process so that
neither logistic regression with $x$ as features nor an RNN model that makes predictions from hidden states alone can perfectly separate this data.

\begin{figure*}
    \centering
    \begin{subfigure}[b]{0.24\linewidth}
        \includegraphics[width=\linewidth]{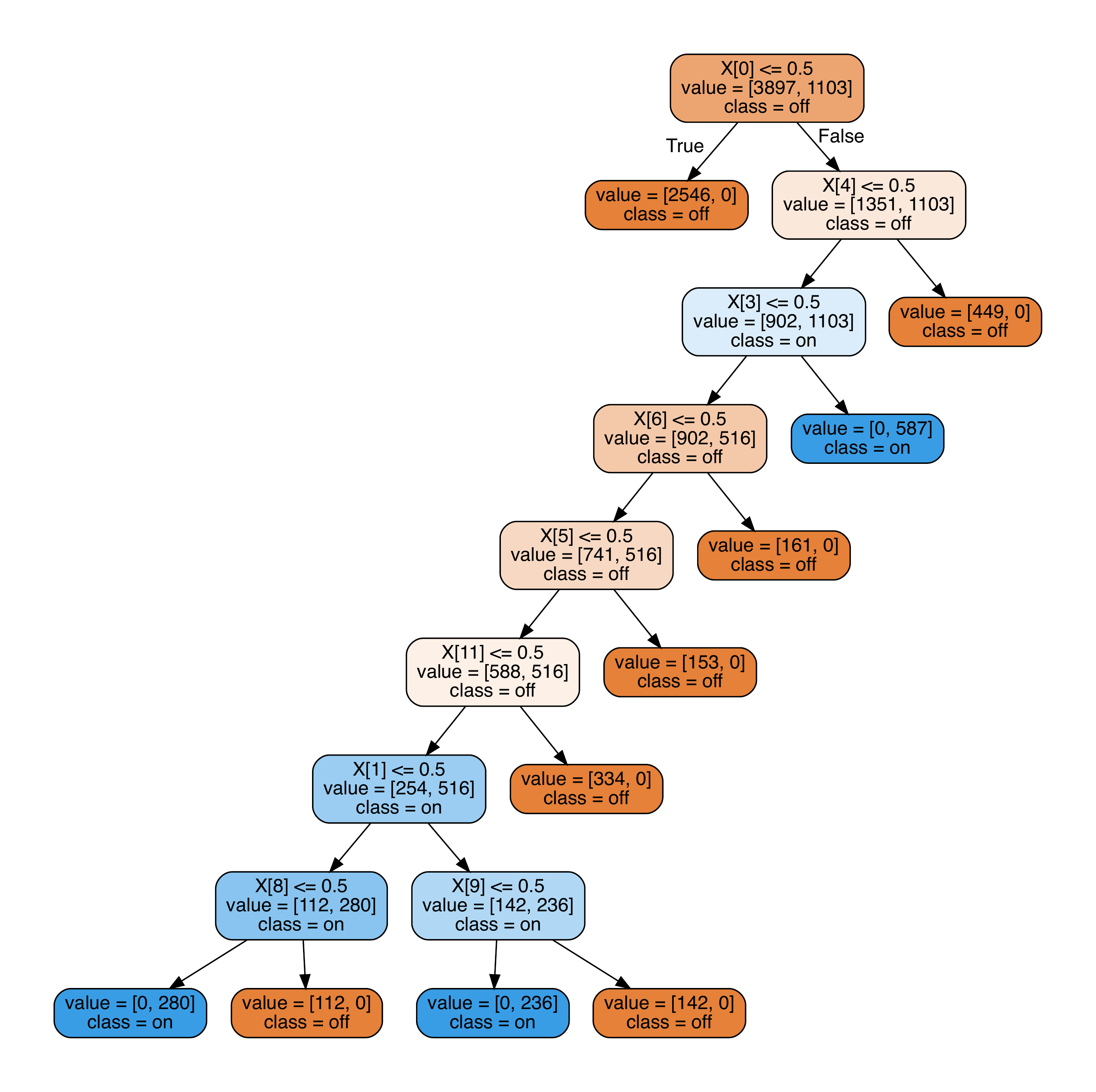}
        \label{fig:hmm2:tree:1}
        \caption{GRU $\lambda = 1$}
    \end{subfigure}
    \begin{subfigure}[b]{0.24\linewidth}
        \includegraphics[width=\linewidth]{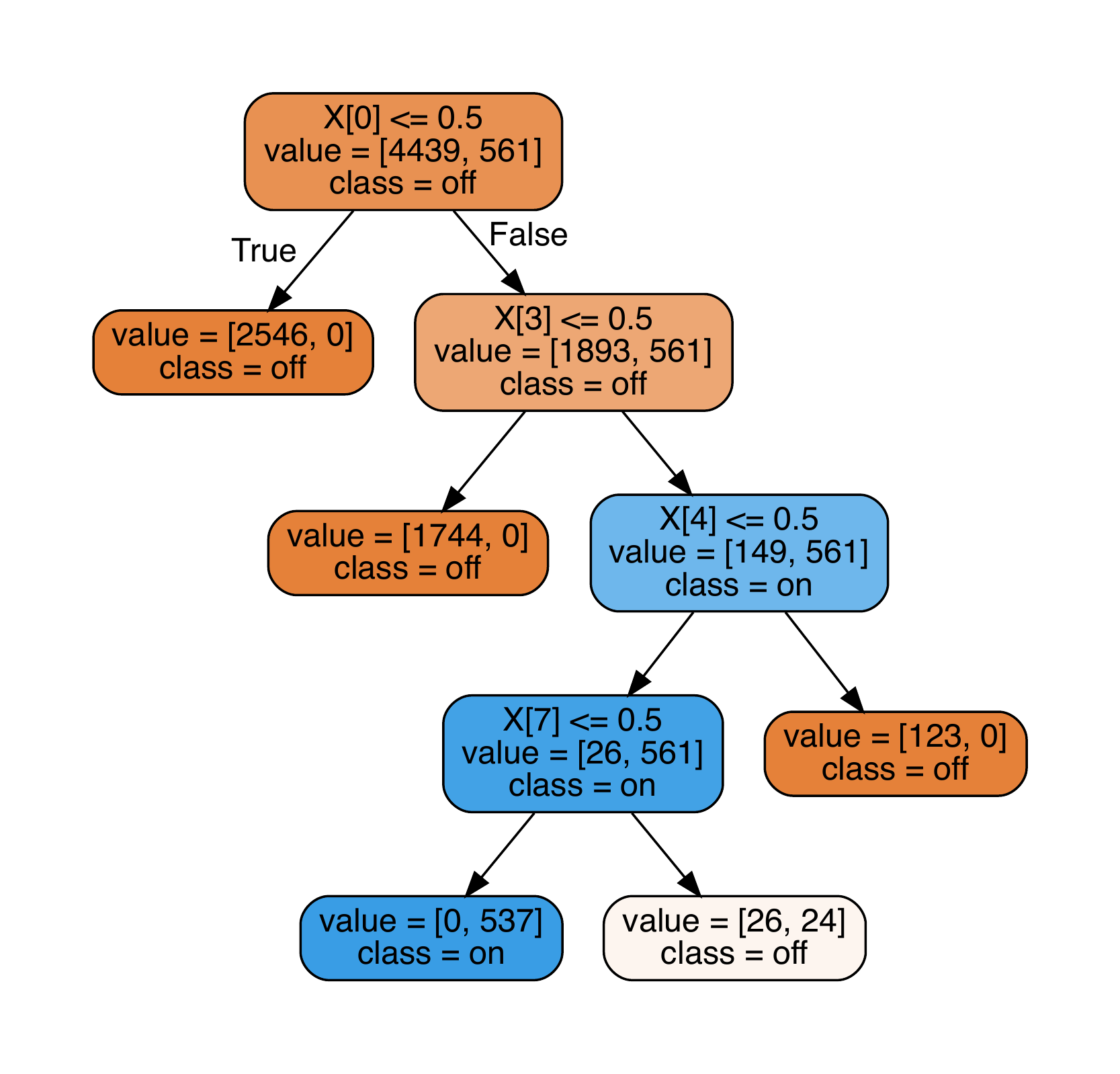}
        \label{fig:hmm2:tree:800}
        \caption{GRU $\lambda = 800$}
    \end{subfigure}
    \begin{subfigure}[b]{0.24\linewidth}
        \includegraphics[width=\linewidth]{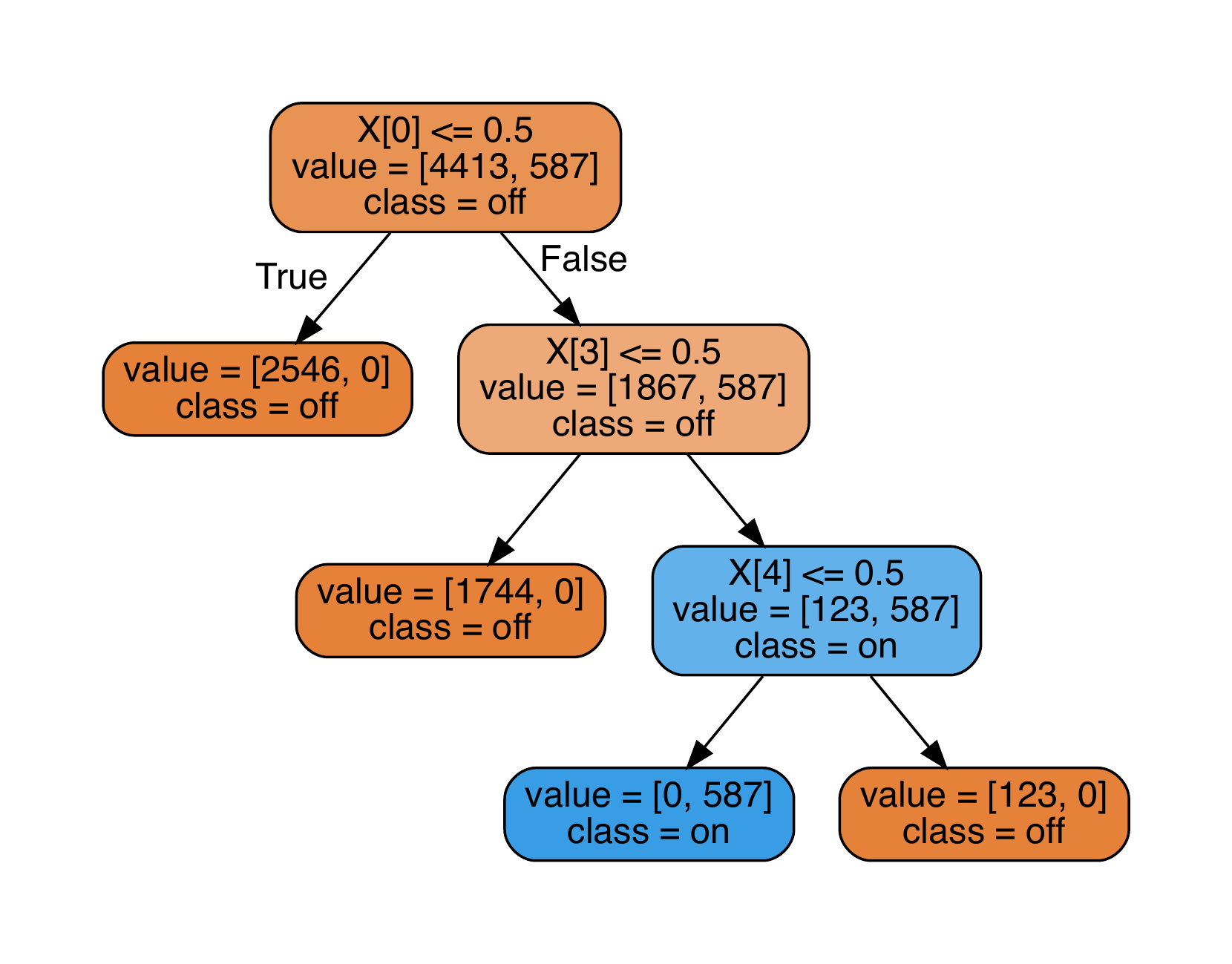}
        \label{fig:hmm2:tree:1000}
        \caption{GRU $\lambda = 1\,000$}
    \end{subfigure}
    \begin{subfigure}[b]{0.24\linewidth}
        \includegraphics[width=\linewidth]{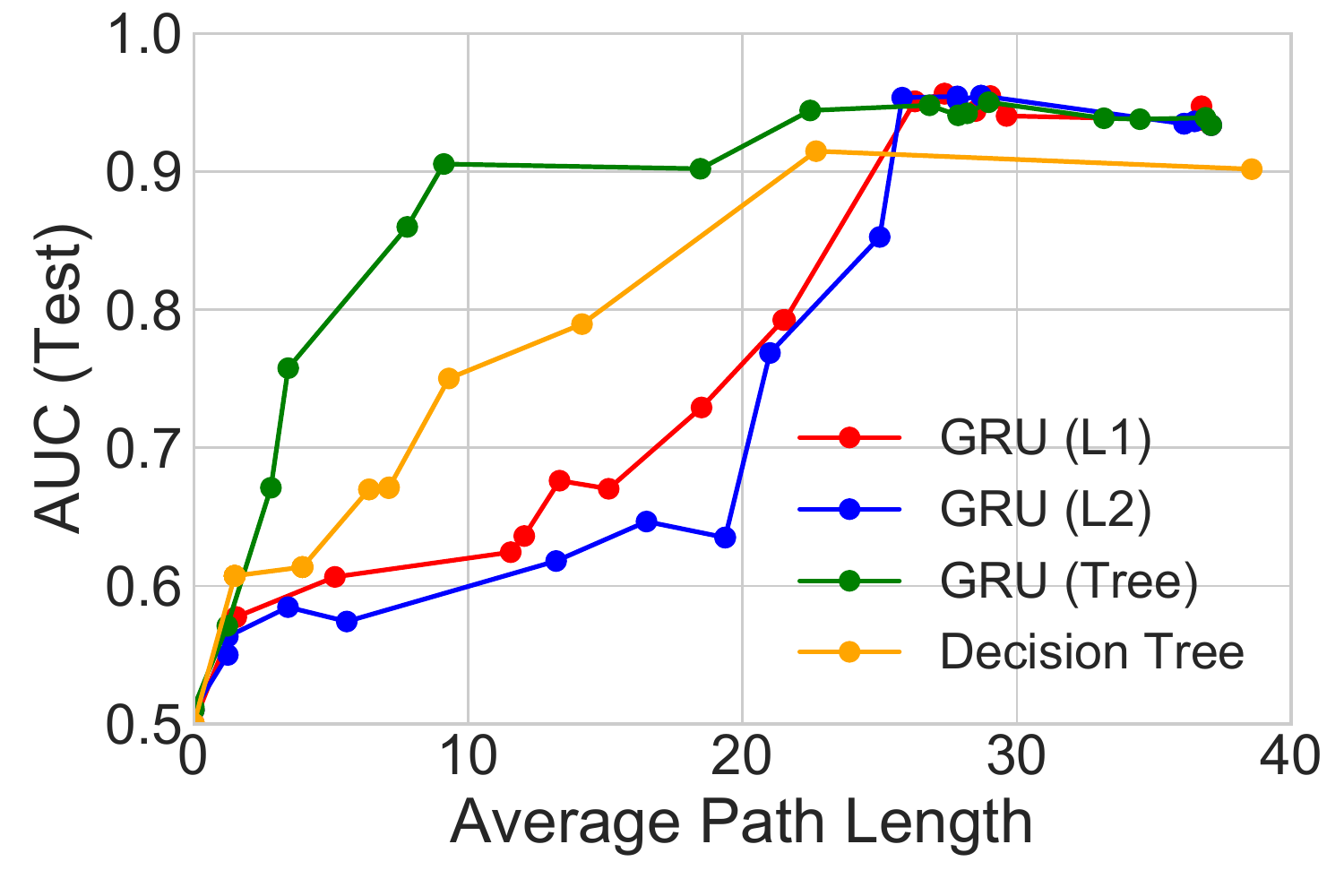}
        \caption{GRU}
        \label{fig:toy:gruhmm:plot}
    \end{subfigure}
    \caption{
    \emph{Toy Signal-and-Noise HMM Task:}
\emph{(a)-(c)} Decision trees trained to mimic predictions of GRU models with 25 hidden states
 at different regularization strengths $\lambda$; as expected, increasing $\lambda$ decreases the size of the learned trees (see supplement for more trees).
Decision tree (c) suggests the model learns to predict positive output (blue) if and only if ``x[0] == 1 and x[3] == 1 and x[4] == 0'', which is consistent with the true rule we  used to generate labels: assign positive label only if first dimension is on (x[0] == 1) and first state is active (emission probabilities for this state: [.5 .5 .5 .5 0 $\ldots$]).
\emph{(d)}
Tree-regularized GRU models reach a sweet spot of small path lengths yet high AUC predictions that alternatives cannot reach at any tested value of $\lambda$.
}
    \label{fig:results:toy-signal-and-noise-hmm}
\end{figure*}

\paragraph{Real-World Tasks:}

We tested our approach on several real tasks: predicting
medical outcomes of hospitalized septic patients, predicting HIV therapy outcomes, and
identifying stop phonemes in English speech
recordings. To normalize scales, we
independently standardized features $x$ via z-scoring.

\begin{itemize}
\item
Sepsis Critical Care: We study time-series data for 11\,786
septic ICU patients from the public MIMIC III dataset \cite{johnson2016mimiciii}. We observe at each hour $t$ a data vector $x_{nt}$ of 35 vital signs and lab results as well as a label vector $y_{nt}$ of 5 binary outcomes.
Hourly data $x_{nt}$ measures continuous features such as respiration rate (RR), blood oxygen levels (paO$_{2}$), fluid levels, and more.
Hourly binary labels $y_{nt}$ include
whether the patient died in hospital and if mechanical ventilation was applied.
Models are trained to predict all 5 output
dimensions concurrently from one shared
embedding.
The average sequence length is 15
hours. 7\,070
patients are used in training, 1\,769 for validation, and 294 for test.

\begin{figure*}
    \centering
    \begin{subfigure}[b]{0.23\linewidth}
        \includegraphics[width=\linewidth]{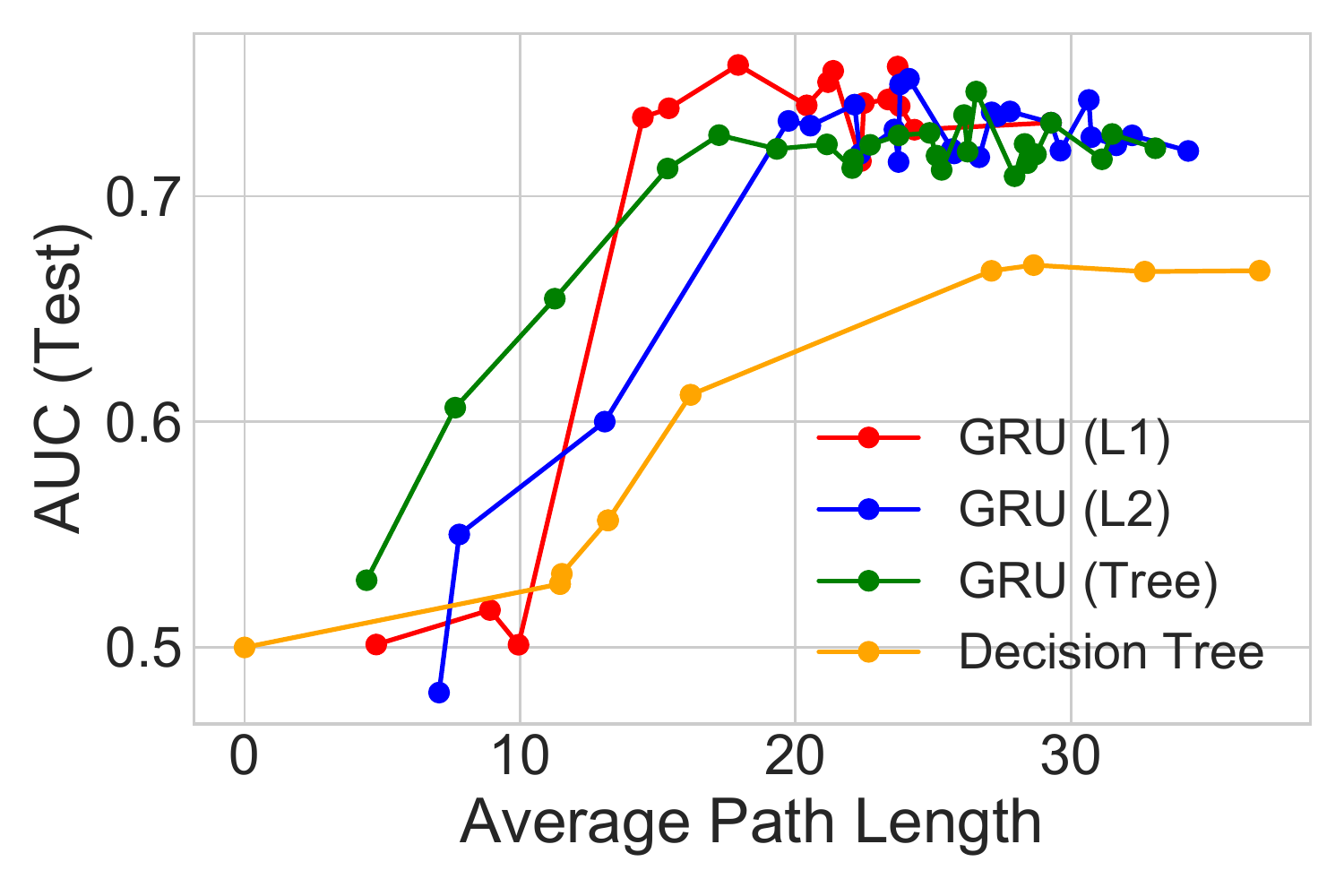}
        \caption{In-Hospital Mortality}
        \label{fig:results:sepsis:gru:mortality:trace_plots}
    \end{subfigure}
    \begin{subfigure}[b]{0.25\linewidth}
        \includegraphics[width=\linewidth,height=3.0cm]{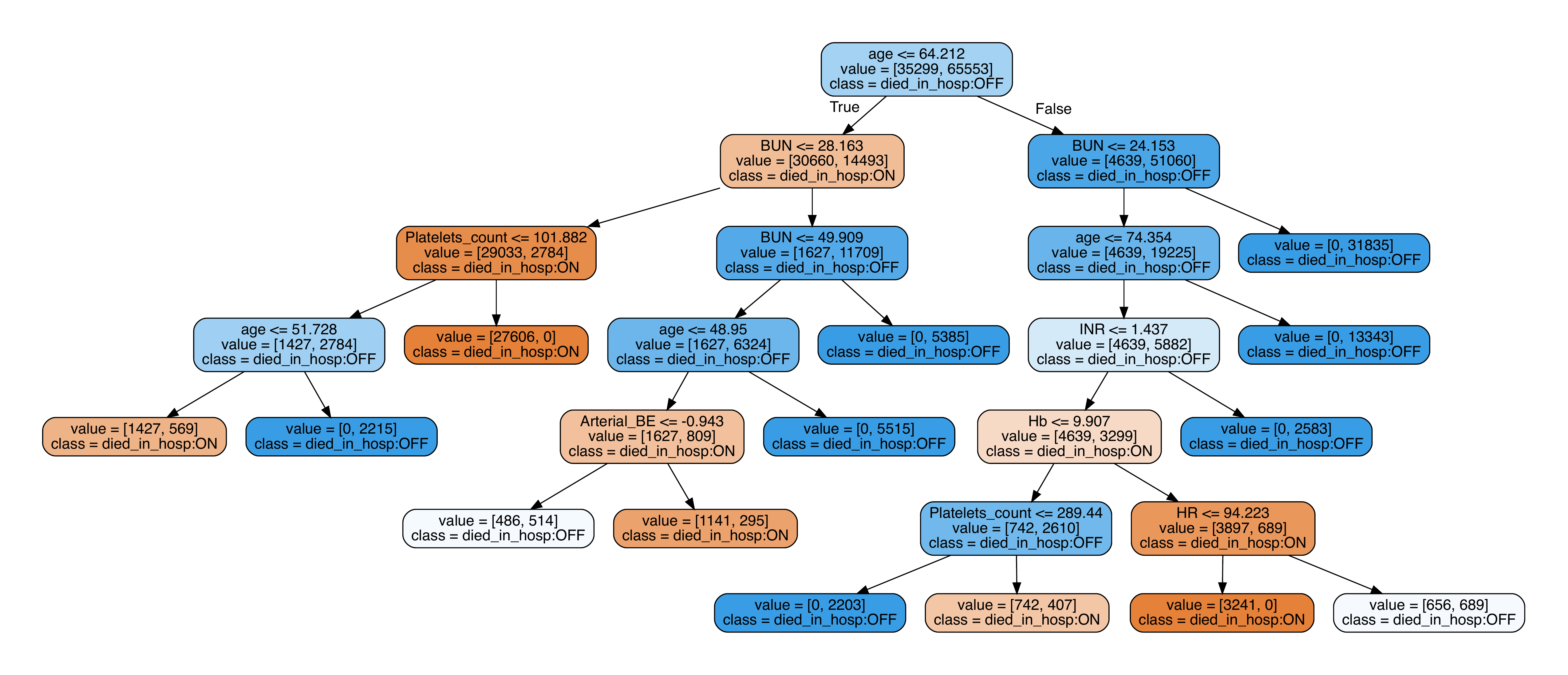}
        \caption{In-Hospital Mortality}
        \label{fig:results:sepsis:gru:mortality:tree}
    \end{subfigure}
    \begin{subfigure}[b]{0.23\linewidth}
        \includegraphics[width=\linewidth]{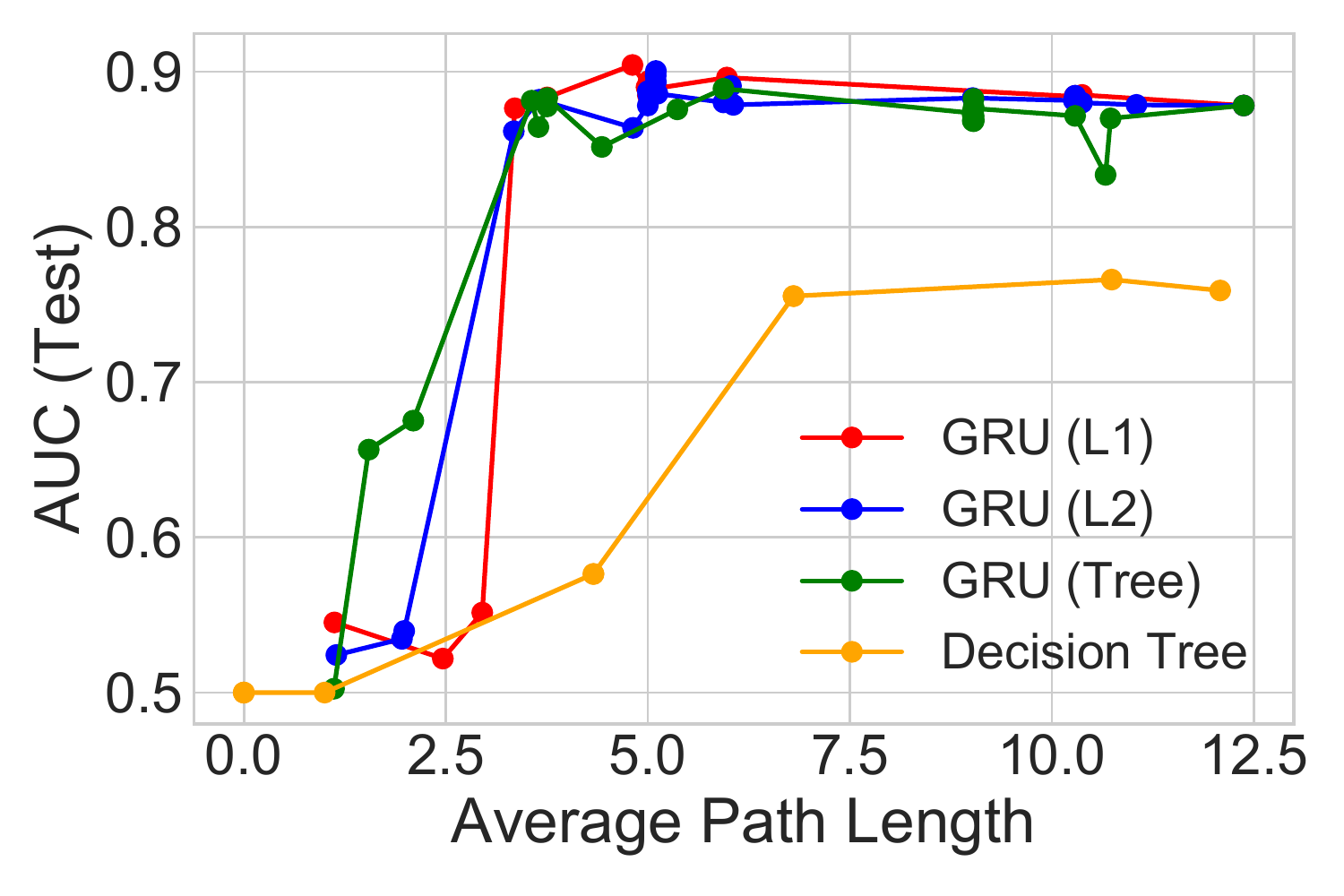}
        \caption{Mechanical Ventilation}
        \label{fig:results:sepsis:gru:mechvent:trace_plots}
    \end{subfigure}
    \begin{subfigure}[b]{0.22\linewidth}
        \includegraphics[width=\linewidth,height=3.0cm]{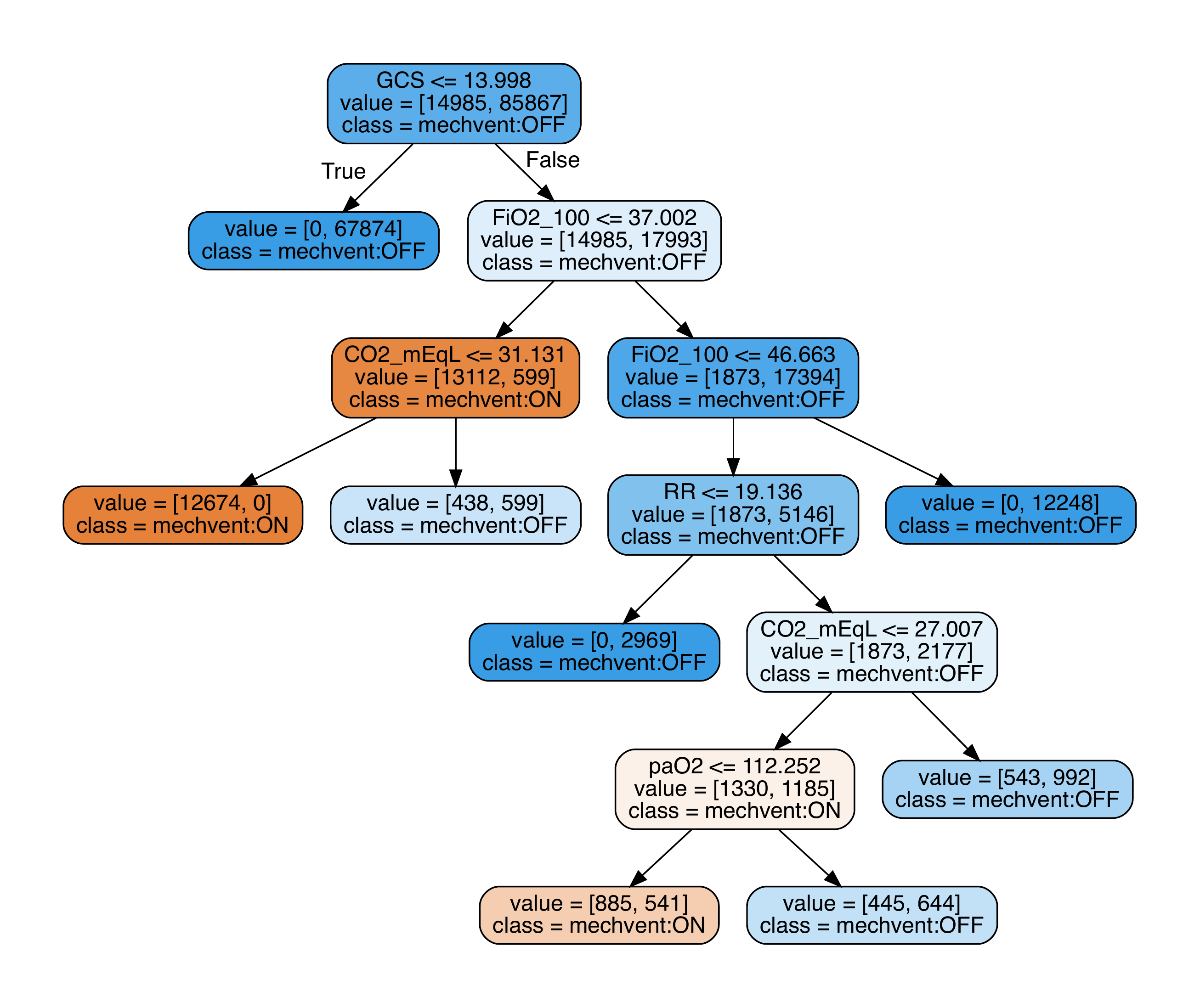}
        \caption{Mechanical Ventilation}
        \label{fig:results:sepsis:gru:mechvent:tree}
    \end{subfigure}
    \caption{
\emph{Sepsis task:}
Study of different regularization techniques for GRU model with 100 states, trained to jointly predict 5 binary outcomes.  Panels (a) and (c) show AUC vs. average path length for 2 of the 5 outcomes (remainder in the supplement); in both cases, tree-regularization provides higher accuracy in the target regime of low-complexity decision trees. Panels (b) and (d) show the associated decision trees for $\lambda = 2\,000$; these were found by clinically interpretable by an ICU clinician (see main text).
}
\label{fig:results:sepsis}
\end{figure*}

\item
HIV Therapy Outcome (HIV): We use the EuResist Integrated Database \cite{Euresist} for 53\,236  patients diagnosed with HIV. We consider 4-6 month intervals (corresponding to hospital visits) as time steps.
Each data vector $x_{nt}$ has 40 features, including blood counts, viral load measurements and lab results. Each output vector $y_{nt}$ has 15 binary labels, including
whether a therapy was successful in reducing viral load to below detection limits,
if therapy caused CD4 blood cell counts to drop to dangerous levels (indicating AIDS),
or if the patient suffered adherence issues to medication.
The average sequence length is 14 steps. 37\,618 patients are used for training; 7\,986 for testing, and 7\,632 for validation.

\item
Phonetic Speech (TIMIT): We have recordings of 630 speakers of eight major dialects of American English
reading ten phonetically rich sentences \cite{garofolo1993timit}. Each sentence contains
time-aligned transcriptions of 60 phonemes. We focus on distinguishing stop phonemes (those that stop the flow of air, such as ``b'' or ``g'') from non-stops.
Each timestep has one binary label $y_{nt}$ indicating if a stop phoneme occurs or not.
Each input $x_{nt}$ has 26
continuous features: the acoustic signal's Mel-frequency cepstral
coefficients and derivatives.  There are 6\,303
sequences, split into 3\,697 for training, 925 for validation,
and 1\,681 for testing. The average length is 614.
\end{itemize}

\begin{figure*}
    \centering{
    \begin{subfigure}[b]{0.23\linewidth}
        \includegraphics[width=\linewidth]{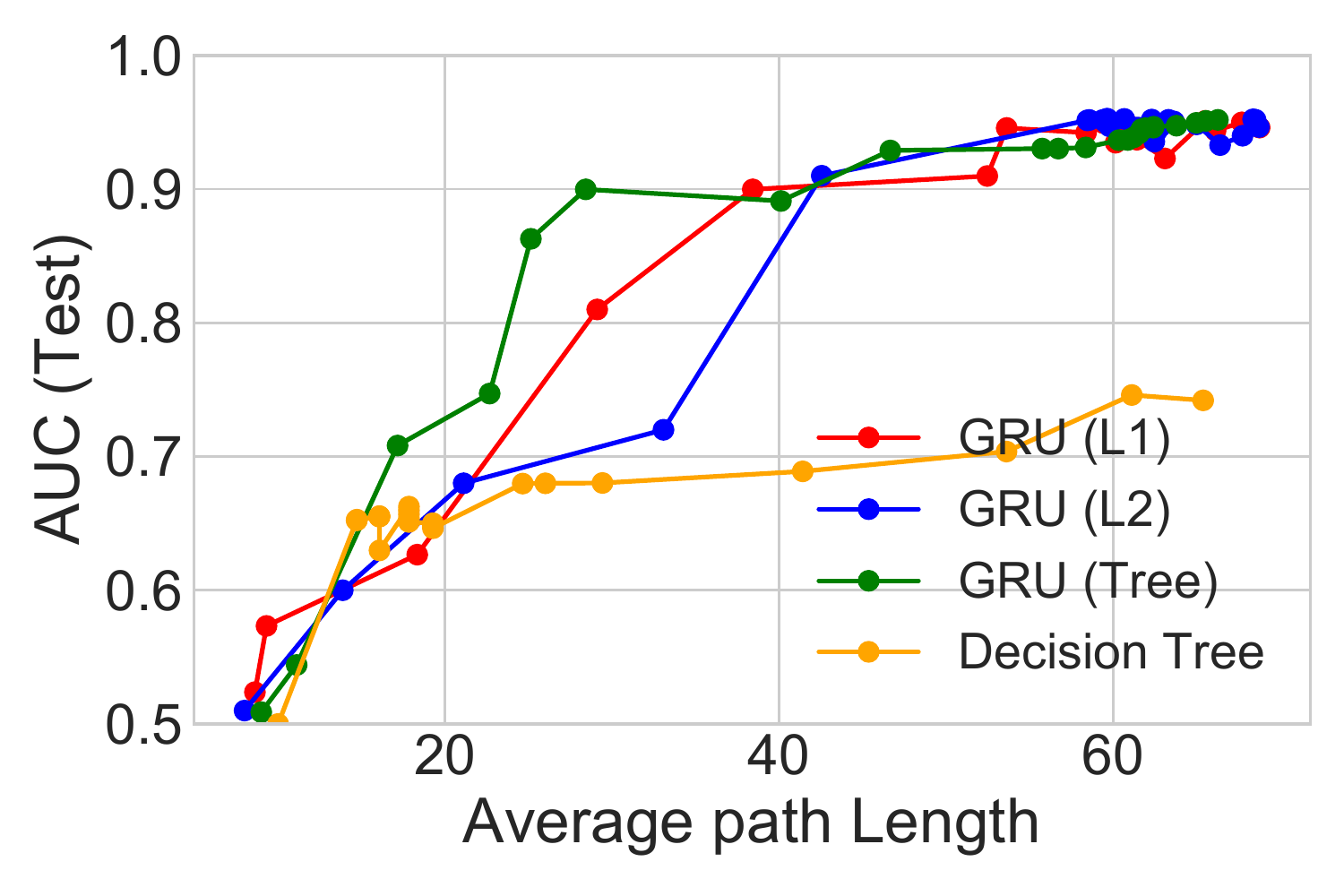}
        \caption{TIMIT Stop Phonemes}
        \label{fig:timit:gru:trace_plots}
    \end{subfigure}
    \begin{subfigure}[b]{0.23\linewidth}
        \includegraphics[width=\linewidth]{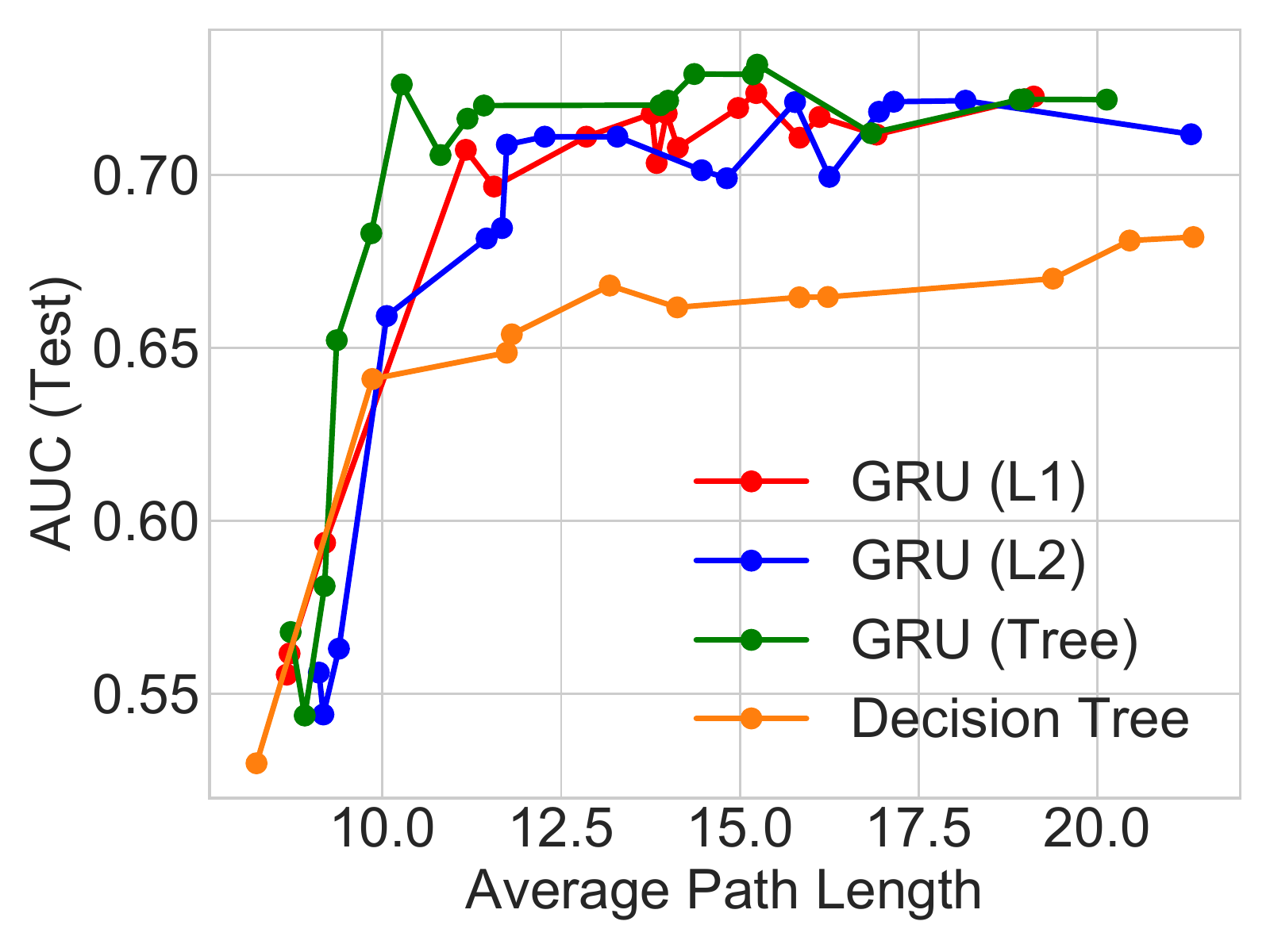}
        \caption{HIV:  CD4$^{+}$ $\leq$ 200 cells/ml}
        \label{fig:hiv:cd4}
    \end{subfigure}
       \begin{subfigure}[b]{0.23\linewidth}
        \includegraphics[width=\linewidth]{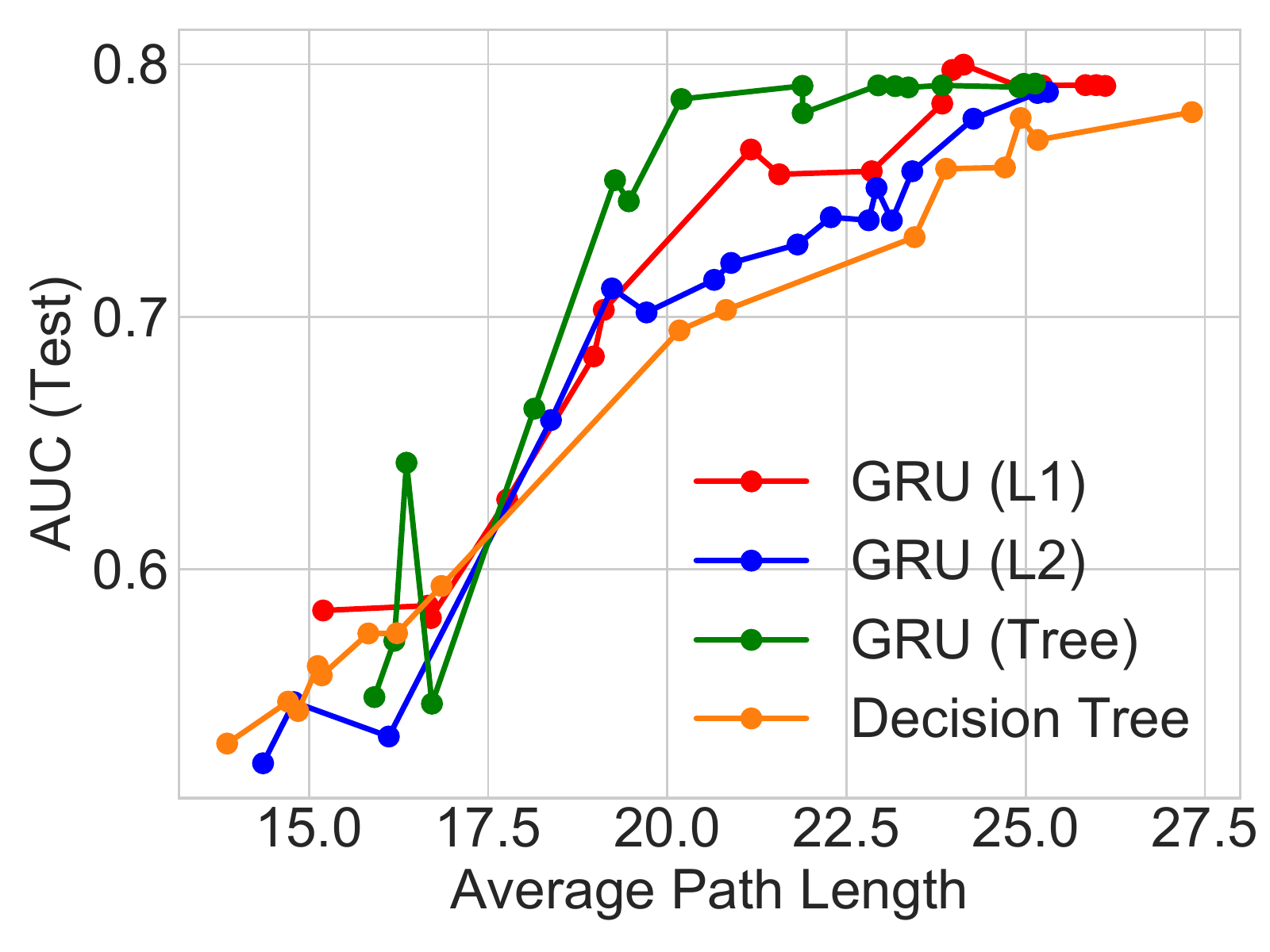}
        \caption{HIV Therapy Adherence}
        \label{fig:hiv:therapy}
    \end{subfigure}
    \begin{subfigure}[b]{0.23\linewidth}
        \includegraphics[width=\linewidth]{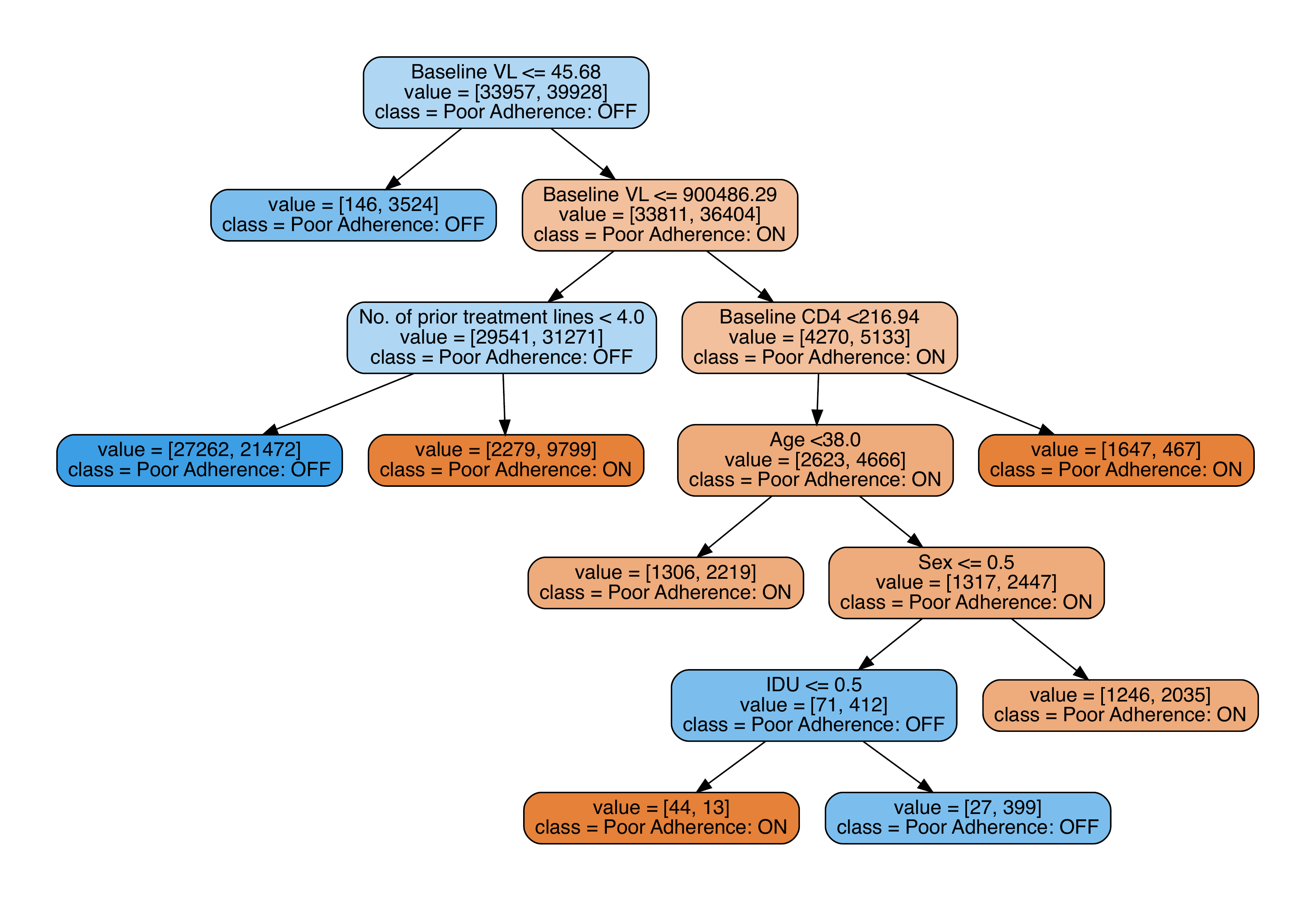}
        \caption{HIV Therapy Adherence}
        \label{fig:hiv:adherence}
    \end{subfigure}
    \label{fig:results:timit}}
\caption{\emph{TIMIT and HIV tasks:}
Study of different regularization techniques for GRU model with 75 states. Panels (a)-(c) are tradeoff curves showing how AUC predictive power and decision-tree complexity evolve with increasing regularization strength under L1, L2 or tree regularization on both TIMIT and HIV tasks. The GRU is trained to jointly predict 15 binary outcomes for HIV, of which 2 are shown here in Panels (b) - (c). The GRU's decision tree proxy for HIV Adherence is shown in (d).}
\end{figure*}

\subsection{Results}
The major conclusions of our experiments comparing GRUs with various regularizations are outlined below.

\paragraph{Tree-regularized models have fewer nodes than other forms of regularization.}
Across tasks, we see that in the target regime of small decision trees
(low average-path lengths), our proposed tree-regularization achieves
higher prediction quality (higher AUCs).  In the signal-and-noise HMM
task, tree regularization (green line in
Fig.~\ref{fig:results:toy-signal-and-noise-hmm}(d)) achieves AUC
values near 0.9 when its trees have an average path length of
10. Similar models with L1 or L2 regularization reach this AUC only
with trees that are nearly double in complexity (path length over 25).
On the Sepsis task (Fig.~\ref{fig:results:sepsis})
we see AUC gains of 0.05-0.1 at path lengths of 2-10.
On the TIMIT task (Fig.~\ref{fig:timit:gru:trace_plots}), we see
AUC gains of 0.05-0.1 at path lengths of 20-30.
Finally, on the HIV CD4 blood cell count task in Fig.~\ref{fig:hiv:cd4}, we see AUC differences of between 0.03 and 0.15 for path lengths of 10-15. The HIV adherence task in Fig.~\ref{fig:hiv:adherence} has AUC gains of between 0.03 and 0.05 in the path length range of 19 to 25 while at smaller paths all methods are quite poor, indicating the problem's difficulty.
Overall, these AUC gains are particularly useful in determining how to administer subsequent HIV therapies.

We emphasize that our tree-regularization usually achieves a sweet spot of high AUCs at short path lengths not possible with standalone decision trees (orange lines),
L1-regularized deep models (red lines) or L2-regularized deep models (blue lines).
In unshown experiments, we also tested elastic net regularization~\cite{zou2005elasticnet}, a
 linear combination of L1 and L2 penalities.
We found elastic nets to
follow the same trend lines as L1 and L2, with no visible differences.
In domains where human-simulatability is required, increases in prediction
accuracy in the small-complexity regime can mean the difference between models that provide value on
a task and models that are unusable, either because performance
is too poor or predictions are uninterpretable.

\paragraph{Our learned decision tree proxies are interpretable.}
Across all tasks, the decision trees which mimic the predictions of tree-regularized deep models are small enough to simulate by hand (path length $\leq 25$) and help users grasp the model's nonlinear prediction logic.
Intuitively, the trees for our synthetic task in Fig.~\ref{fig:results:toy-signal-and-noise-hmm}(a)-(c) decrease in size as the strength $\lambda$ increases. The logic of these trees also matches the true labeling process: even the simplest tree (c) checks a relevant subset of input dimensions necessary to verify that both the first state and the first output dimension are active.

In Fig.~\ref{fig:results:sepsis}, we show
decision tree proxies for our deep models on two sepsis prediction tasks: mortality and need for ventilation.  We consulted a clinical expert on sepsis treatment, who noted that
the trees helped him understand what the models might be doing and thus determine if he would trust the deep model.
For example, he said that using FiO$_{2}$, RR, CO$_{2}$
and paO$_{2}$ to predict need for mechanical ventilation (Fig.~\ref{fig:results:sepsis:gru:mechvent:tree}) was sensible, as these all measure breathing quality.
In contrast, the in-hospital mortality tree (Fig.~\ref{fig:results:sepsis:gru:mortality:tree})
predicts that some young patients with no organ failure have high mortality rates while other young patients with organ failure have low mortality.  These counter-intuitive results led to hypotheses about how uncaptured variables impact the
training process. Such reasoning would not be possible from
simple sensitivity analyses of the deep model.

Finally, we have verified that the decision tree proxies of our tree-regularized deep models of the HIV task in Fig. \ref{fig:hiv:adherence} are interpretable for understanding why a patient has trouble adhering to a prescription; that is, taking drugs regularly as directed.
Our clinical collaborators confirm that the baseline viral load and number of prior treatment lines, which are prominent attributes for the decisions in Fig. \ref{fig:hiv:adherence}, are useful predictors of a patient with adherence issues. Several medical studies~\cite{Langford2007,JIA2} suggest that patients with higher baseline viral loads tend to have faster disease progression, and hence have to take several drug cocktails to combat resistance. Juggling many drugs typically makes it difficult for these patients to adhere as directed.
We hope interpretable predictive models for adherence could help assess a patient's overall prognosis~\cite{paterson2000adherence} and offer opportunities for intervention (e.g. with alternative single-tablet regimens).
\paragraph{Decision trees trained to mimic deep models make faithful predictions.}
Across datasets, we find that each tree-regularized deep time-series model has predictions that agree with its corresponding decision tree proxy in about 85-90\% of test examples. Table 1 shows exact fidelty scores for each dataset. Thus, the simulatable paths of the decision tree will be trustworthy in a majority of cases.

\paragraph{Practical runtimes for tree regularization are less than twice that of simpler L2.}
While our tree-regularized GRU with 10 states takes 3977 seconds per epoch on TIMIT, a similar L2-regularized GRU takes 2116 seconds per epoch.
Thus, our new method has cost less than twice the baseline \emph{even when the surrogate is serially computed}.
Because the surrogate $\hat{\Omega}(W)$ will in general be
a much smaller model than the predictor $\hat{y}(x,W)$, we expect one could get faster per-epoch times
by parallelizing the creation of $(W,\Omega(W))$ training pairs and the training of the surrogate $\hat{\Omega}(W)$.
Additionally, 3977 seconds includes the time needed to train
the surrogate. In practice, we do this sparingly, only once every 25 epochs, yielding an amortized per-epoch cost of 2191 seconds (more runtime results are in the supplement).

\paragraph{Decision trees are stable over multiple optimization runs.}
When tree regularization is strong (high $\lambda$), the decision trees trained to match the predictions of deep models are stable. For both signal-and-noise and sepsis tasks, multiple runs from different random restarts have nearly identical tree shape and size, perhaps differing by a few nodes. This stability is crucial to building trust in our method. On the signal-and-noise task ($\lambda = 7000$), 7 of 10 independent runs with random initializations resulted in trees of exactly the same structure, and the others closely resembled those sharing the same subtrees and features (more details in supplement).

\begin{table}[h!]
    \centering
    \begin{tabular}{ l | c}
        Dataset & Fidelity \\
        \hline
        signal-and-noise HMM & 0.88 \\
        SEPSIS (In-Hospital Mortality) & 0.81\\
        SEPSIS (90-Day Mortality) & 0.88\\
        SEPSIS (Mech. Vent.) & 0.90\\
        SEPSIS (Median Vaso.) & 0.92\\
        SEPSIS (Max Vaso.) & 0.93\\
        HIV (CD4$^{+}$ below 200) & 0.84 \\
        HIV (Therapy Success) & 0.88 \\
        HIV (Mortality) & 0.93\\
        HIV (Poor Adherence) & 0.90 \\
        HIV (AIDS Onset) & 0.93\\
        TIMIT & 0.85\\
    \end{tabular}
    \caption{
    Fidelity of predictions from our trained deep GRU-RNN and its corresponding decision tree.
    Fidelity is defined as the percentage of test examples on which the prediction made by a tree agrees with the deep model~\cite{craven1996extracting}.
     We used 20 hidden GRU states for signal-and-noise task, 50 states for all others.
    }
    \label{table:fidelity}
\end{table}

\begin{figure}[h]
    \begin{subfigure}[b]{0.49\linewidth}
        \centering
        \includegraphics[width=\linewidth]{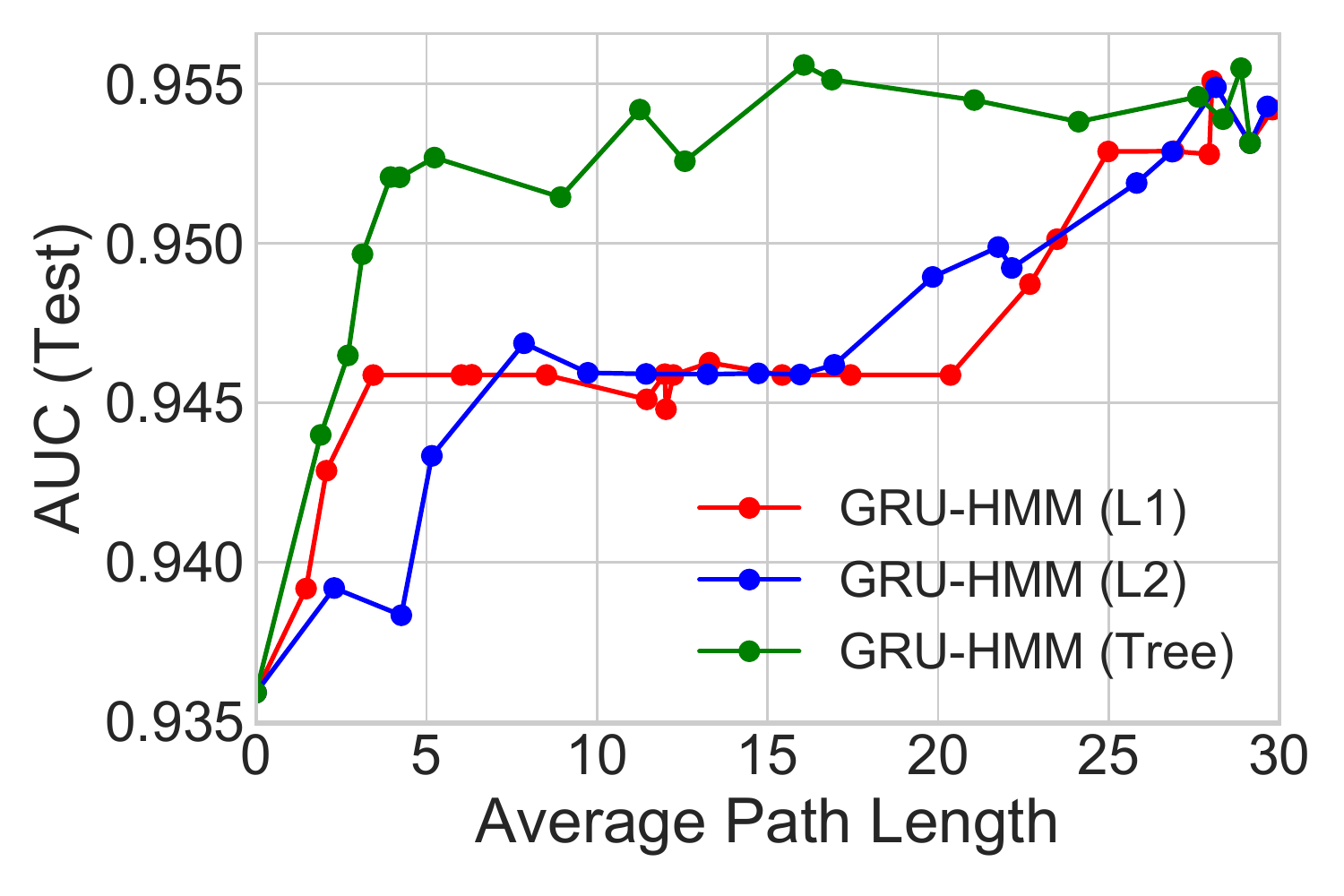}
        \caption{Signal-and-noise {20+5}}
        \label{fig:2hmm:gruhmm:plot}
    \end{subfigure}
    \begin{subfigure}[b]{0.49\linewidth}
        \centering
        \includegraphics[width=\linewidth]{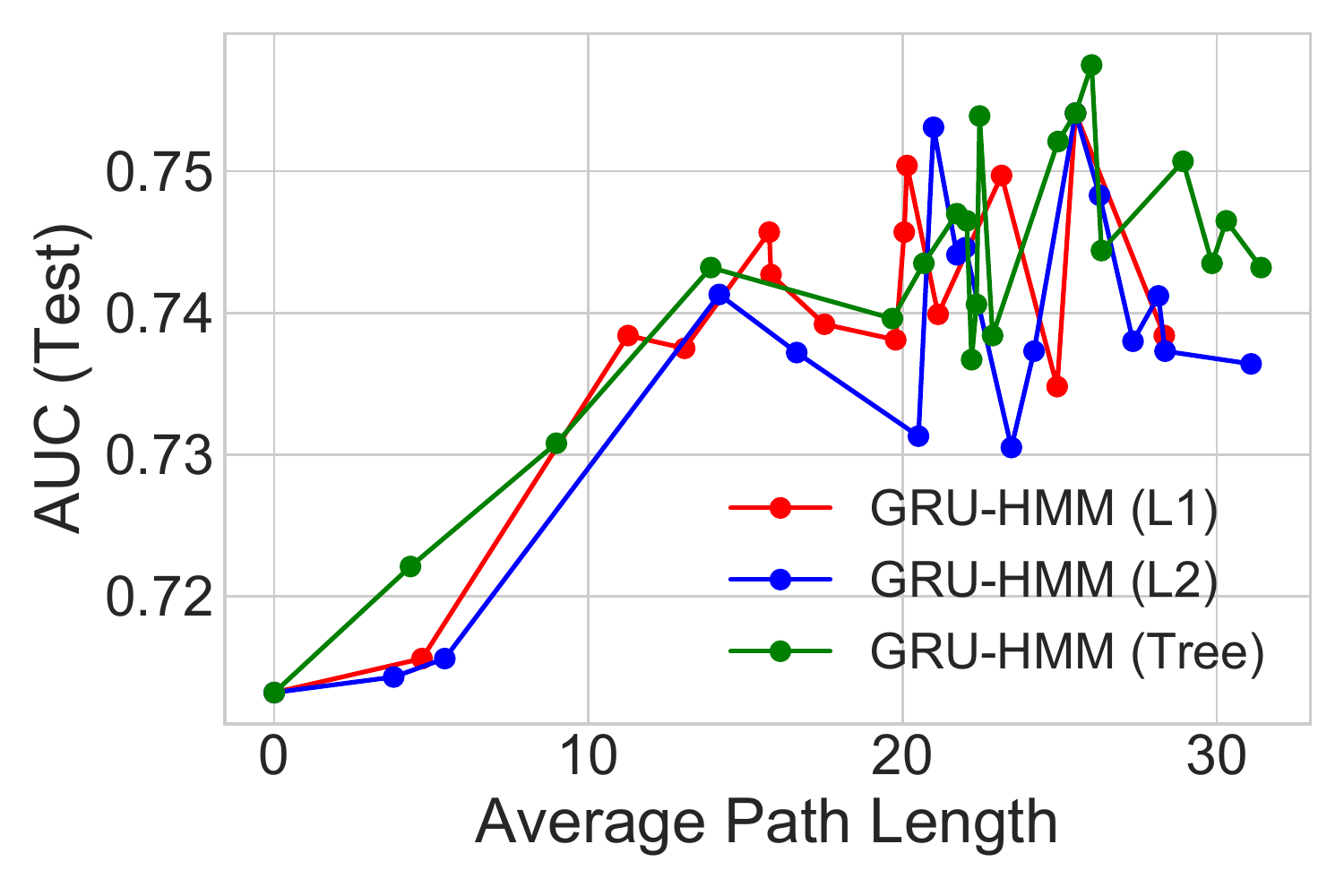}
        \caption{In-Hosp. Mort. {50+50}}
        \label{fig:sepsis:mortality:gruhmm:plot}
    \end{subfigure}
    \begin{subfigure}[b]{0.49\linewidth}
        \centering
        \includegraphics[width=\linewidth]{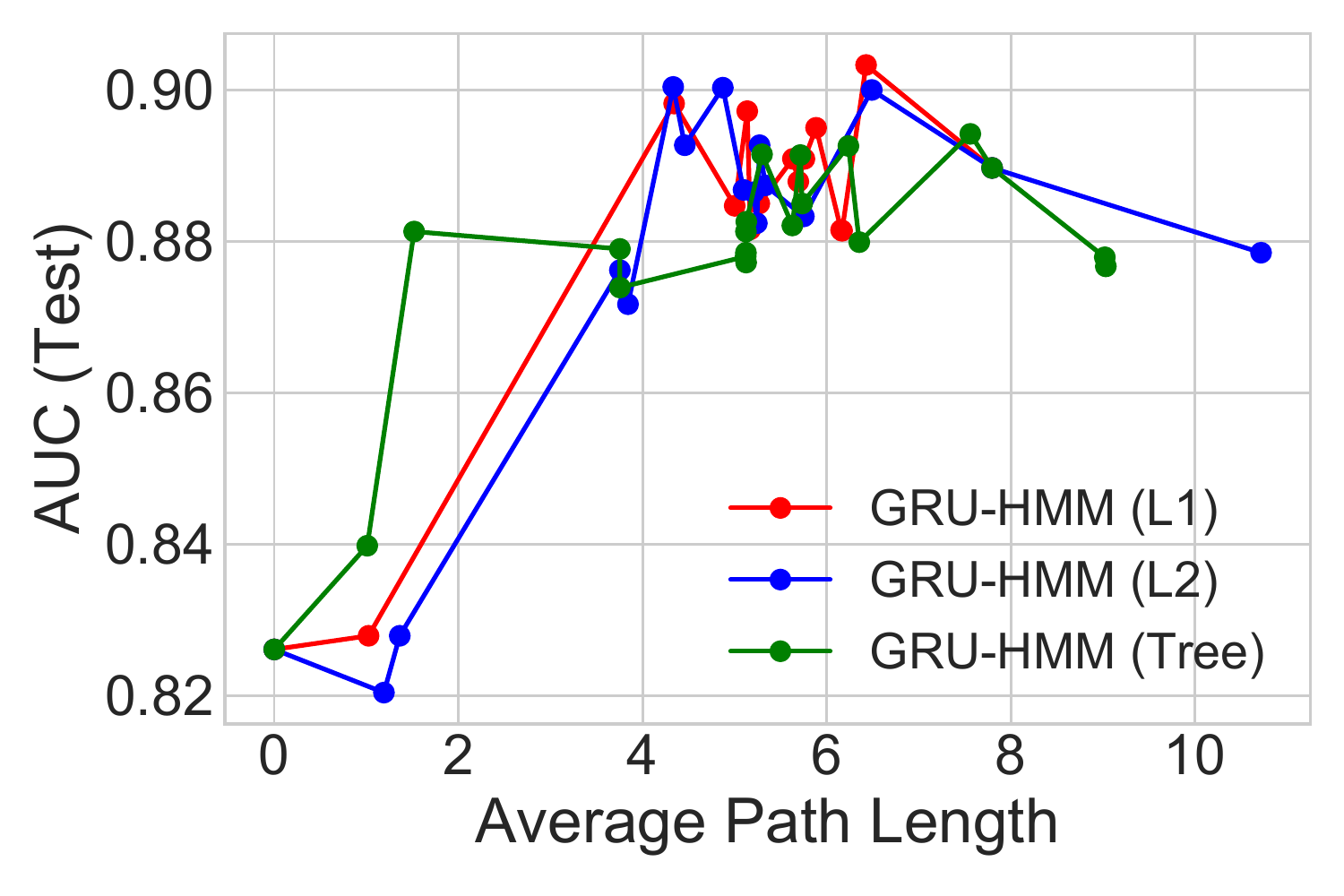}
        \caption{Mech. Vent. {50+50}}
        \label{fig:sepsis:vent:gruhmm:plot}
    \end{subfigure}
    \begin{subfigure}[b]{0.49\linewidth}
        \centering
        \includegraphics[width=\linewidth]{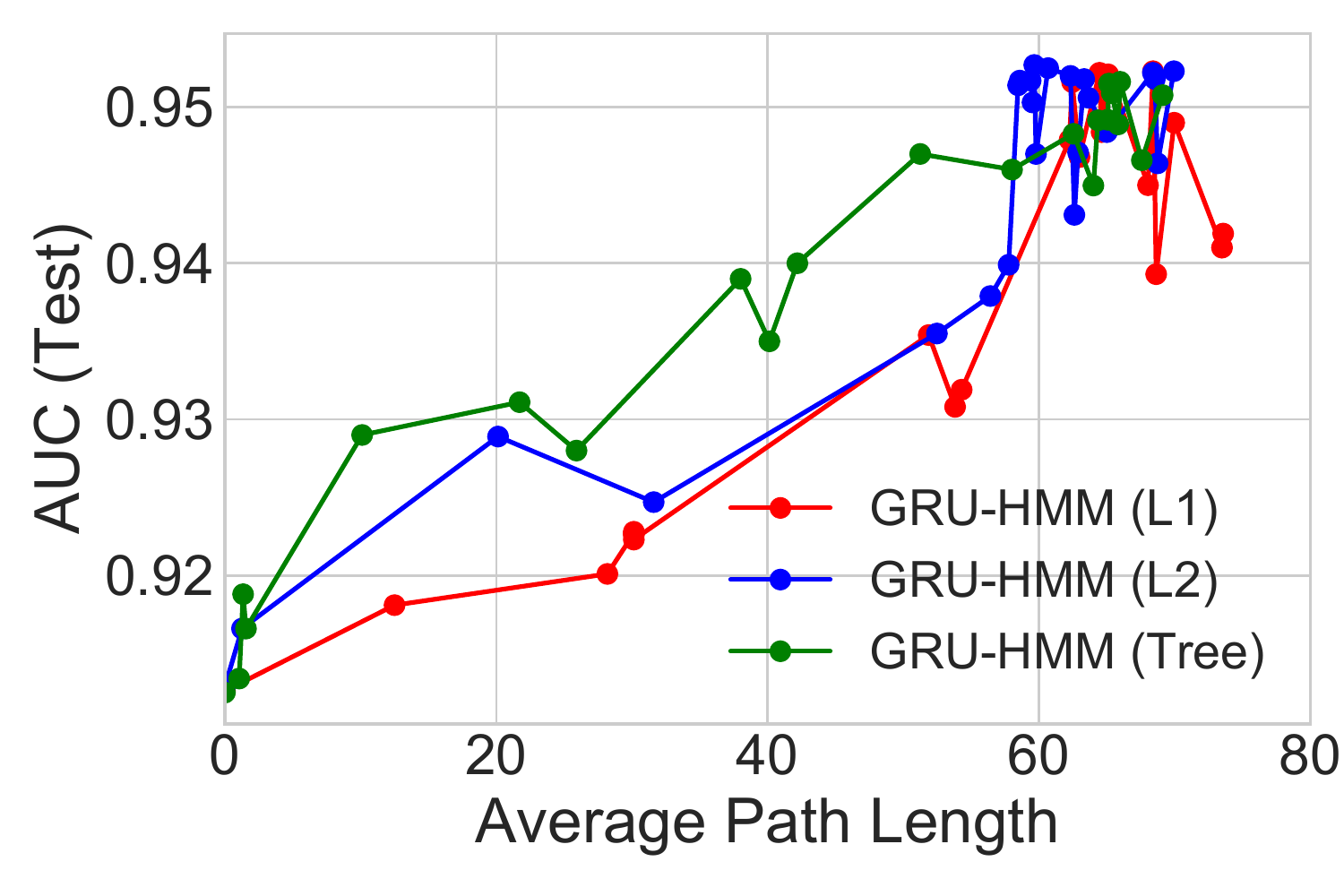}
        \caption{Stop Phonemes {50+25}}
        \label{fig:timit:gruhmm:plot}
    \end{subfigure}
    \caption{
    Fitness curves for the GRU-HMM, showing prediction quality (AUC) vs. complexity (path length) across range of regularization strengths $\lambda$.
    Captions show the number of HMM states plus the number of GRU states.
    See earlier figures to compare these GRU-HMM numbers to simpler GRU and decision tree baselines.
}
\label{fig:results:gruhmm}
\end{figure}

\paragraph{The deep residual GRU-HMM achieves high AUC with less complexity.}
So far, we have focused on regularizing standard deep models, such as
MLPs or GRUs.  Another option is to use a deep model as a residual
on another model that is already interpretable: for example,
discrete HMMs partition timesteps into clusters, each of which can be inspected,
but its predictions might have limited accuracy.
In Fig.~\ref{fig:results:gruhmm}, we show the performance of jointly training a \emph{GRU-HMM},
a new model which combines an HMM with a tree-regularized GRU to improve its predictions (details and further results in the
supplement). Here, the ideal path length is zero, indicating only the HMM makes predictions. For small average-path-lengths, the GRU-HMM improves the original HMM's
predictions \emph{and} has simulatability gains over earlier GRUs.
On the mechanical ventilation task, the GRU-HMM requires an average path length of only 28 to reach AUC of 0.88, while the GRU alone with the same number of states requires a path length of 60 to reach the same AUC.
This suggests that jointly-trained deep residual models may provide even better interpretability.

\section{Discussion and Conclusion}
We have introduced a novel tree-regularization technique that encourages the complex decision boundaries of any differentiable model to be well-approximated by human-simulatable functions, allowing domain experts to quickly understand and approximately \emph{compute} what the more complex model is doing.  Overall, our training procedure is robust and efficient; future work could continue to explore and increase the stability of the learned models as well as identify ways to apply our approach to situations in which the inputs are not inherently interpretable (e.g. pixels in an image).

Across three complex, real-world domains -- HIV treatment, sepsis treatment, and human speech processing -- our tree-regularized models provide gains in prediction accuracy in the regime of simpler, approximately human-simulatable models.  Future work could apply tree regularization to local, example-specific approximations of a loss~\cite{ribeiro2016should} or to representation learning tasks (encouraging embeddings with simple boundaries).  More broadly, our general training procedure could apply tree-regularization or other procedure-regularization to a wide class of popular models, helping us move beyond sparsity toward models humans can easily simulate and thus trust.





\section*{Acknowledgements}
MW is supported by the U.S. National Science Foundation. MCH is supported by Oracle Labs. SP is supported by the Swiss National Science Foundation project 51MRP0\_158328. The authors thank the EuResist Network for providing HIV data for this study, and thank Matthieu Komorowski for the preprocessed sepsis data \cite{raghu2017sepsis}. Computations were supported by the FAS Research Computing Group at Harvard and sciCORE (http://scicore.unibas.ch/)~scientific computing core facility at University of Basel.

\bibliographystyle{aaai}
\bibliography{paper}

\counterwithin{figure}{section}
\counterwithin{table}{section}
\appendix
\newpage
\onecolumn

\begin{center}
{\textbf \Large Supplementary Material}
\end{center}

\section{Details for Decision-Tree Training}

\paragraph{Training decision trees with post-pruning.}
Our average path length function $\Omega(W)$ for determining the complexity of a deep model with parameters $W$ -- defined in the main paper in Alg. 1 --  
assumes that we have a robust,
black-box way to train binary decision-trees called \textsc{TrainTree}
given a labeled dataset $\{x_n, \hat{y}_n \}$. For this we use
the \texttt{DecisionTree} module distributed in Python's sci-kit
learn, which optimizes information gain with Gini impurity.  The specific syntax we use
(for reproducibility) is:
\begin{verbatim}
tree = DecisionTree(min_sample_count=5)
tree.fit(x_train, y_train)
tree = prune_tree(tree, x_valid, y_valid)
\end{verbatim}
The provided keyword options force the tree to have at least 5 examples from the training set in every leaf.  We found that tuning hyperparameters of the \textsc{TrainTree} subprocedure, such as the minimum size of a leaf node, to be important for making useful trees.

Generally, the runtime cost of sklearn's fitting procedure scales superlinearly with the number of examples $N$ and linearly with the number of features $F$ -- a total complexity of $O(FN\log(N))$. In practice, we found that with $N=1000$ examples, $F=10$ features, tree construction takes 15.3 microseconds.

The pruning procedure is a heuristic to create simpler trees, summarized in algorithm \ref{alg:tree_prune}. After \textsc{TrainTree} delivers a working decision tree,
we iterative propose removing each remaining leaf node, accepting the proposal if the squared prediction error on a validation set improves.
This pruning removes sub-trees that don't generalize to unseen data.

\begin{algorithm}
\caption{Post-pruning for training decision trees.}
\begin{algorithmic}[1]
\Require{
\Statex $ T $ : initial decision tree
\Statex $\textsc{ErrOnVal}(\cdot)$ : squared error on validation data
\Statex $\qquad \textsc{ErrOnVal}(T) \triangleq \sum_{n=1}^N (T(x_n) - y_n )^2$
}
\Procedure{PruneTree}{ $T$, $err$ }
\State $e \gets \textsc{ErrOnVal}(T)$.
\For{node $n \in \textsc{SortLeafToRoot}(T.nodes)$}
    \State $T' \gets \textsc{RemoveNode}(T, n)$
    \State $e_{new} \gets \textsc{ErrOnVal}(T')$
    \If{$e_{new} < e$} $T \gets T'$
    \EndIf
\EndFor
\State Return $T$
\EndProcedure
\end{algorithmic}
\label{alg:tree_prune}
\end{algorithm}

\begin{figure}[!t]
    \captionsetup[subfigure]{aboveskip=-5pt,belowskip=-2pt}
    \centering
    \begin{subfigure}[b]{0.7\linewidth}
        \includegraphics[width=\linewidth]{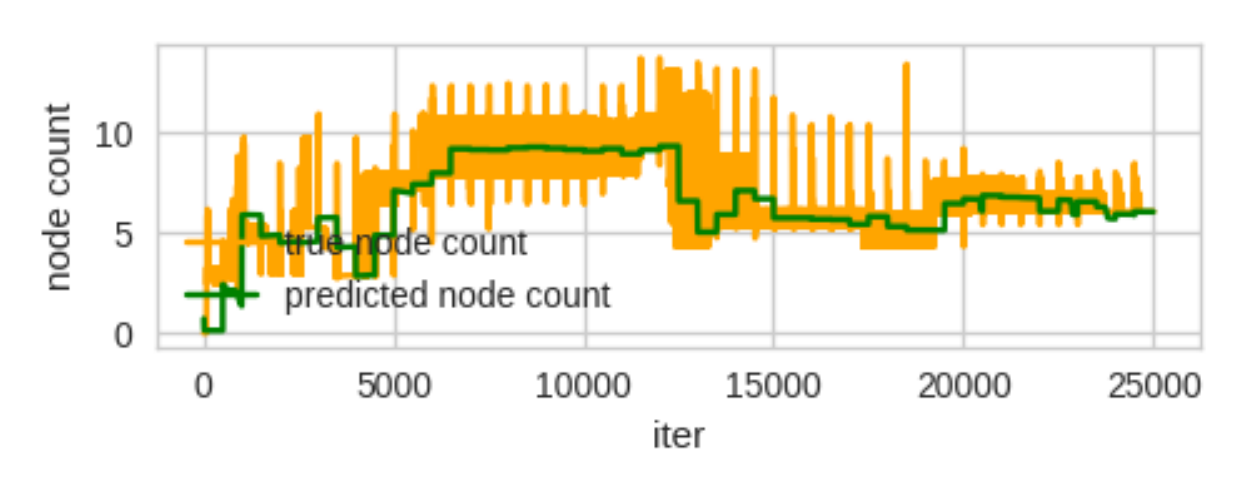}
        \caption{Path length estimates $\hat{\Omega}$ for 2D Parabola task}
        \label{fig:regressor:parabola}
    \end{subfigure}
    %
    \begin{subfigure}[b]{0.7\linewidth}
        \includegraphics[width=\linewidth]{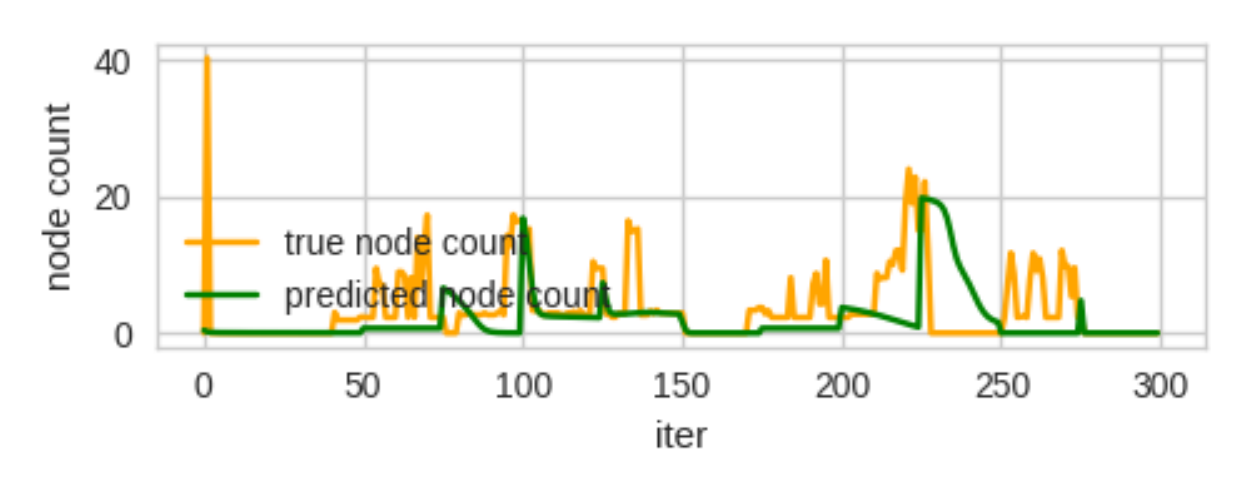}
        \caption{Path length estimates $\hat{\Omega}$ for Signal-and-noise HMM task}
        \label{fig:regressor:2hmm}
    \end{subfigure}
\caption{
True average path lengths (yellow) and surrogate estimates
$\hat{\Omega}$ (green) across many iterations of network parameter
training iterations.  }
\label{fig:nodecount-training}
\end{figure}

\paragraph{Sanity check: Surrogate path length closely follow true path length.}
Fig.~\ref{fig:nodecount-training} shows that our
surrogate predictor $\hat{\Omega}(\cdot)$ tracks the true average path
length as we train the target predictor $\hat{y}(\cdot, W)$ on several different datasets.

\paragraph{Sensitivity to different choices for surrogate training.}

\begin{figure}[!h]
    \center
    \includegraphics[width=0.9\linewidth]{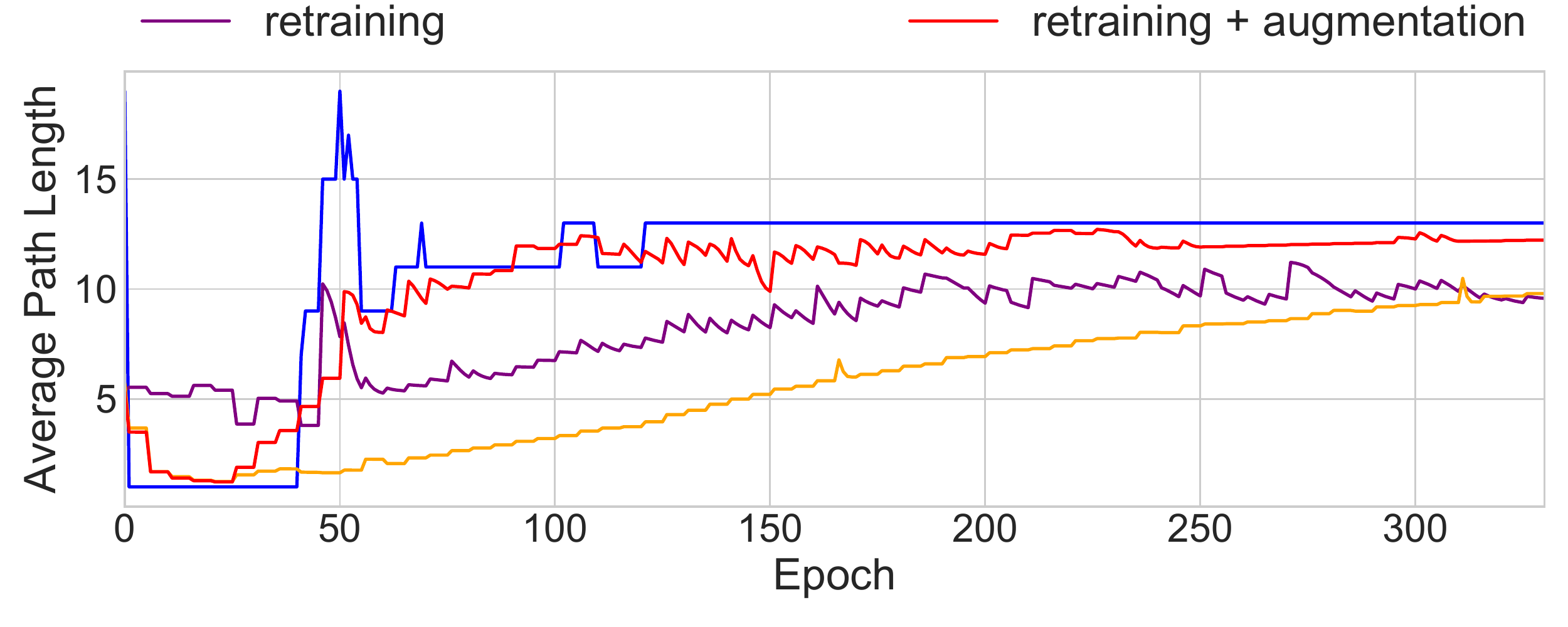}
    \caption{This figure shows the effects of weight augmentation and retraining. The blue line is the true average path length of the decision tree at each epoch. All other lines show predicted path lengths using the surrogate MLP. By randomly sampling weights and intermittently retraining the surrogate, we significantly improve the ability of the surrogate model to track the changes in the
    ground truth.}
    \label{fig:weight2node_tricks}
\end{figure}

In Fig.~\ref{fig:weight2node_tricks}, we show sample learning
curves for variations of methods for approximating the average path length (also called ``node count'') in a decision tree. In blue is the true value. Each of the other 3 lines use the same surrogate model: an MLP with 25 hidden nodes. Increasing its capacity too much, i.e. 100 hidden nodes, leads to overfitting where the surrogate is able to predict the average path length extremely well for a small number of iterations, while the performance quickly decays. With an MLP of the right capacity, four additional tricks: (1) weight augmentation, (2) random restarts with an unregularized model, (3) fixed window of data, and (4) surrogate retraining greatly improve the accuracy of the average path length predictions. 

Normally, if our differentiable model is a GRU, we compile examples using the GRU weights at every batch and calculate the true average path length. This dataset is used to train the surrogate model. If examples are very sparse, surrogate predictions may be unstable. Augmentation addresses this by randomly sampling weight vectors and computing the average path length to artificially create a larger dataset. Early epochs are especially problematic when it comes to lacking data. In addition to augmentation, we use random restarts to separately train unregularized GRUs (each with different weight initializations) to grow a dataset of weight vectors prior to training the regularized model.  

As the GRU parameters take steps away from their initial values, our examples from those early epochs no longer describe the current state of the model. Retraining and a fixed window of data address this by re-learning the surrogate function at a fixed frequency using examples only from the last $J$ epochs. In practice, both the augmentation size, the retraining frequency, and $J$ are functions of the learning rate and the dataset size. See table \ref{fig:learning_rate_for_datasets} for exact numbers.

\clearpage
\section{Experimental Protocol}

See table \ref{fig:learning_rate_for_datasets} for model hyperparameters for each dataset. For standard recurrent models such as HMM or GRU, the decision trees were trained on the input data and the predictions of the model's output node. For our deep residual GRU-HMM, the decision trees were trained on the predictions on the GRU's output node only. For both synthetic and real-world datasets, our surrogate to the tree loss is a multilayer perceptron with 1 hidden layer of 25 nodes. For each dataset, when we investigated several regularization strengths ($\lambda$), we initialize the model weights using the same random seed. We use the Adam algorithm~\cite{kingma2014adam} for all optimization.

\begin{table*}[h!]
    \centering
    \resizebox{\linewidth}{!}{%
    \begin{tabular}{ c | c c c c c c c c c c}
        Dataset & Total Num. Sequences & Avg. seq. length & Learning Rate & Batch size & Minimum Leaf Sample & Post-pruned & Epochs (Model) & Epochs (Surrogate) & Retraining Freq. & $J$ \\
        \hline 
        parabola & n/a & n/a & 1e-2 & 32 & 0 & N & 250 & 500 & 100 & n/a \\
        signal-and-noise HMM & 100 & 50 & 1e-2 & 10 & 25 & Y & 300 & 1000 & 50 & 50\\
        HIV & 53\,236 & 14 & 1e-3 & 256 & 1\,000 & Y & 300 & 5000 & 25 & 100 \\
        SEPSIS & 11\,786 & 15 & 1e-3 & 256 & 1\,000 & Y & 300 & 5000 & 25 & 100 \\
        TIMIT & 6\,303 & 614 & 1e-3 & 256 & 5\,000 & Y & 200 & 5000 & 25 & 100\\
    \end{tabular}}
    \caption{Dataset summaries and training parameters used in our experiments.}
    \label{fig:learning_rate_for_datasets}
\end{table*}

\subsection{2D Parabola}

\paragraph{Dataset generation.}
The training data consists of 2D input points whose two-class decision
boundary is roughly shaped like a parabola.  The true decision
function is defined by $y = 5*(x-0.5)^{2} + 0.4$. We sampled all 200
input points $x_n$ uniformly within the unit square $[0,1] \times
[0,1]$ and labeled those above the decision function as positive.  To
add randomness, we flipped 10\% of the points in the region near the
boundary between $y = 5*(x-0.5)^{2} + 0.2$ and $y = 5*(x-0.5)^{2} +
0.6$.

\paragraph{Regularization strengths.}
Tested values of regularization strength parameter $\lambda$:
0.1, 0.5, 1, 5, 10, 25, 50, 75, 100, 250, 500, 750, 1\,000, 2\,500, 5\,000, 7\,500, 10\,000, 25\,000, 50\,000, 75\,000, 100\,000


\subsection{Signal-and-noise HMM}

\paragraph{Dataset generation}
The transition and emission matrices describing the generative process used to create the signal-and-noise HMM are shown in Fig.~\ref{fig:toy-hmm}. The output $y_{n}$ at every timestep is created by concatenating a one-hot vector of an emitted state and the 7-dimensional binary input vector. We emphasize that to output 1, the HMM must be in state 1 and the first input feature must be 1.

\begin{figure}[!h]
    \begin{subfigure}[b]{0.49\linewidth}
        \[
            \begin{psmallmatrix}
            .5 & .5 & .5 & .5 & 0 & 0 & 0 \\
            .5 & .5 & .5 & .5 & .5 & 0 & 0 \\
            .5 & .5 & .5 & 0 & .5 & 0 & 0 \\
            .5 & .5 & .5 & 0 & 0 & .5 & 0 \\
            .5 & .5 & .5 & 0 & 0 & 0 & .5
            \end{psmallmatrix}
        \]
        \caption{}
        \label{fig:emission:hmm}
    \end{subfigure}
    \begin{subfigure}[b]{0.49\linewidth}
        \[
            \begin{psmallmatrix}
            .7 & .3 & 0 & 0 & 0 \\
            .5 & .25 & .25 & 0 & 0 \\
            0 & .25 & .5 & .25 & 0 \\
            0 & 0 & .25 & .25 & .5 \\
            0 & 0 & 0 & .5 & .5
            \end{psmallmatrix}
        \]
        \caption{}
        \label{fig:transition:hmm}
    \end{subfigure}
    \begin{subfigure}[b]{0.49\linewidth}
        \[
            \begin{psmallmatrix}
            .5 & .5 & .5 & 0 & 0 & 0 & 0 \\
            0 & .5 & .5 & .5 & 0 & 0 & 0 \\
            0 & 0 & .5 & .5 & .5 & 0 & 0 \\
            0 & 0 & 0 & .5 & .5 & .5 & 0 \\
            0 & 0 & 0 & 0 & .5 & .5 & .5
            \end{psmallmatrix}
        \]
        \caption{}
        \label{fig:emission:hmm2}
    \end{subfigure}
    \begin{subfigure}[b]{0.49\linewidth}
        \[
            \begin{psmallmatrix}
            .2 & .2 & .2 & .2 & .2 \\
            .2 & .2 & .2 & .2 & .2 \\
            .2 & .2 & .2 & .2 & .2 \\
            .2 & .2 & .2 & .2 & .2 \\
            .2 & .2 & .2 & .2 & .2
            \end{psmallmatrix}
        \]
        \caption{}
        \label{fig:transition:hmm}
    \end{subfigure}
\caption{Emission (5 states vs 7 features) and transition probabilities for the signal HMM (a, b) and noise HMM (c, d).}
  \label{fig:toy-hmm}
\end{figure}

\paragraph{Training Details.}
With synthetic datasets, we explore
(1, 5, 6, 10, 15, 20) GRU nodes, (5, 6, 20) HMM states, and GRU-HMMs with 5 HMM
states and (1, 5, 10, 15) GRU nodes.

\subsection{Sepsis}
\paragraph{Training Details.}
We explore
(1, 5, 6, 10, 11, 15, 20, 25, 26, 30, 35, 50, 51, 55, 60, 75, 100) GRU nodes,
(5, 6, 10, 11, 15, 20, 25, 26, 30, 35, 50, 51, 55, 60, 75, 100) HMM states, and
GRU-HMMs with (5, 10, 25, 50) HMM states and (1, 5, 10, 25, 50) GRU nodes. The input features are z-scored prior to training.

\subsection{HIV}

\paragraph{Training Details.} We explore
(1, 5, 6, 10, 11, 15, 20, 25, 26, 30, 35, 50, 51, 55, 60, 75) GRU nodes,
(5, 6, 10, 11, 15, 20, 25, 26, 30, 35, 50, 51, 55, 60, 75) HMM states, and
GRU-HMMs with (5, 10, 25) HMM states and (1, 5, 10, 25, 50) GRU nodes. 

\subsection{TIMIT}

\paragraph{Training Details.} We explore
(1, 5, 6, 10, 11, 15, 20, 25, 26, 30, 35, 50, 51, 55, 60, 75) GRU nodes,
(5, 6, 10, 11, 15, 20, 25, 26, 30, 35, 50, 51, 55, 60, 75) HMM states, and
GRU-HMMs with (5, 10, 25) HMM states and (1, 5, 10, 25, 50) GRU nodes. Like Sepsis, the input features are z-scored prior to training.

\section{Extended Results}

For signal-to-noise HMM, Sepsis, and TIMIT, we first show expanded versions of the fitness trace plots and the tree visualizations. For Sepsis and HIV, we show the additional output dimensions not in the paper.

We also include tables of the test AUC performance for our synthetic and real data
sets over a vast array of parameter settings (GRU node counts, HMM
state counts, regularization strengths). Consistent with the common wisdom of
training deep models, we found that larger models, with
regularization, tended to perform the best.

\subsection{Signal-and-noise HMM: Plots}

\begin{figure*}[!h]
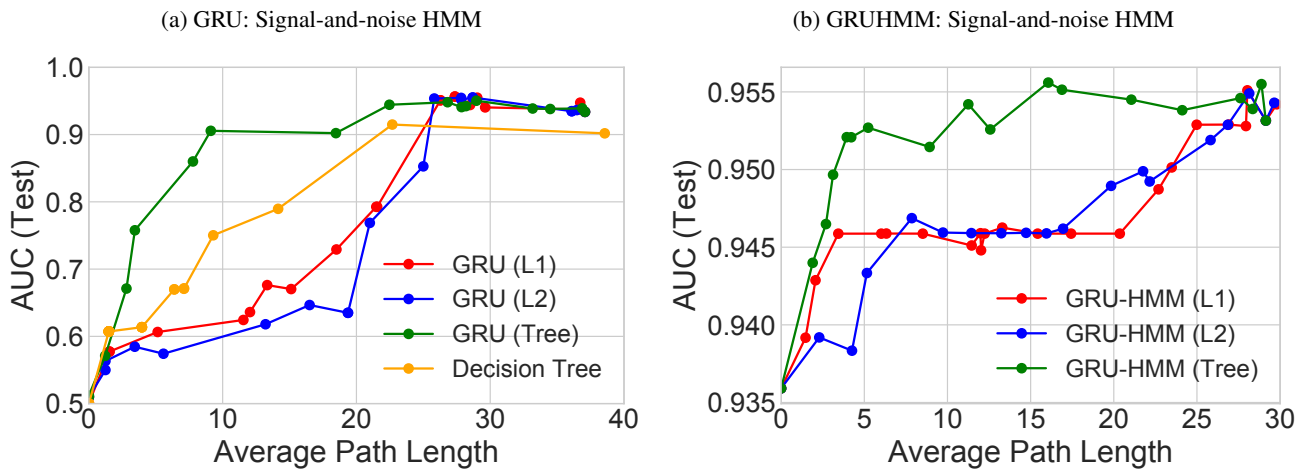

    \centering
    \begin{subfigure}[b]{0.49\linewidth}
        \caption{GRU: Signal-and-noise HMM}
        \includegraphics[width=\linewidth]{hmm2_gru.pdf}
        \label{fig:2hmm:gru:plot}
    \end{subfigure}
    \begin{subfigure}[b]{0.49\linewidth}
        \caption{GRUHMM: Signal-and-noise HMM}
        \includegraphics[width=\linewidth]{hmm2_gruhmm.pdf}
        \label{fig:2hmm:gruhmm:plot}
    \end{subfigure}
    \caption{Performance and complexity trade-offs using L1, L2, and Tree regularization on (a) GRU and (b) GRU-HMM performance on the Signal-and-noise HMM dataset. Note the differences in scale.}
\end{figure*}

\newpage
\subsection{Signal-and-noise HMM: Tree Visualization}

\begin{figure}[!h]
    \centering
    \begin{subfigure}[b]{0.19\linewidth}
        \includegraphics[width=\linewidth]{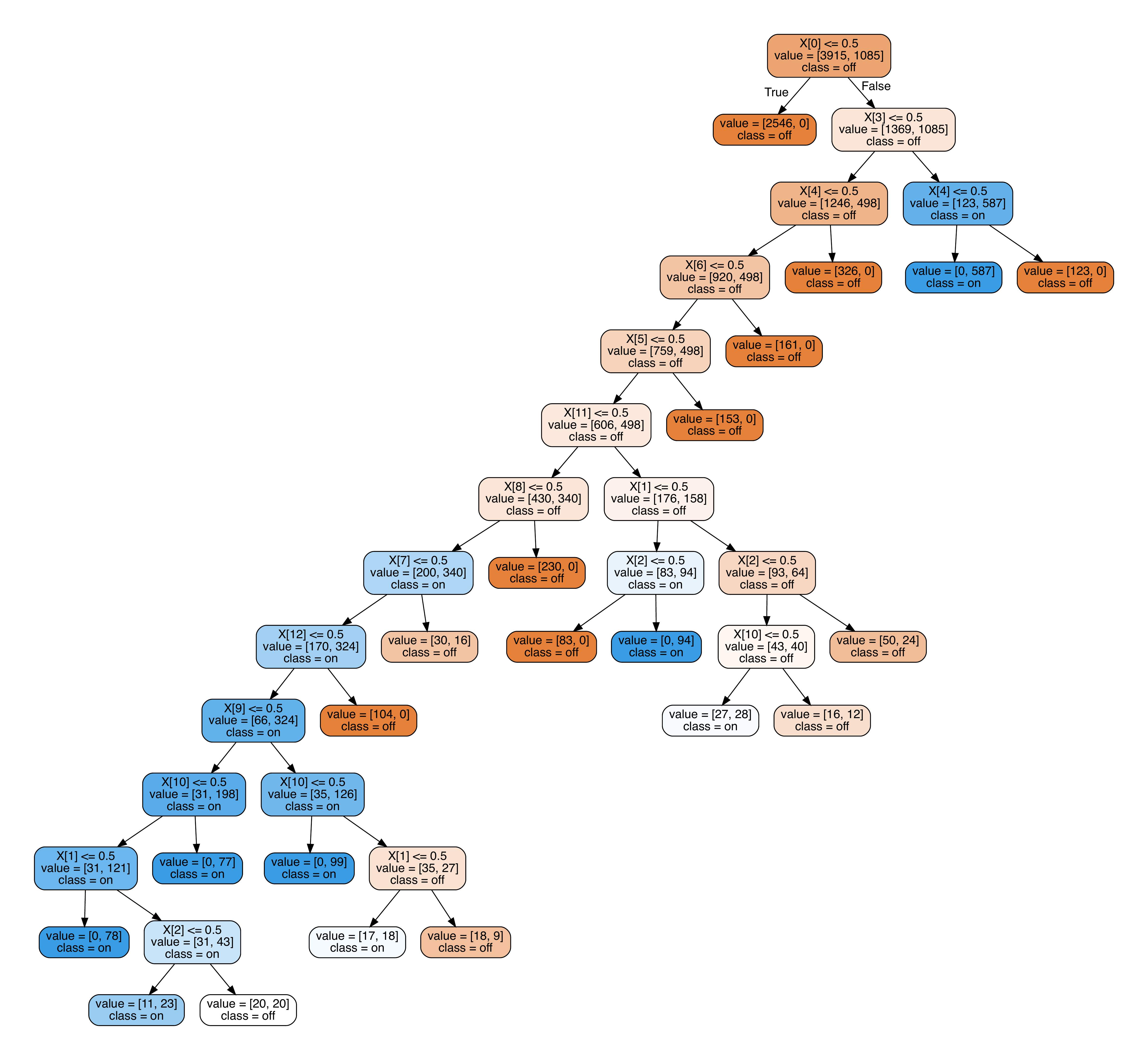}
        \label{fig:hmm2:tree:0.1}
        \caption{GRU:0.1}
    \end{subfigure}
    \begin{subfigure}[b]{0.19\linewidth}
        \includegraphics[width=\linewidth]{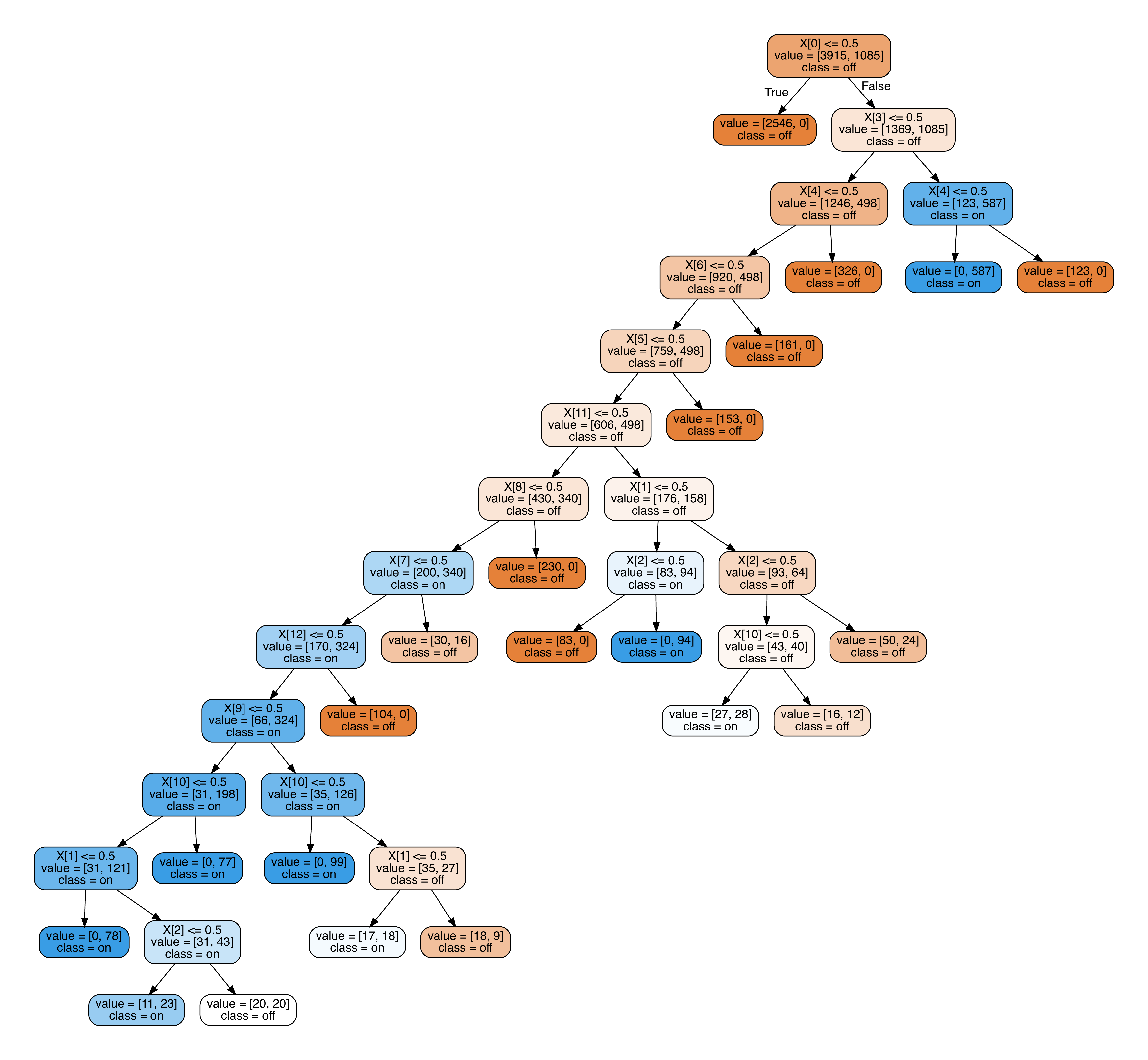}
        \label{fig:hmm2:tree:0.1}
        \caption{GRU:0.1}
    \end{subfigure}
    \begin{subfigure}[b]{0.19\linewidth}
        \includegraphics[width=\linewidth]{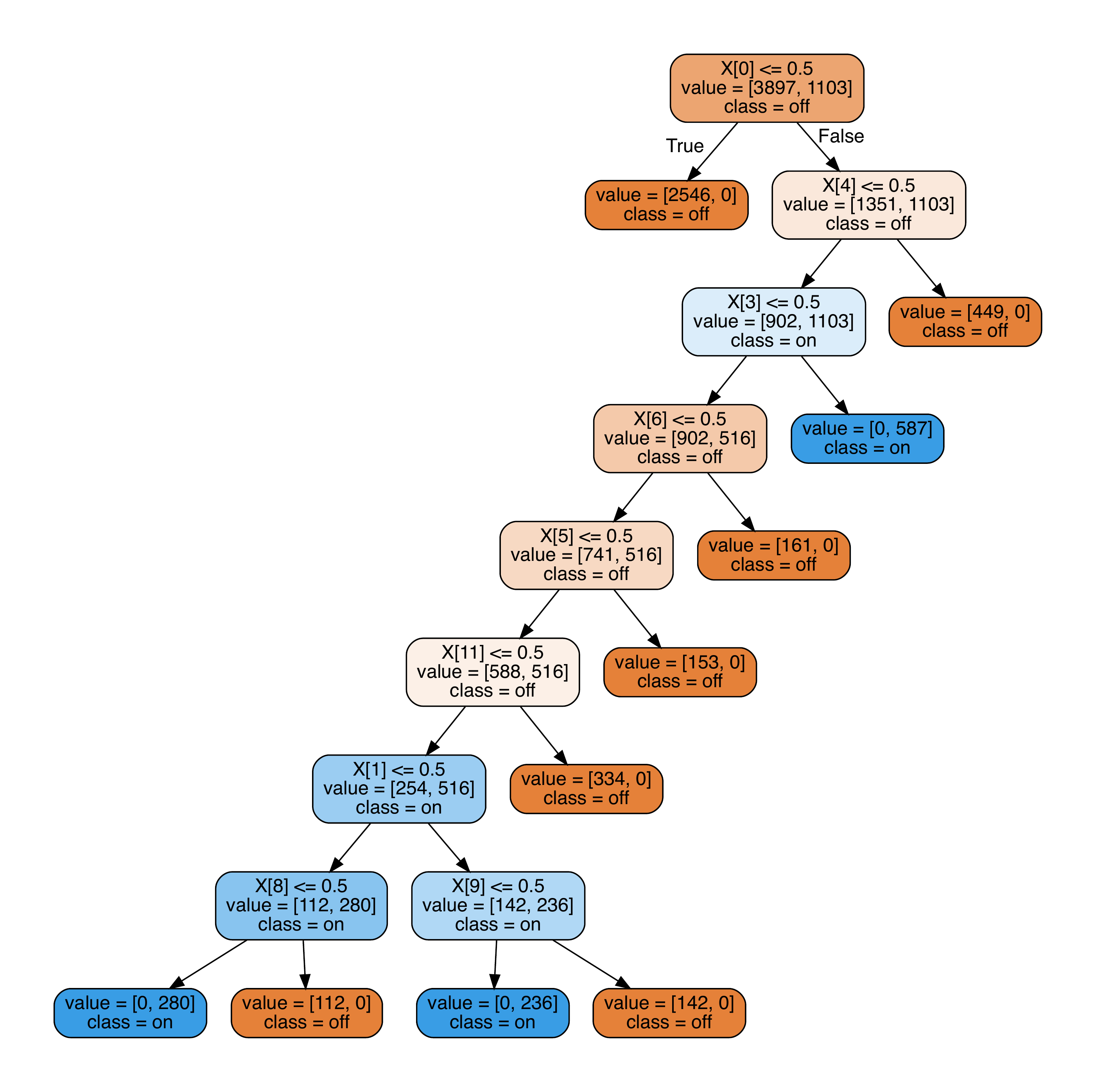}
        \label{fig:hmm2:tree:1}
        \caption{GRU:1.0}
    \end{subfigure}
    \begin{subfigure}[b]{0.19\linewidth}
        \includegraphics[width=\linewidth]{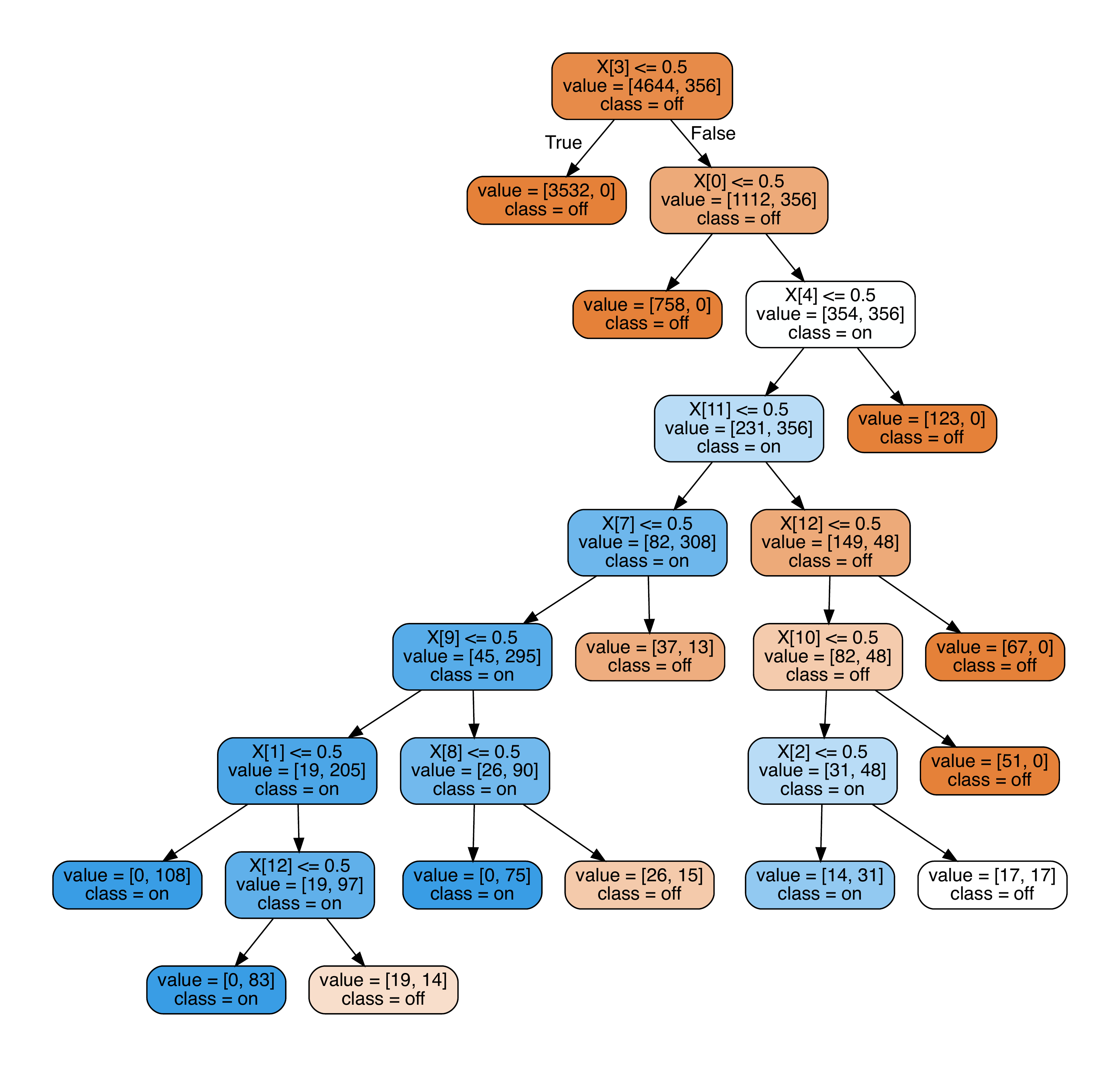}
        \label{fig:hmm2:tree:10}
        \caption{GRU:10}
    \end{subfigure}
    \begin{subfigure}[b]{0.19\linewidth}
        \includegraphics[width=\linewidth]{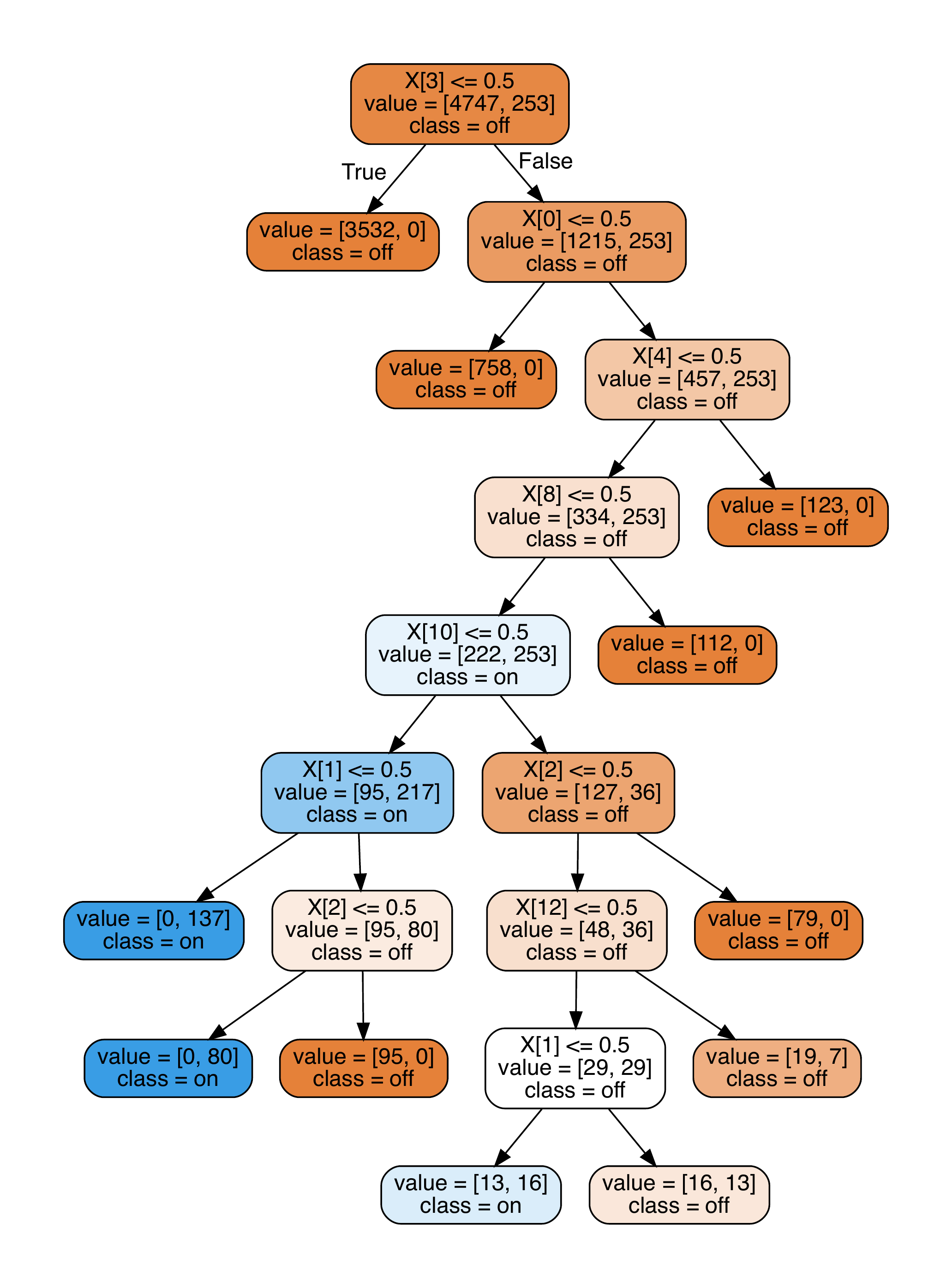}
        \label{fig:hmm2:tree:20}
        \caption{GRU:20}
    \end{subfigure}
    \begin{subfigure}[b]{0.19\linewidth}
        \includegraphics[width=\linewidth]{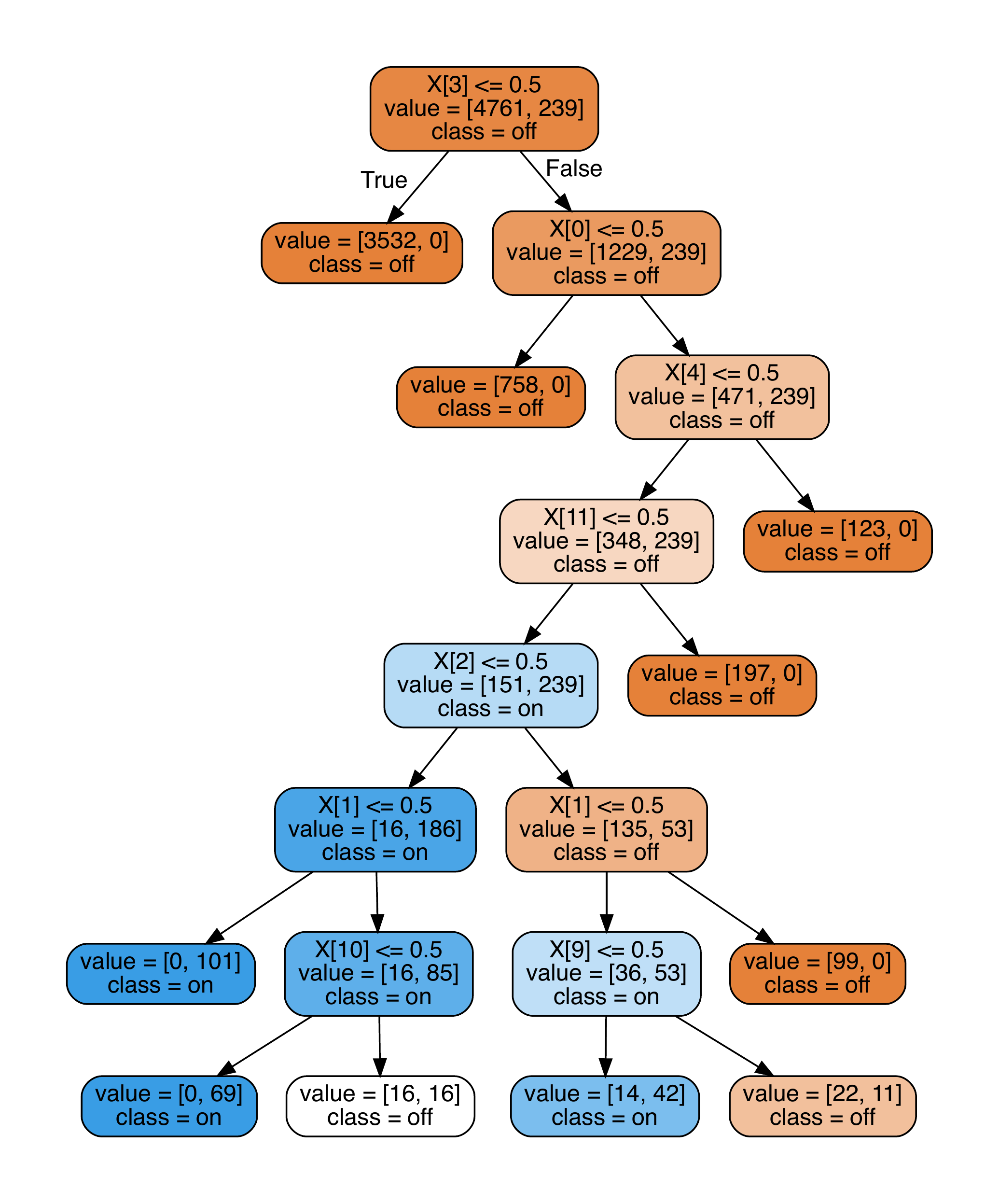}
        \label{fig:hmm2:tree:100}
        \caption{GRU:100}
    \end{subfigure}
    \begin{subfigure}[b]{0.19\linewidth}
        \includegraphics[width=\linewidth]{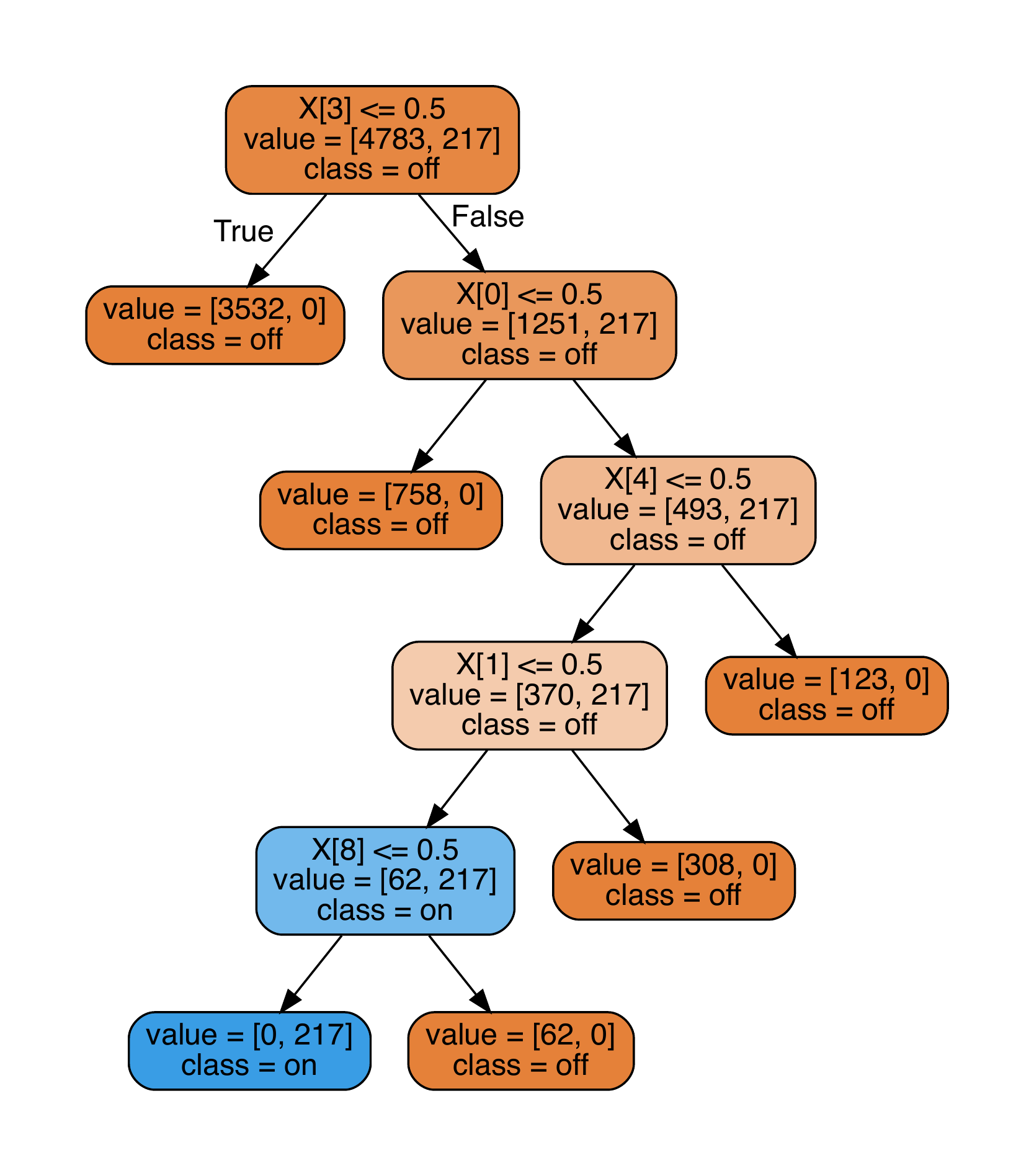}
        \label{fig:hmm2:tree:400}
        \caption{GRU::400}
    \end{subfigure}
    \begin{subfigure}[b]{0.19\linewidth}
        \includegraphics[width=\linewidth]{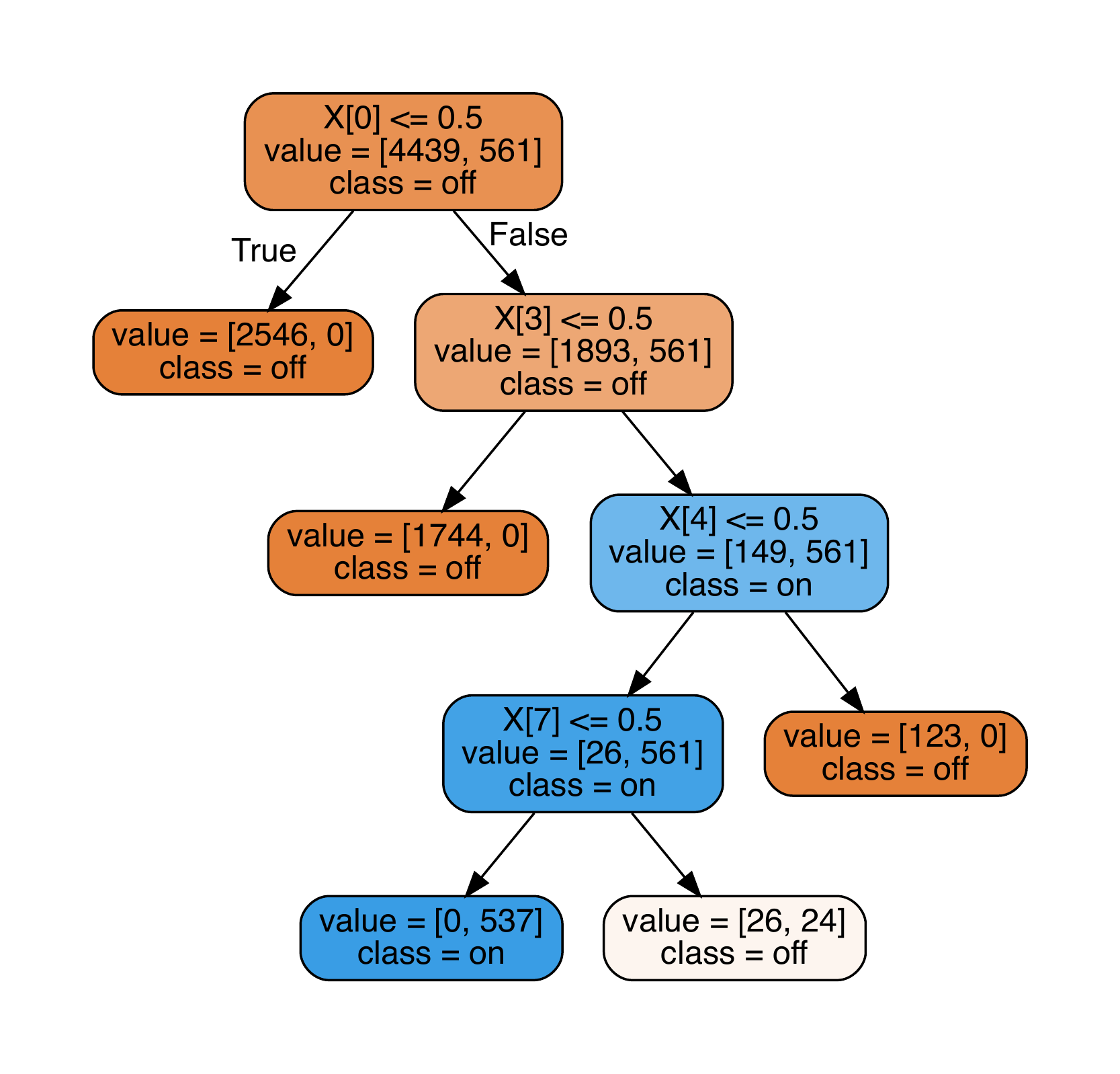}
        \label{fig:hmm2:tree:800}
        \caption{GRU:800}
    \end{subfigure}
    \begin{subfigure}[b]{0.19\linewidth}
        \includegraphics[width=\linewidth]{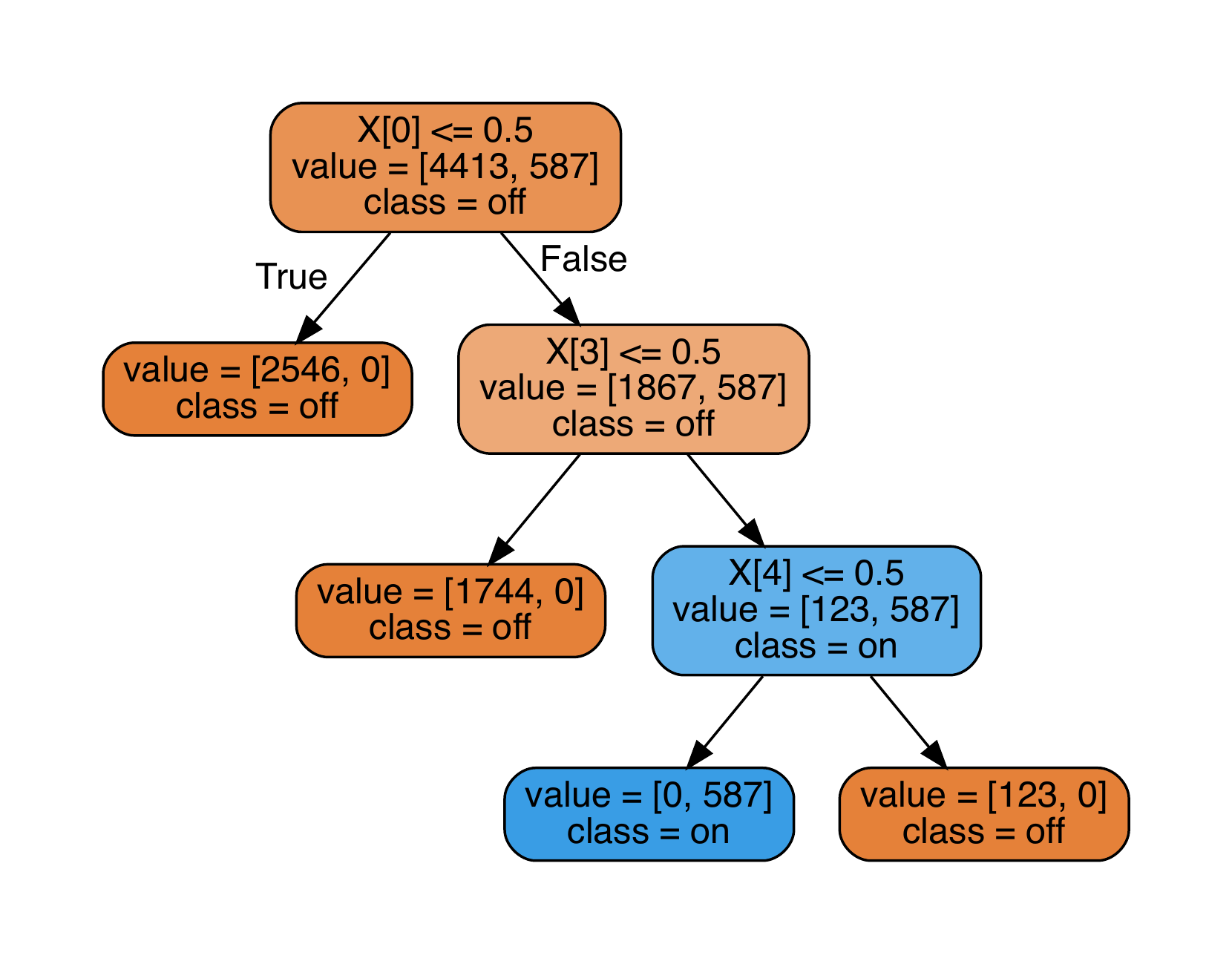}
        \label{fig:hmm2:tree:1000}
        \caption{GRU:1\,000}
    \end{subfigure}
    \begin{subfigure}[b]{0.19\linewidth}
        \includegraphics[width=\linewidth]{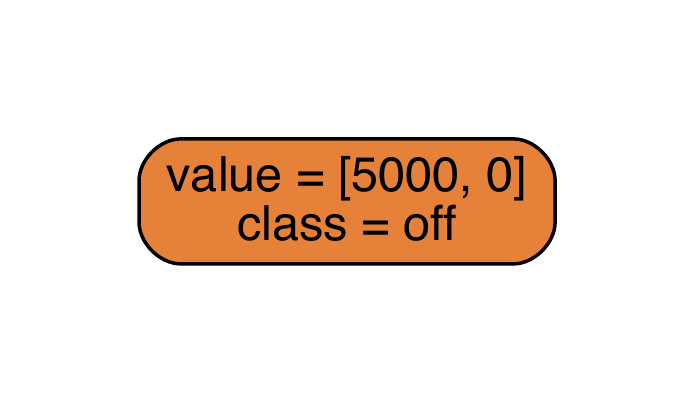}
        \label{fig:hmm2:tree:10000}
        \caption{GRU:10\,000}
    \end{subfigure}
    \caption{Decision trees trained under varying tree regularization strengths for GRU models on the signal-and-noise HMM dataset dataset.
    As the tree regularization increases, the number of nodes collapses to a single
    one. If we focus on (h), we see that the tree resembles the ground truth
    data-generating function quite closely.}
\end{figure}

\newpage
\subsection{Signal-and-noise HMM: AUCs}

\begin{table}[h!]
    \centering
    \begin{minipage}{1.0\textwidth}
      \centering
      \resizebox{0.85\linewidth}{!}{
      \begin{tabular}{r | c c c c}
        Model & AUC (Test) & Average Path Length & Parameter Count \\ [0.5ex]
        \hline
        logreg & 0.91832 & 17.302 & 6\\
        \hline
        decision tree & 0.92050 & 29.4424 & - \\
        \hline
        hmm (5) & 0.93591 & 25.5736 & 71\\
        hmm (20) & 0.94177 & 27.2784 & 581\\
        \hline
        gru (1) & 0.65049 & 1.8876 & 29\\
        gru (5) & 0.94812 & 26.304 & 205\\
        gru (6) & 0.94883 & 27.2118 & 264\\
        gru (10) & 0.94962 & 28.563 & 560\\
        gru (15) & 0.93982 & 30.7172 & 1\,065\\
        gru (20) & 0.93368 & 37.0844 & 1\,720\\
        \hline
        grutree (20/10.0) & 0.94226 & 28.1850 & 1\,720\\
        grutree (20/200.0) & 0.94806 & 26.8140 & 1\,720\\
        grutree (20/7\,000.0) & 0.94431 & 22.4646 & 1\,720\\
        grutree (20/9\,000.0) & 0.90555 & 9.1127 & 1\,720\\
        grutree (20/10\,000.0) & 0.82770 & 3.4400 & 1\,720\\
        \hline
        gruhmm (5/1) & 0.95146 & 18.2202 & 100\\
        gruhmm (5/5) & 0.95584 & 27.258 & 276\\
        gruhmm (5/10) & 0.95773 & 30.9624 & 631\\
        gruhmm (5/15) & 0.94857 & 36.7188 & 1\,136\\
        \hline
        gruhmmtree (5/15/1.0) & 0.95382 & 24.115 & 1\,136\\
        gruhmmtree (5/15/10.0) & 0.95180 & 16.883 & 1\,136\\
        gruhmmtree (5/15/50.0) & 0.95258 & 12.573 & 1\,136\\
        gruhmmtree (5/15/200.0) & 0.95145 & 8.926 & 1\,136\\
        gruhmmtree (5/15/500.0) & 0.95769 & 5.231 & 1\,136\\
        gruhmmtree (5/15/900.0) & 0.95708 & 3.942 & 1\,136\\
        gruhmmtree (5/15/2\,000.0) & 0.95648 & 2.694 & 1\,136\\
        gruhmmtree (5/15/5\,000.0) & 0.95399 & 1.896 & 1\,136\\
        gruhmmtree (5/15/7\,000.0) & 0.93591 & 0.000 & 1\,136\\
      \end{tabular}}
    \end{minipage}
    \caption{Performance metrics across models on the signal-and-noise HMM dataset. The parameter count is included as a measure of the model capacity.}
    \label{table:synthetic-results}
\end{table}

\newpage
\subsection{Sepsis: Plots}

\begin{figure*}[!h]
    \centering
    \begin{subfigure}[b]{0.32\linewidth}
        \caption{In-Hospital Mortality}
        \includegraphics[width=\linewidth]{sepsis_grutree_0.pdf}
        \label{fig:sepsis:grutree:0}
    \end{subfigure}
    \begin{subfigure}[b]{0.32\linewidth}
        \caption{90-Day Mortality}
        \includegraphics[width=\linewidth]{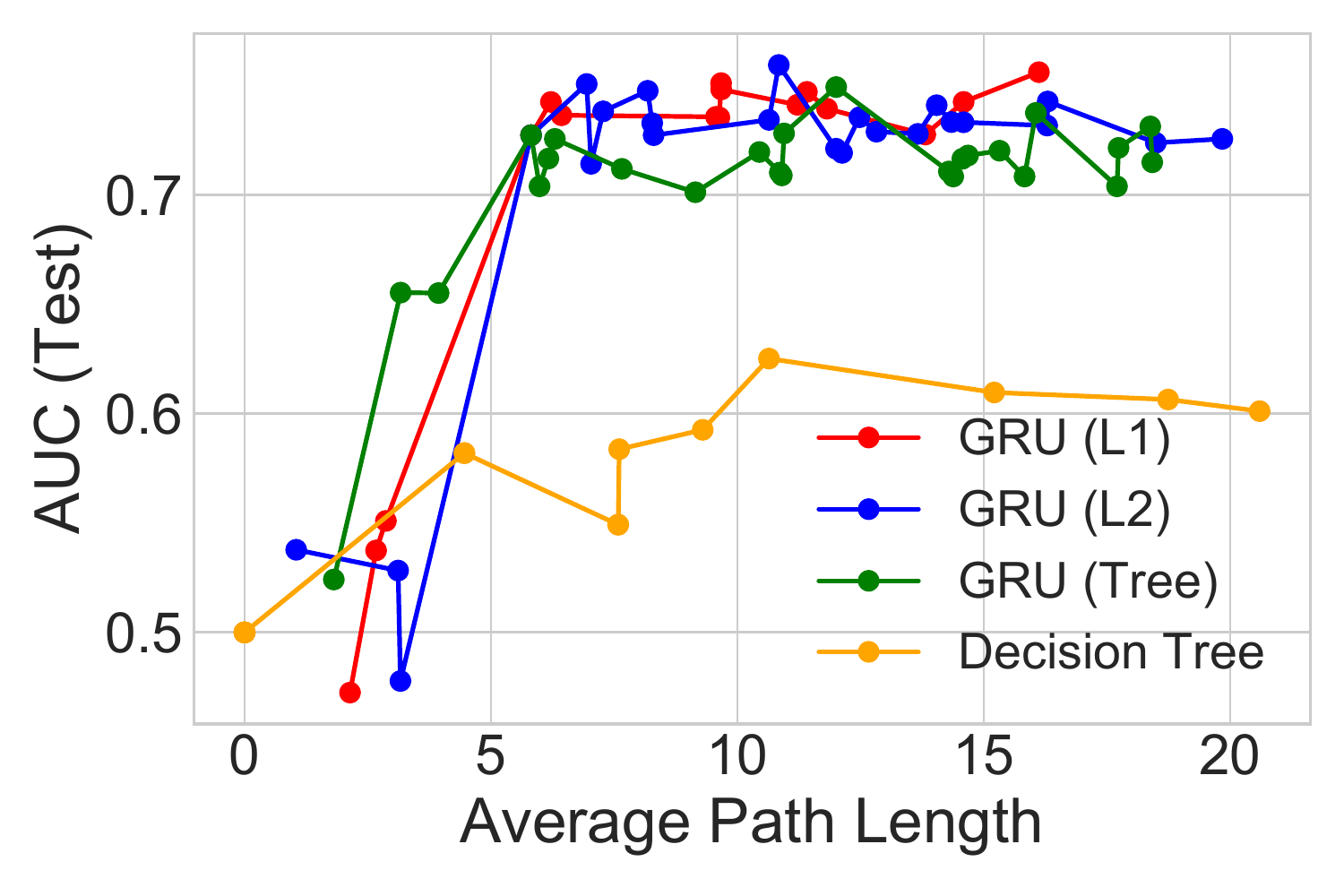}
        \label{fig:sepsis:grutree:1}
    \end{subfigure}
    \begin{subfigure}[b]{0.32\linewidth}
        \caption{Mechanical Ventilation}
        \includegraphics[width=\linewidth]{sepsis_grutree_2.pdf}
        \label{fig:sepsis:grutree:2}
    \end{subfigure}
    \begin{subfigure}[b]{0.32\linewidth}
        \caption{Median Vasopressor}
        \includegraphics[width=\linewidth]{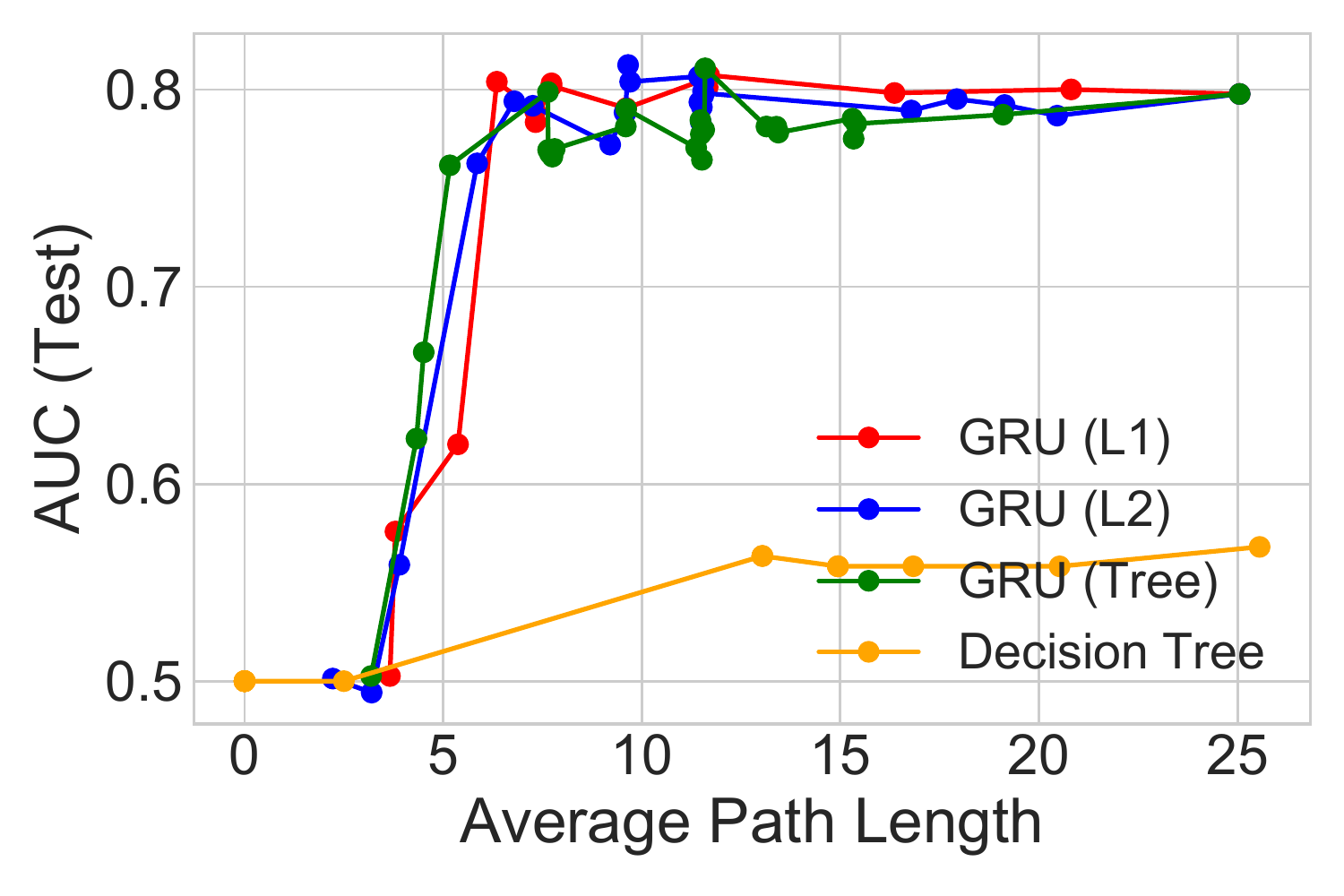}
        \label{fig:sepsis:grutree:3}
    \end{subfigure}
    \begin{subfigure}[b]{0.32\linewidth}
        \caption{Max Vasopressor}
        \includegraphics[width=\linewidth]{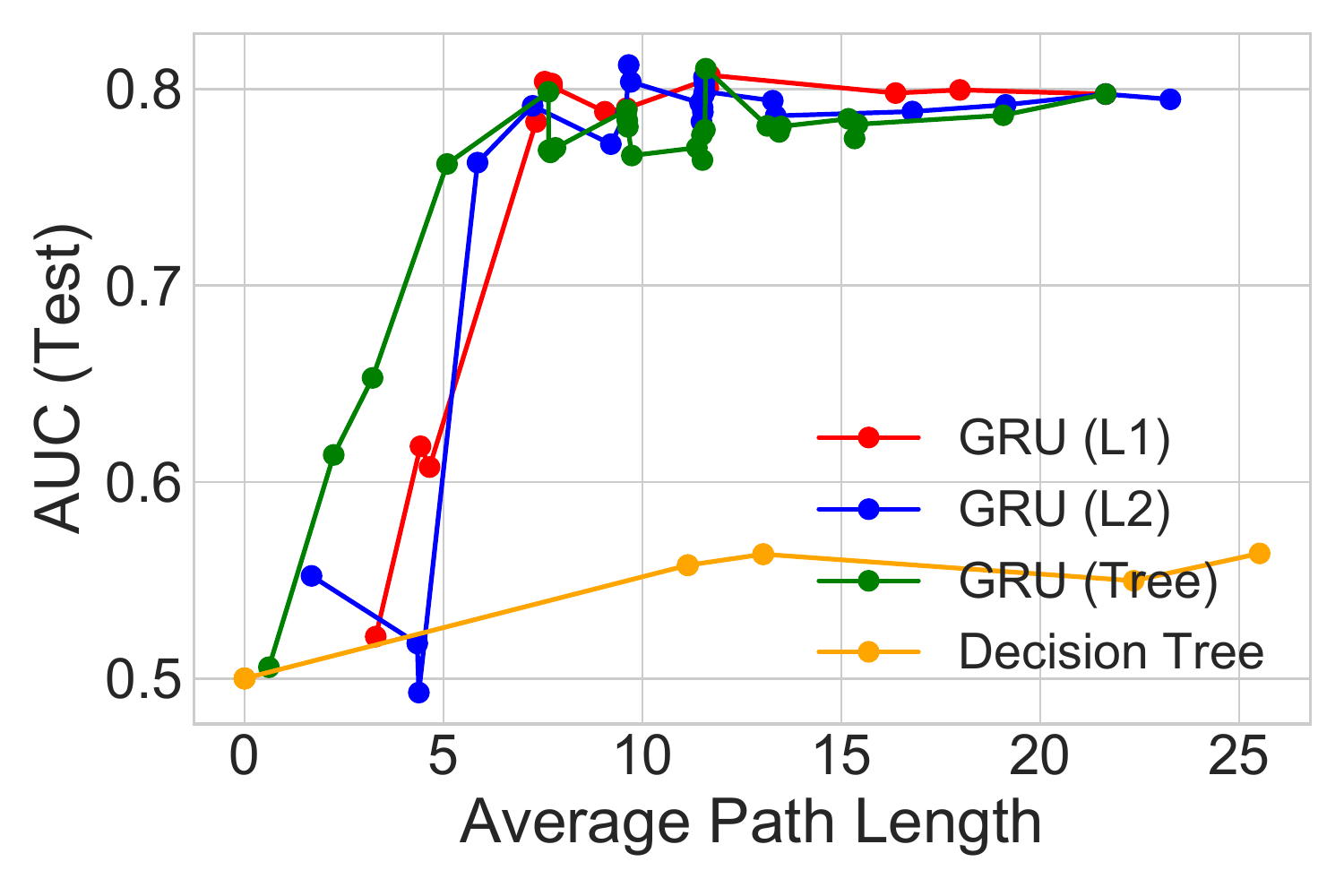}
        \label{fig:sepsis:grutree:4}
    \end{subfigure}
    \caption{Performance and complexity trade-offs using L1, L2, and Tree
    regularization on GRU performance on the Sepsis dataset.}
    \label{fig:sepsis:grutree:all}
\end{figure*}

\subsection{Sepsis: Tree Visualization}
\begin{figure*}[!ht]
    \centering
    \begin{subfigure}[b]{0.19\linewidth}
        \includegraphics[width=\linewidth]{sparse_trees/figurec4a}
        \label{fig:sepsis:gru:tree:dim:1}
        \caption{In-Hospital Mortality}
    \end{subfigure}
    \begin{subfigure}[b]{0.19\linewidth}
        \includegraphics[width=\linewidth]{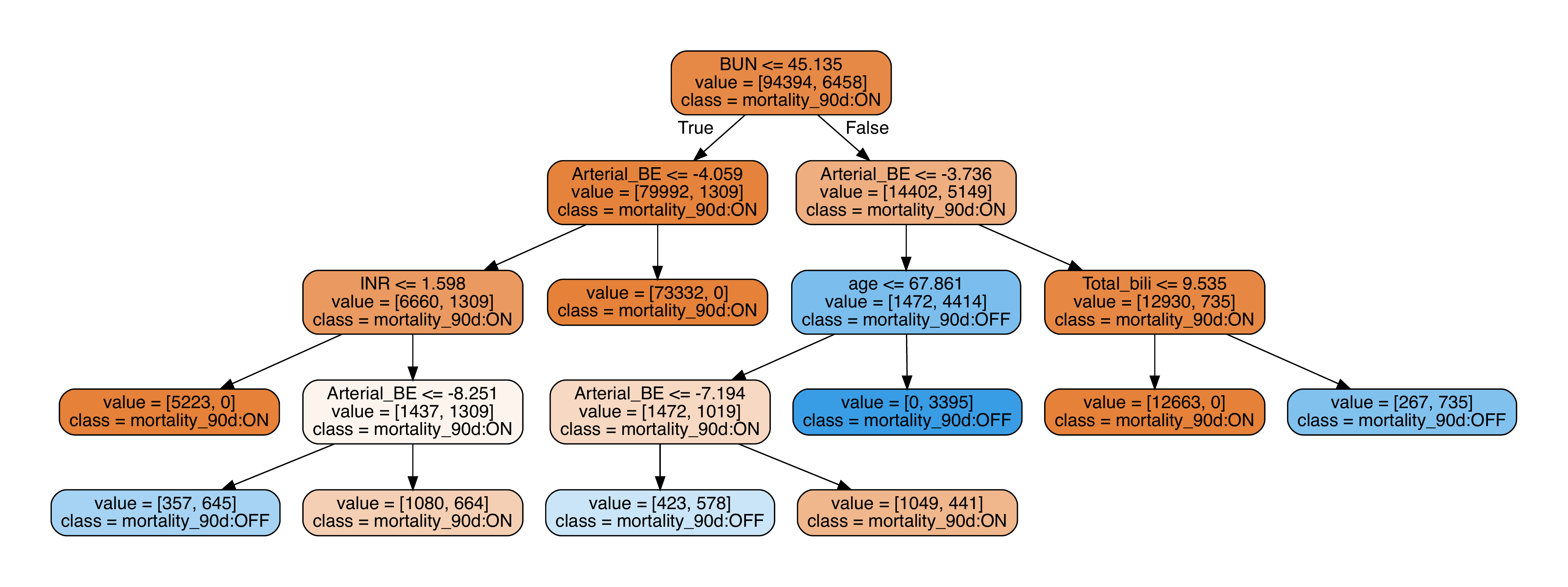}
        \label{fig:sepsis:gru:tree:dim:2}
        \caption{90-Day Mortality}
    \end{subfigure}
    \begin{subfigure}[b]{0.19\linewidth}
        \includegraphics[width=\linewidth]{sparse_trees/figurec4c}
        \label{fig:sepsis:gru:tree:dim:3}
        \caption{Mechanical Ventilation}
    \end{subfigure}
    \begin{subfigure}[b]{0.19\linewidth}
        \includegraphics[width=\linewidth]{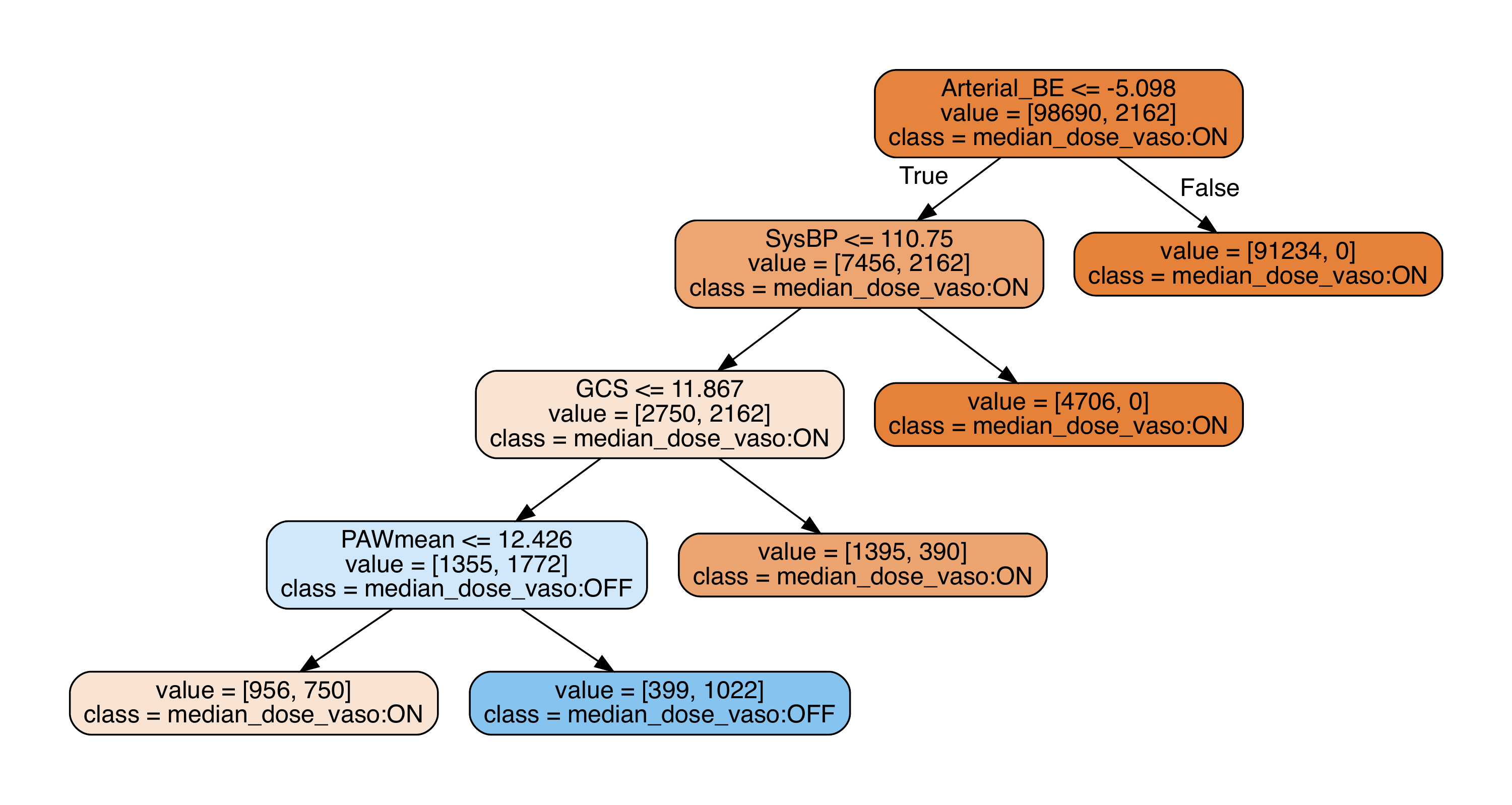}
        \label{fig:sepsis:gru:tree:dim:4}
        \caption{Median Vasopressor}
    \end{subfigure}
    \begin{subfigure}[b]{0.19\linewidth}
        \includegraphics[width=\linewidth]{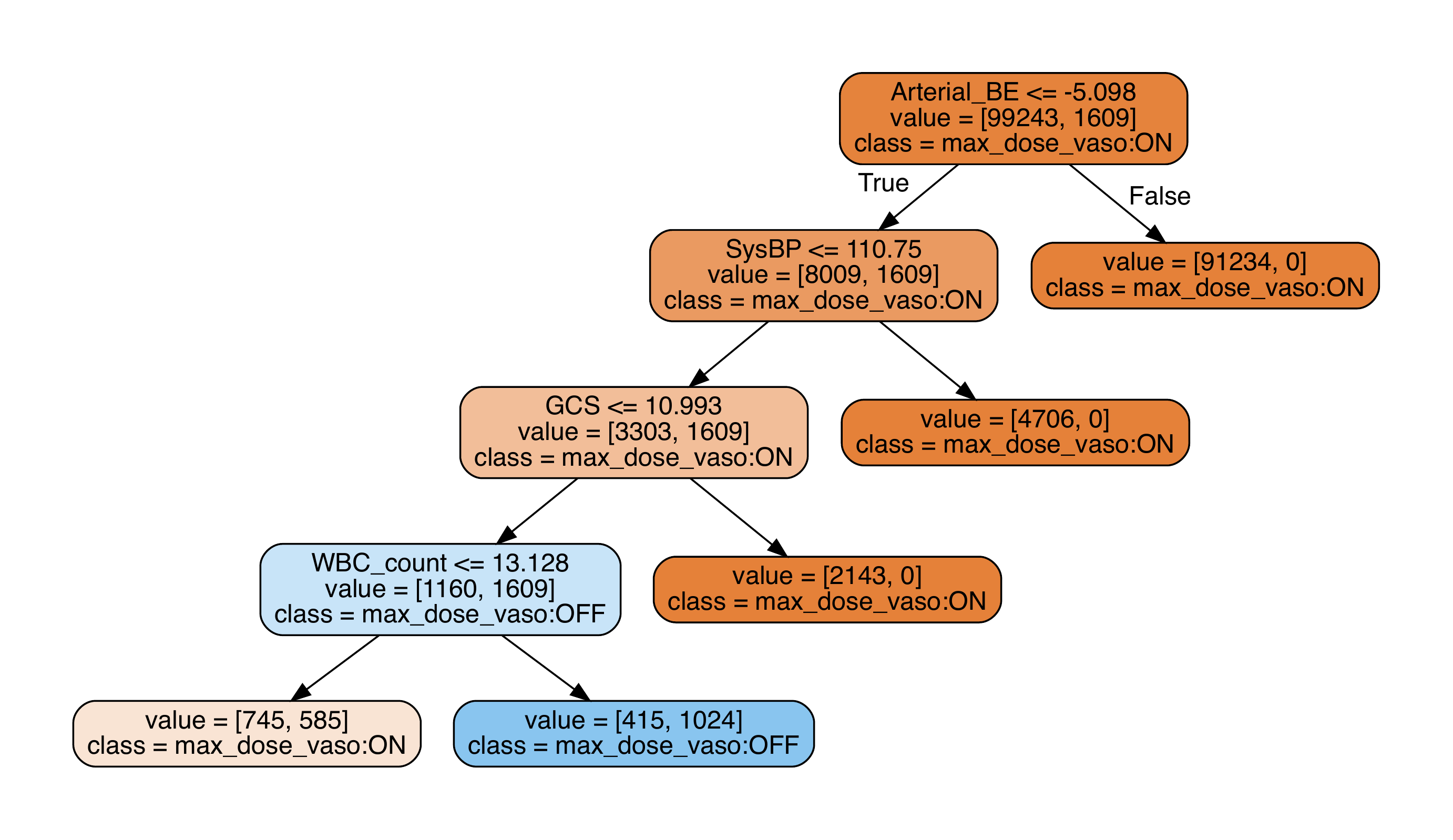}
        \label{fig:sepsis:gru:tree:dim:5}
        \caption{Max Vasopressor}
    \end{subfigure}
    \caption{Decision trees trained using $\lambda=800.0$ for a GRU model using Sepsis. The 5 output dimensions are jointly trained.}
    \label{table:sepsis-gru-tree}
\end{figure*}

\newpage
\subsection{Sepsis: AUCs}
\begin{table*}[ht!]
    \centering
    \resizebox{0.8\linewidth}{!}{%
    \begin{tabular}{r | l l l l l r r}
        Model & \shortstack{In-Hospital \\ Mortality} & \shortstack{90-Day \\ Mortality} & \shortstack{Mechanical \\ Ventilation} & \shortstack{Median \\ Vasopressor} & \shortstack{Max \\ Vasopressor} & \shortstack{Total Average \\ Path Length} & \shortstack{Parameter \\ Count}\\ [0.5ex]
        \hline
        logreg & 0.6980 & 0.6986 & 0.8242 & 0.7392 & 0.7392 & 32.489 & 180\\
        \hline
        decision tree & 0.7017 & 0.7016 & 0.8509 & 0.7439 & 0.7427 & 76.242 & - \\
        \hline
        hmm (5) & 0.7128 & 0.7095 & 0.6979 & 0.7295 & 0.7290 & 35.125 & 405\\
        hmm (10) & 0.7227 & 0.7297 & 0.8237 & 0.7409 & 0.7405 & 57.629 & 860\\
        hmm (15) & 0.7216 & 0.7282 & 0.8188 & 0.7346 & 0.7341 & 61.832 & 1\,365\\
        hmm (20) & 0.7233 & 0.7350 & 0.8218 & 0.7371 & 0.7364 & 62.353 & 1\,920\\
        hmm (25) & 0.7147 & 0.7321 & 0.8089 & 0.7313 & 0.7310 & 63.415 & 2\,525\\
        hmm (30) & 0.7164 & 0.7297& 0.8099 & 0.7316 & 0.7311 & 65.164 & 3\,180\\
        hmm (35) & 0.7177 & 0.7237 & 0.8095 & 0.7201 & 0.7195 & 65.474 & 3\,885\\
        hmm (50) & 0.7267 & 0.7357 & 0.8373 & 0.7335 & 0.7328 & 66.317 & 6\,300\\
        hmm (75) & 0.7254 & 0.7361 & 0.8059 & 0.7434 & 0.7430 & 72.553 & 11\,325\\
        hmm (100) & 0.7294 & 0.7354 & 0.8129 & 0.7408 & 0.7403 & 80.415 & 17\,600\\
        \hline
        gru (1) & 0.3897 & 0.6400 & 0.4761 & 0.7414 & 0.7411 & 31.816 & 117\\
        gru (5) & 0.7357 & 0.7296 & 0.8795 & 0.7866 & 0.7862 & 45.395 & 645\\
        gru (10) & 0.7488 & 0.7445 & 0.8892 & 0.7983 & 0.7979 & 58.102 & 1\,440\\
        gru (15) & 0.7529 & 0.7450 & 0.8912 & 0.8020 & 0.8021 & 61.025 & 2\,385\\
        gru (20) & 0.7535 & 0.7497 & 0.8887 & 0.8018 & 0.8017 & 61.214 & 3\,480\\
        gru (25) & 0.7578 & 0.7486 & 0.8902 & 0.8113 & 0.8114 & 62.029 & 4\,725\\
        gru (30) & 0.7602 & 0.7508 & 0.8927 & 0.8063 & 0.8061 & 72.854 & 6\,120\\
        gru (35) & 0.7522 & 0.7483 & 0.8900 & 0.8095 & 0.8091 & 74.091 & 7\,665\\
        gru (50) & 0.7431 & 0.7390 & 0.8895 & 0.8054 & 0.8051 & 76.543 & 13\,200\\
        gru (75) & 0.7408 & 0.7239 & 0.8837 & 0.8006 & 0.8000 & 87.422 & 25\,425\\
        gru (100) & 0.7325 & 0.7273 & 0.8781 & 0.7977 & 0.7975 & 94.161 & 41\,400\\
        \hline
        grutree (100/0.01) & 0.7276 & 0.7314 & 0.8776 & 0.7873 & 0.7867 & 91.797 & 41\,400 \\
        grutree (100/1.0) & 0.7147 & 0.7040 & 0.8741 & 0.7812 & 0.7810 & 82.019 & 41\,400 \\
        grutree (100/8.0) & 0.7232 & 0.7203 & 0.8763 & 0.7845 & 0.7840 & 73.767 & 41\,400 \\
        grutree (100/20.0) & 0.7123 & 0.7085 & 0.8733 & 0.7813 & 0.7813 & 65.035 & 41\,400 \\
        grutree (100/70.0) & 0.7360 & 0.7376 & 0.8813 & 0.7988 & 0.7986 & 61.012 & 41\,400 \\
        grutree (100/300.0) & 0.7210 & 0.7197 & 0.8681 & 0.7676 & 0.7678 & 54.177 & 41\,400 \\
        grutree (100/2\,000.0) & 0.7230 & 0.7167 & 0.8335 & 0.7616 & 0.7619 & 48.206 & 41\,400 \\
        grutree (100/5\,000.0) & 0.6546 & 0.6552 & 0.6752 & 0.6668 & 0.6530 & 26.085 & 41\,400 \\
        grutree (100/7\,000.0) & 0.6063 & 0.6554 & 0.6565 & 0.6230 & 0.6138 & 20.214 & 41\,400 \\
        grutree (100/8\,000.0) & 0.5298 & 0.5242 & 0.5025 & 0.5026 & 0.5057 & 13.383 & 41\,400 \\
        \hline
        gruhmm (1/5) & 0.4222 & 0.6472 & 0.4678 & 0.7478 & 0.7477 & 41.583 & 722\\
        gruhmm (1/10) & 0.4007 & 0.6295 & 0.4730 & 0.7418 & 0.7419 & 61.041 & 1\,517\\
        gruhmm (1/25) & 0.4019 & 0.6207 & 0.4773 & 0.7353 & 0.7352 & 65.955 & 4\,802\\
        gruhmm (1/50) & 0.3999 & 0.6162 & 0.4772 & 0.7120 & 0.7121 & 70.534 & 13\,277\\
        gruhmm (5/5) & 0.7430 & 0.7372 & 0.8798 & 0.8009 & 0.8006 & 47.639 & 1\,050\\
        gruhmm (5/10) & 0.7408 & 0.7320 & 0.8819 & 0.7991 & 0.7988 & 63.627 & 1\,845\\
        gruhmm (5/25) & 0.7365 & 0.7279 & 0.8776 & 0.7955 & 0.7952 & 68.215 & 5\,130\\
        gruhmm (5/50) & 0.7222 & 0.7107 & 0.8660 & 0.7814 & 0.7811 & 71.572 & 13\,605\\
        gruhmm (10/5) & 0.7468 & 0.7467 & 0.8949 & 0.8098 & 0.8097 & 50.902 & 1\,505\\
        gruhmm (10/10) & 0.7490 & 0.7478 & 0.8958 & 0.8098 & 0.8096 & 63.522 & 2\,300\\
        gruhmm (10/25) & 0.7422 & 0.7407 & 0.8916 & 0.8055 & 0.8054 & 70.919 & 5\,585\\
        gruhmm (10/50) & 0.7254 & 0.7221 & 0.8824 & 0.7903 & 0.7903 & 71.297 & 14\,060\\
        gruhmm (25/5) & 0.7580 & 0.7568 & 0.8941 & 0.8236 & 0.8235 & 51.794 & 3\,170\\
        gruhmm (25/10) & 0.7592 & 0.7563 & 0.8945 & 0.8225 & 0.8225 & 64.223 & 3\,965\\
        gruhmm (25/25) & 0.7525 & 0.7508 & 0.8912 & 0.8186 & 0.8184 & 72.480 & 7\,250\\
        gruhmm (25/50) & 0.7604 & 0.7583 & 0.8954 & 0.8106 & 0.8103 & 79.127 & 11\,025 \\
        gruhmm (50/5) & 0.7655 & 0.7592 & 0.9006 & 0.8228 & 0.8226 & 64.229 & 6\,945\\
        gruhmm (50/10) & 0.7648 & 0.7568 & 0.9003 & 0.8220 & 0.8219 & 69.281 & 7\,740\\
        gruhmm (50/25) & 0.7600 & 0.7555 & 0.8981 & 0.8205 & 0.8203 & 85.503 & 11\,025\\
        gruhmm (50/50) & 0.7412 & 0.7373 & 0.8910 & 0.8056 & 0.8055 & 101.637 & 19\,500\\
        \hline
        gruhmmtree (50/50/0.5) & 0.7432 & 0.7492 & 0.879 & 0.7854 & 0.7849 & 84.188 & 19\,500 \\
        gruhmmtree (50/50/20.0) & 0.7435 & 0.747 & 0.8826 & 0.7914 & 0.7906 & 77.815 & 19\,500 \\
        gruhmmtree (50/50/50.0) & 0.7384 & 0.7548 & 0.8914 & 0.7922 & 0.7918 & 71.719 & 19\,500 \\
        gruhmmtree (50/50/200.0 & 0.747 & 0.7502 & 0.8767 & 0.7832 & 0.7824 & 69.715 & 19\,500 \\
        gruhmmtree (50/50/300.0) & 0.7539 & 0.7623 & 0.8942 & 0.8092 & 0.8091 & 66.9 & 19\,500 \\
        gruhmmtree (50/50/600.0 & 0.7435 & 0.7453 & 0.8821 & 0.7909 & 0.7905 & 63.703 & 19\,500 \\
        gruhmmtree (50/50/1\,000.0) & 0.7575 & 0.7502 & 0.8739 & 0.7882 & 0.7873 & 60.949 & 19\,500 \\
        gruhmmtree (50/50/3\,000.0) & 0.7396 & 0.7484 & 0.8926 & 0.8013 & 0.8011 & 54.751 & 19\,500 \\
        gruhmmtree (50/50/4\,000.0) & 0.7432 & 0.7511 & 0.8915 & 0.802 & 0.8024 & 44.868 & 19\,500 \\
        gruhmmtree (50/50/7\,000.0) & 0.7308 & 0.7477 & 0.8813 & 0.7881 & 0.7882 & 27.836 & 19\,500 \\
        gruhmmtree (50/50/9\,000.0) & 0.7132 & 0.7319 & 0.8261 & 0.7301 & 0.7299 & 0.0 & 19\,500 \\
    \end{tabular}}
    \caption{Performance metrics for multi-dimensional classification on a held-out portion of the Sepsis dataset. \emph{Total Average Path Length} refers to the summed average path lengths across the 5 output dimensions. Refer to Fig.~ \ref{fig:sepsis:grutree:all} for average-path-lengths split across dimensions.}
    \label{table:sepsis-results}
\end{table*}
\newpage
\subsection{HIV:Plots}
\begin{figure*}[!h]
    \centering
    \begin{subfigure}[b]{0.32\linewidth}
        \caption{Therapy Success}
        \includegraphics[width=\linewidth]{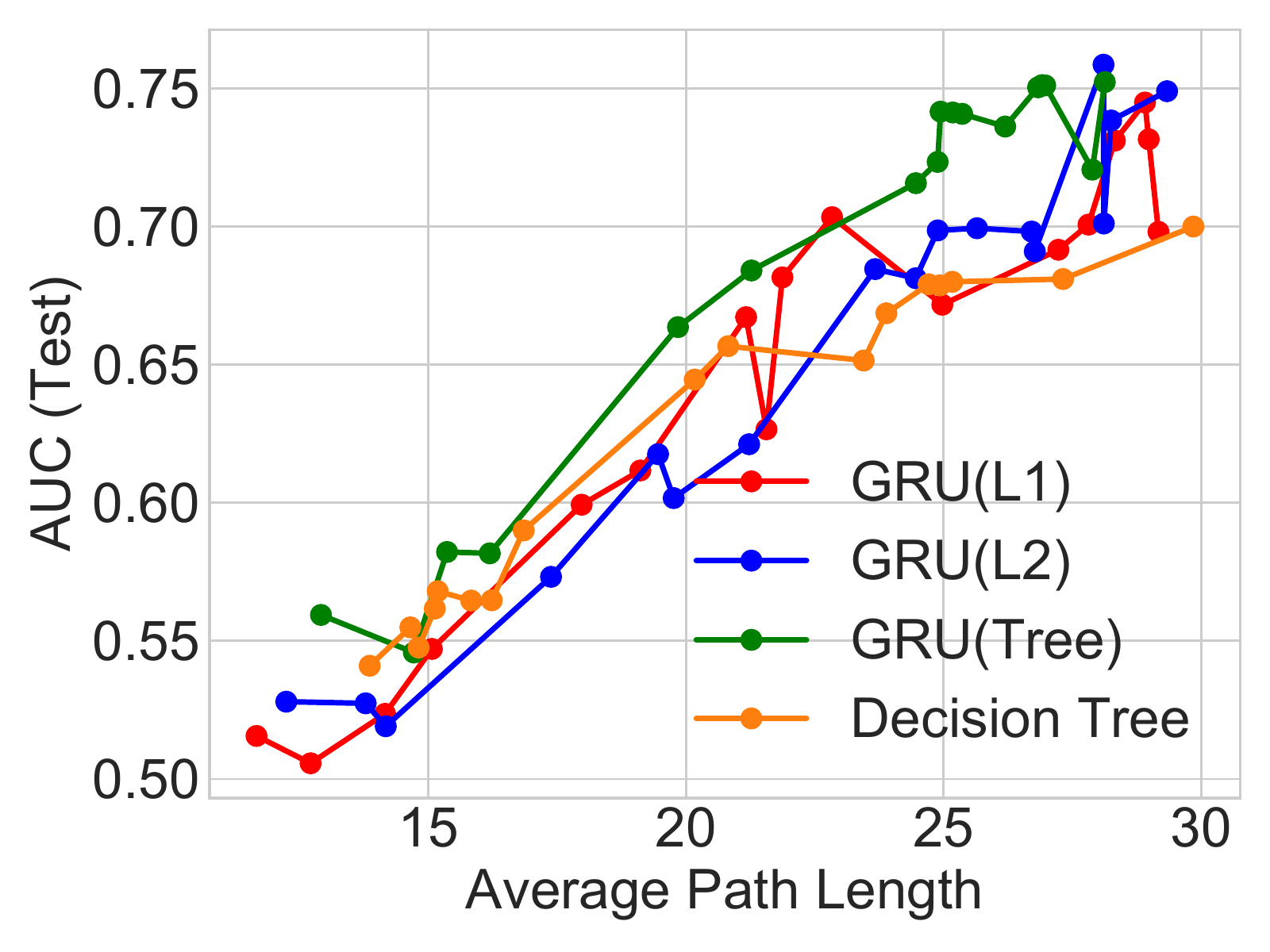}
        \label{appfig:hiv:therapy}
    \end{subfigure}
    \begin{subfigure}[b]{0.32\linewidth}
        \caption{CD4$^{+}$ $\leq$ 200 cells/ml}
        \includegraphics[width=\linewidth]{HIV/CD4_below200_GRU.pdf}
        \label{appfig:hiv:cd4}
    \end{subfigure}
    \begin{subfigure}[b]{0.32\linewidth}
        \caption{Adherence}
        \includegraphics[width=\linewidth]{HIV/adherence_gru.pdf}
        \label{appfig:hiv:adherence}
    \end{subfigure}
    \begin{subfigure}[b]{0.32\linewidth}
        \caption{Mortality}
        \includegraphics[width=\linewidth]{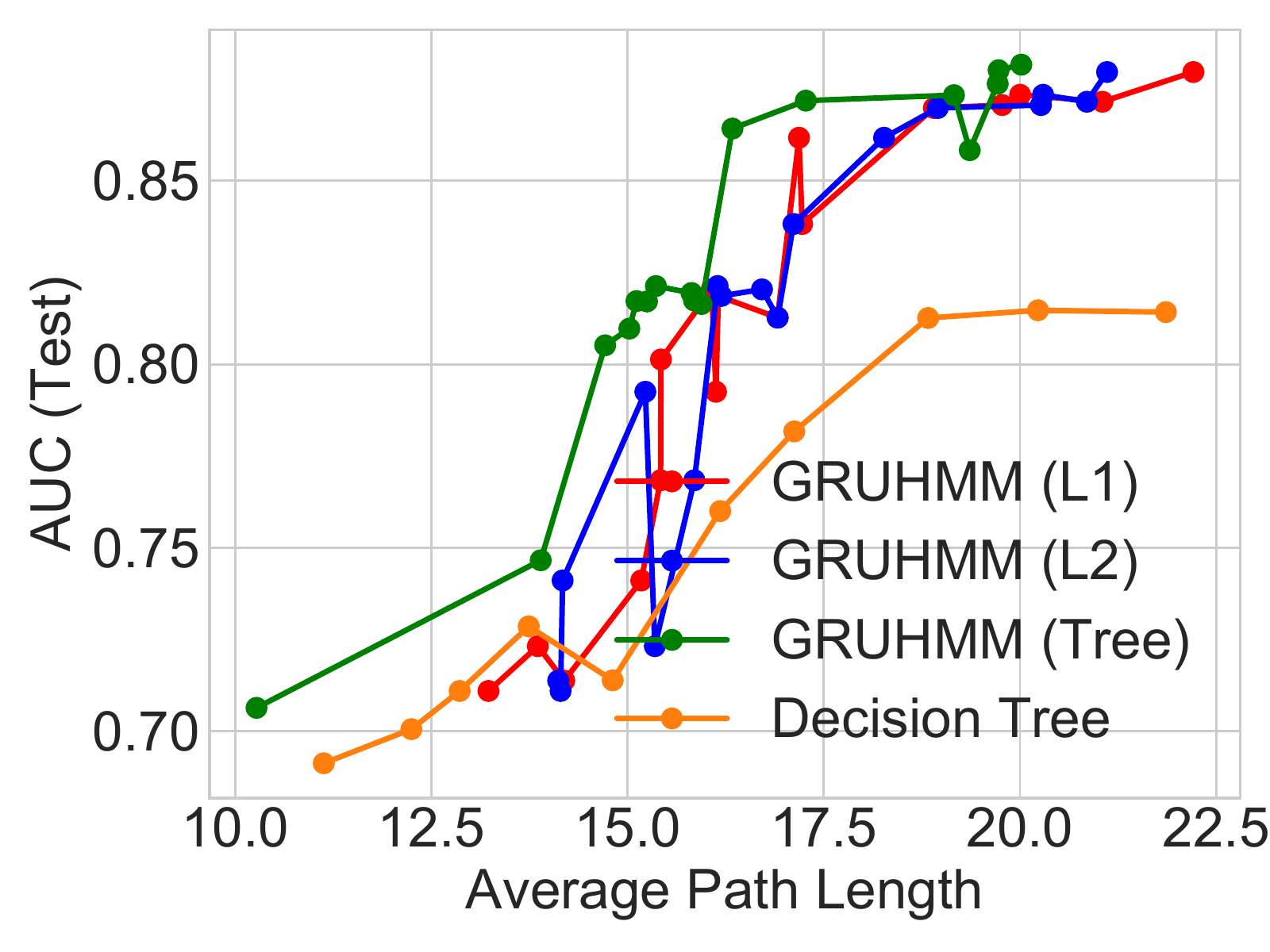}
        \label{appfig:hiv:onsetaids}
    \end{subfigure}
    \begin{subfigure}[b]{0.32\linewidth}
        \caption{Onset of AIDS}
        \includegraphics[width=\linewidth]{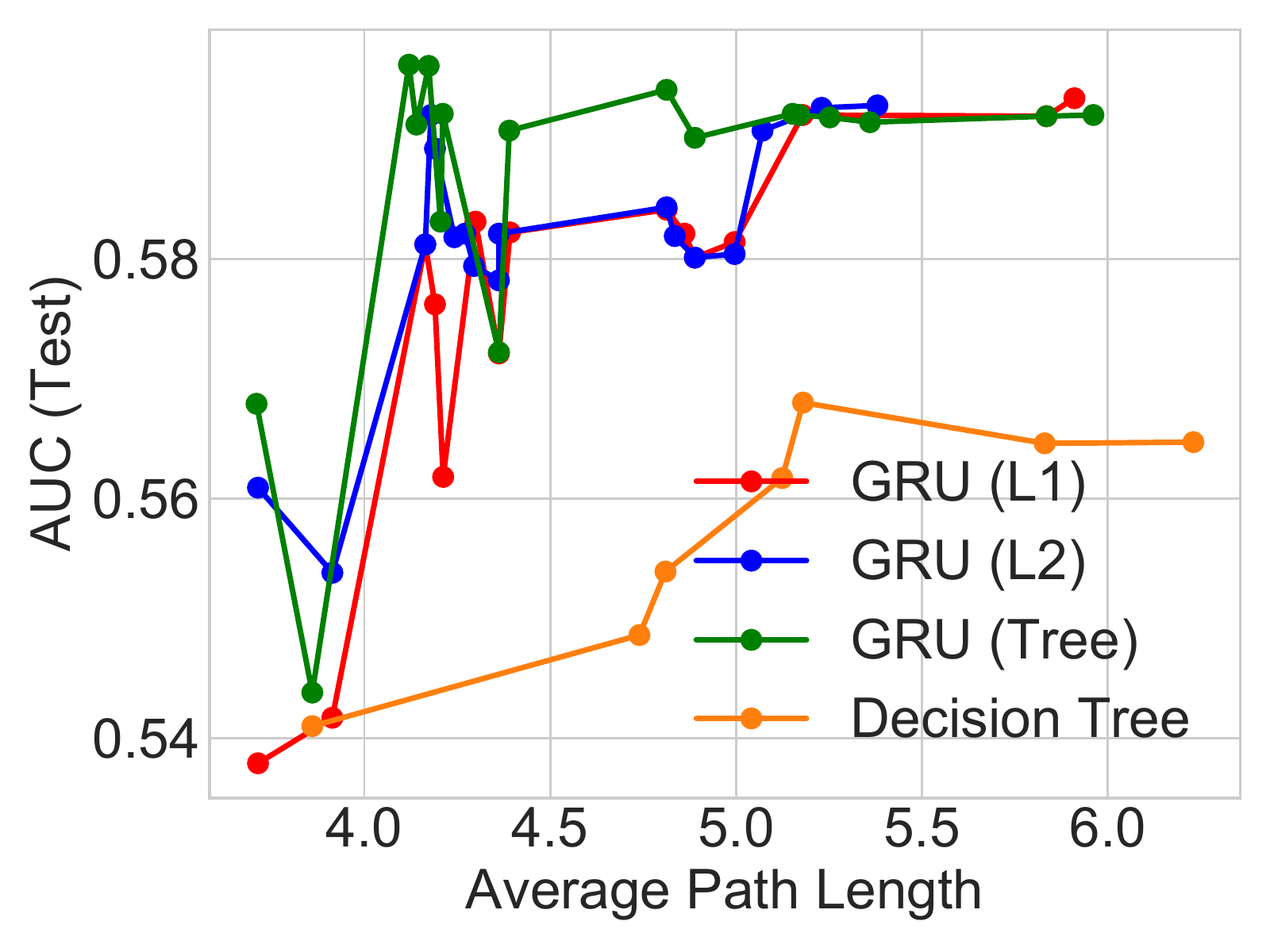}
        \label{appfig:hiv:mortality}
    \end{subfigure}
    \caption{Performance and complexity trade-offs using L1, L2, and Tree regularization on GRU for the HIV dataset.
    The 5 outputs shown here were trained jointly.}
\end{figure*}

\newpage
\subsection{HIV: AUCs}
\begin{table*}[h]
    \centering
    \resizebox{0.8\linewidth}{!}{%
    \begin{tabular}{r | l l l r r r r}
        Model & \shortstack{Poor \\ Adherence} & \shortstack{Mortality} & \shortstack{CD4$^{+}$ \\ Count $\leq$ 200} & \shortstack{Therapy\\ Success} & \shortstack{Total Average \\ Path Length} & \shortstack{Parameter \\ Count}\\ [0.5ex]
        \hline
        logreg & 0.6884 & 0.7031 & 0.5741 & 0.6092 &38.942  & 1155 \\
        \hline
        decision tree & 0.7100 & 0.7601 & 0.5937 & 0.6286 & 62.150  & - \\
        \hline
        hmm (5) & 0.7106 & 0.7611 & 0.6012 & 0.6265 &41.864  & 865 \\
        hmm (10) & 0.7287 & 0.7627 & 0.6237 & 0.6409 & 46.309 & 1780 \\
        hmm (25) & 0.7243 & 0.7627 & 0.6327 & 0.6384 & 56.159 & 4825 \\
        hmm (50) & 0.7181 & 0.7639 & 0.6412 & 0.6370 & 69.014 & 10900 \\
        hmm (75) & 0.7244 & 0.7661 & 0.6294 & 0.6518 & 70.476 & 18225 \\
        hmm (100) & 0.7261 & 0.7657 & 0.6287 & 0.6524 & 71.159 & 26800 \\
        \hline
        gru (5) & 0.6457 & 0.6814 & 0.6695 & 0.6834 & 58.347 & 1310 \\
        gru (25) & 0.7516 & 0.7986 & 0.7073 & 0.6991 & 60.072 & 8050 \\
        gru (50) & 0.7011 & 0.8290 & 0.6995 & 0.7054 & 67.513 & 19850 \\
        gru (75) & 0.7623 & 0.8514 & 0.7117 & 0.7490 & 64.870 & 35400 \\
        gru (100) & 0.7340 & 0.8216 & 0.6981 & 0.7235 & 67.183 & 54700 \\
        \hline
        grutree (100/0.01) & 0.7176 & 0.7948 & 0.7046 & 0.6803 & 91.020 & 54700 \\
        grutree (100/1.0) & 0.7134 & 0.7997 & 0.7138 & 0.6892  & 86.774 & 54700 \\
        grutree (100/20.0) & 0.7157 & 0.8066 & 0.7216 & 0.7114 & 76.025 & 54700 \\
        grutree (100/70.0) & 0.7485 & 0.8210 & 0.7413 & 0.7060 & 68.952 & 54700 \\
        grutree (100/300.0) & 0.7251 & 0.8178 & 0.7264 & 0.6746 & 54.058 & 54700 \\
        grutree (100/2\,000.0) & 0.7030 & 0.8169 & 0.6342 & 0.6627 & 49.839 & 54700 \\
        grutree (100/5\,000.0) & 0.6549 & 0.7582 & 0.6142 & 0.6352 & 23.895 & 54700 \\
        grutree (100/7\,000.0) & 0.6167 & 0.7524 & 0.5740 & 0.5634 & 15.283 & 54700 \\
        grutree (100/8\,000.0) & 0.5874 & 0.7412 & 0.5003 & 0.5027 & 7.391 & 54700 \\
        \hline
        gruhmm (5/5) & 0.6430 & 0.6647 & 0.5418 & 0.6479 & 67.619 & 2175 \\
        gruhmm (5/10) & 0.6708 & 0.6720 & 0.5879 & 0.6517 & 72.137 & 3090 \\
        gruhmm (5/25) & 0.6951 & 0.6981 & 0.6476 & 0.6955 &  68.200 & 6135 \\
        gruhmm (5/50) & 0.6810 & 0.7002 & 0.6760 & 0.7114 & 71.518 & 12210 \\
        gruhmm (10/5) & 0.7018 & 0.7147 & 0.7049 & 0.7208 & 64.852 & 3635 \\
        gruhmm (10/10) & 0.7190 & 0.7378 & 0.7136 & 0.7578 & 73.252 & 4550 \\
        gruhmm (10/25) & 0.7264 & 0.7457 & 0.7217 & 0.7951 & 70.884 & 7595 \\
        gruhmm (10/50) & 0.7570 & 0.7522 & 0.7224 & 0.8234 & 69.726 & 13670 \\
        gruhmm (25/10) & 0.7462 & 0.7861 & 0.7152 & 0.8217 & 68.241 & 9830 \\
        gruhmm (25/25) & 0.7435 & 0.8102 & 0.7425 & 0.8186 & 79.261 & 12875 \\
        gruhmm (25/50) & 0.7484 & 0.7714 & 0.7501 & 0.8006 & 76.174 & 18950 \\
        gruhmm (50/10) & 0.7437 & 0.7668 & 0.7813 & 0.8260 & 70.081 & 21630 \\
        gruhmm (50/25) & 0.7380 & 0.7557 & 0.7824 & 0.8215 & 88.617 & 24675 \\
        gruhmm (50/50) & 0.7317 & 0.7684 & 0.7920 & 0.8007 & 97.864 & 30750 \\
        \hline
        gruhmmtree (50/50/0.5) & 0.7432 & 0.7692 & 0.8790 & 0.7804 & 73.168 & 30750 \\
        gruhmmtree (50/50/50.0) & 0.7426 & 0.8152 & 0.8914 & 0.7979 & 67.729 & 30750 \\
        gruhmmtree (50/50/200.0 & 0.7461 & 0.8308 & 0.8767 & 0.8032  & 59.025 & 30750 \\
        gruhmmtree (50/50/600.0 & 0.7467 & 0.8820 & 0.8821 & 0.8293 & 52.128  & 30750 \\
        gruhmmtree (50/50/1\,000.0) & 0.7375 & 0.8951 & 0.8739 & 0.7882 & 48.247 & 30750 \\
        gruhmmtree (50/50/4\,000.0) & 0.7242 & 0.8461 & 0.8515 & 0.8030 & 14.868 & 30750 \\
        gruhmmtree (50/50/7\,000.0) & 0.7280 & 0.8462 & 0.8313 & 0.7484 & 1.836 & 30750 \\
    \end{tabular}}
    \caption{Performance metrics for multi-dimensional classification on a held-out portion of the HIV dataset. \emph{Total Average Path Length} refers to the summed average path lengths across the output dimensions.}
    \label{table:hiv-results}
\end{table*}

\newpage
\subsection{TIMIT:Plots/Tree Visualization}

\begin{figure}[!h]
    \centering
    \begin{subfigure}[b]{0.4\linewidth}
        \includegraphics[width=\linewidth]{timit_grutree.pdf}
        \caption{TIMIT Stop Phonemes}
        \label{timit:results:plot}
    \end{subfigure}
    \begin{subfigure}[b]{\linewidth}
        \includegraphics[width=\linewidth]{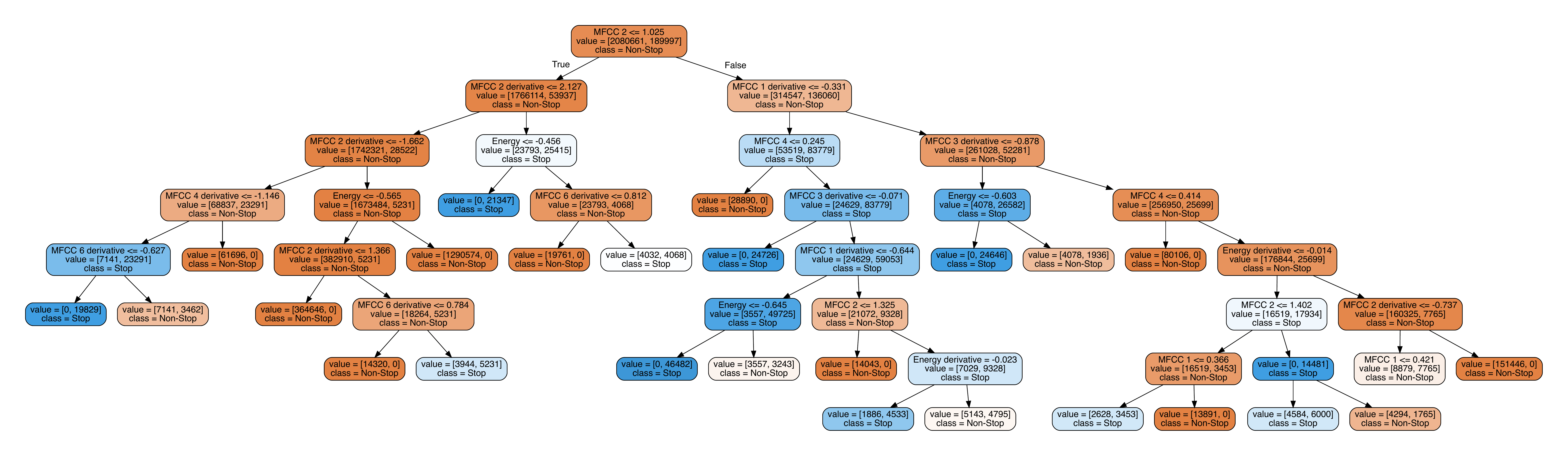}
        \caption{GRU:500}
        \label{timit:results:tree}
    \end{subfigure}
    \caption{(a) Performance and complexity trade-offs using L1, L2, and Tree regularization for GRU models on TIMIT. (b) Decision tree trained using $\lambda=500.0$ tree regularization on GRU.}
\end{figure}

\newpage
\subsection{TIMIT:AUCs}
\begin{table*}[h!]
    \centering
    \resizebox{0.55\linewidth}{!}{%
    \begin{tabular}{r | c c c c}
        Model & AUC & Average Path Length & Parameter Count \\ [0.5ex]
        \hline
        logreg & 0.7747 & 23.460 & 27 \\
        \hline
        decision tree & 0.8668 & 59.2061 & - \\
        \hline
        hmm (5) & 0.8900 & 51.911 & 295 \\
        hmm (10) & 0.8981 & 56.273 & 640 \\
        hmm (25) & 0.9129 & 57.602 & 1\,975 \\
        hmm (50) & 0.9189 & 63.752 & 5\,200 \\
        hmm (75) & 0.9251 & 71.473 & 9\,675 \\
        \hline
        gru (1) & 0.9169 & 42.602 & 86 \\
        gru (5) & 0.9451 & 49.275 & 490 \\
        gru (10) & 0.9509 & 60.079 & 1\,130 \\
        gru (25) & 0.9547 & 62.051 & 3\,950 \\
        gru (50) & 0.9578 & 64.957 & 11\,650 \\
        gru (75) & 0.9620 & 68.998 & 23\,100 \\
        \hline
        gruhmm (1/5) & 0.9419 & 54.9723 & 381 \\
        gruhmm (1/10) & 0.9535 & 53.5642 & 726 \\
        gruhmm (1/25) & 0.9636 & 57.3290 & 2601 \\
        gruhmm (5/5) & 0.9569 & 55.9531 & 785 \\
        gruhmm (5/10) & 0.9575 & 57.6199 & 1\,130 \\
        gruhmm (5/25) & 0.9603 & 59.9925 & 2\,465\\
        gruhmm (10/5) & 0.9626 & 57.0652 & 1\,425 \\
        gruhmm (10/10) & 0.9641 & 60.7877 & 1\,770\\
        gruhmm (10/25) & 0.9651 & 61.0018 & 3\,105 \\
        gruhmm (25/5) & 0.9635 & 57.5288 & 4\,245 \\
        gruhmm (25/10) & 0.9657 & 60.5212 & 4\,590 \\
        gruhmm (25/25) & 0.9663 & 65.0161 & 5\,925\\
        gruhmm (50/5) & 0.9676 & 62.2378 & 11\,945 \\
        gruhmm (50/10) & 0.9679 & 65.1191 & 12\,290 \\
        gruhmm (50/25) & 0.9685 & 67.4301 & 13\,625\\
        \hline
        grutree (75/0.01) & 0.9517 & 66.2801 & 23\,100 \\
        grutree (75/0.1) & 0.9466 & 62.4316 & 23\,100 \\
        grutree (75/0.5) & 0.9367 & 60.8764 & 23\,100 \\
        grutree (75/2.0) & 0.9311 & 58.3659 & 23\,100 \\
        grutree (75/5.0) & 0.9302 & 55.7588 & 23\,100 \\
        grutree (75/10.0) & 0.9288 & 46.6616 & 23\,100 \\
        grutree (75/100.0) & 0.8911 & 40.1123 & 23\,100 \\
        grutree (75/500.0) & 0.8998 & 28.4240 & 23\,100 \\
        grutree (75/700.0) & 0.8628 & 25.136 & 23\,100 \\
        grutree (75/800.0) & 0.7471 & 22.6671 & 23\,100 \\
        grutree (75/1\,000.0) & 0.7082 & 17.1523 & 23\,100 \\
        grutree (75/6\,000.0) & 0.5441 & 11.1108 & 23\,100 \\
        grutree (75/7\,000.0) & 0.5088 & 8.9910 & 23\,100 \\
        \hline
        gruhmmtree (50/25/0.1) & 0.9507 & 69.1110 & 13\,625 \\
        gruhmmtree (50/25/1.0) & 0.9465 & 67.5773 & 13\,625 \\
        gruhmmtree (50/25/6.0) & 0.9515 & 65.1494 & 13\,625 \\
        gruhmmtree (50/25/20.0) & 0.9449 & 64.0072 & 13\,625 \\
        gruhmmtree (50/25/30.0) & 0.9482 & 62.5406 & 13\,625 \\
        gruhmmtree (50/25/70.0) & 0.9460 & 58.0111 & 13\,625 \\
        gruhmmtree (50/25/100.0) & 0.9470 & 51.2417 & 13\,625 \\
        gruhmmtree (50/25/500.0) & 0.9401 & 42.1882 & 13\,625 \\
        gruhmmtree (50/25/700.0) & 0.9352 & 40.1281 & 13\,625 \\
        gruhmmtree (50/25/1\,000.0) & 0.9390 & 38.0072 & 13\,625 \\
        gruhmmtree (50/25/3\,000.0) & 0.9280 & 25.9120 & 13\,625 \\
        gruhmmtree (50/25/4\,000.0) & 0.9311 & 21.7170 & 13\,625 \\
        gruhmmtree (50/25/7\,000.0) & 0.9290 & 10.1122 & 13\,625 \\
        gruhmmtree (50/25/9\,000.0) & 0.9134 & 1.0563 & 13\,625 \\
        gruhmmtree (50/25/10\,000.0) & 0.9125 & 0.0000 & 13\,625 \\
    \end{tabular}}
    \caption{Performance metrics across models on a held-out portion of the TIMIT dataset.}
    \label{table:timit-results}
\end{table*}

\newpage
\section{GRU-HMM: Deep Residual Timeseries Model}
\paragraph{Hidden Markov Model}
For our purposes, Hidden Markov Models (HMMs) can be viewed as
stochastic RNNs which can be interpreted as probabilistic generative
models. In this work, we consider an HMM to generate a latent variable sequence $\textbf{z} = [z_1, \ldots z_{T}]$
via a Markov chain, where each latent indicates one of $K$ possible discrete states: $z_{t} \in \left \{1, ..., K \right \}$. This state sequence is then used to jointly produce the ``data'' $x_{t}$ and ``outcomes'' $y_{t}$ observed at each timestep. The joint
distribution over $\textbf{z}, \textbf{x}, \textbf{y}$ factorizes as:
\begin{align}
p(\textbf{z}, \textbf{y}) = \pi_{0}(z_{0})\prod_{t=1}^{T} p(z_{t}|z_{t-1}, A) \cdot p(x_t | z_t, \phi) \mbox{Bern}(y_{t}| \sigma(\sum_{k} w_k \delta_k(z_{t}))),
\end{align}
where $A$ is a transition matrix such that $A_{i, j}
= \textup{Pr}(z_{t} = i | z_{t-1} = j)$,
$\pi_{0} = p(z_{0})$ is the initial state distribution,
$\left \{ \phi_{k} \right \}_{k=1}^{K}$ are the emission parameters that generate data.
We can then apply the same objective as above for training.


\paragraph{GRU-HMM: Modeling the residuals of an HMM.}

We now consider an additional model, the GRU-HMM, designed for interpretability.
The idea is to use a GRU to
to model the residual errors when predicting the binary target via the HMM belief states.
We can further penalize the complexity of the GRU predictions via our tree regularization, so that higher-quality predictions do not come at the price of a much less interpretable model.

\begin{figure}[!h]
    \centering
    \includegraphics[width=0.75\linewidth]{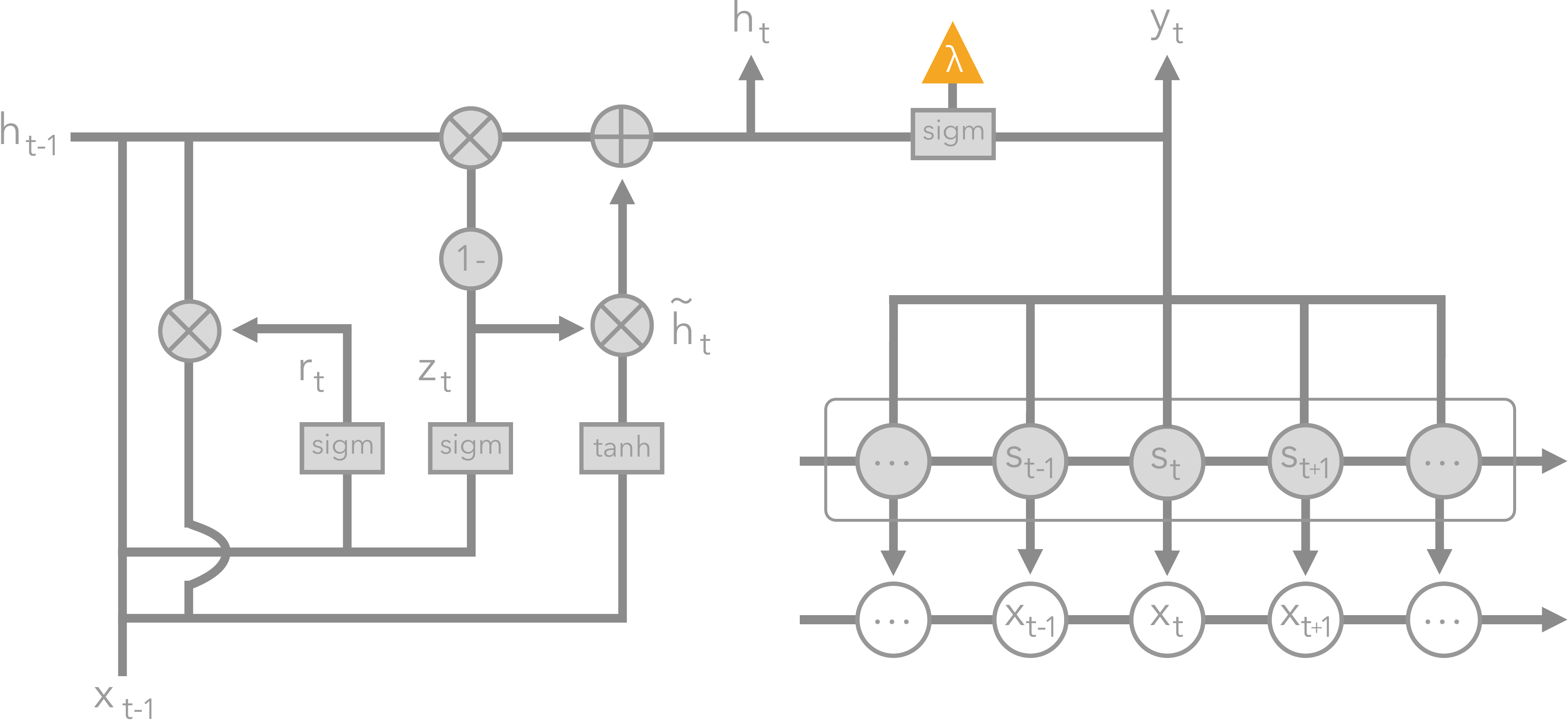}
    \caption{Deep residual model: GRU-HMM.
        The orange triangle indicates the output used
        in surrogate training for tree regularization.}
    \label{fig:gru-hmm-we-regularize:gruhmm}
\end{figure}

We train the deep residual model on the same suite of synthetic and real world
datasets. See Tables \ref{table:synthetic-results}, \ref{table:sepsis-results}, \ref{table:timit-results} for a comparison of GRU-HMM with vanilla GRU and HMM
models under different regularization and expressiveness parameters. We can see that across the datasets, deep residual models perform around 1\% better than
their vanilla equivalents with roughly the same number of model parameters.

\begin{figure*}[!h]
    \centering
    \begin{subfigure}[b]{0.32\linewidth}
        \caption{In-Hospital Mortality}
        \includegraphics[width=\linewidth]{sepsis_gruhmmtree_0.pdf}
        \label{fig:sepsis:gruhmmtree:0}
    \end{subfigure}
    \begin{subfigure}[b]{0.32\linewidth}
        \caption{90-Day Mortality}
        \includegraphics[width=\linewidth]{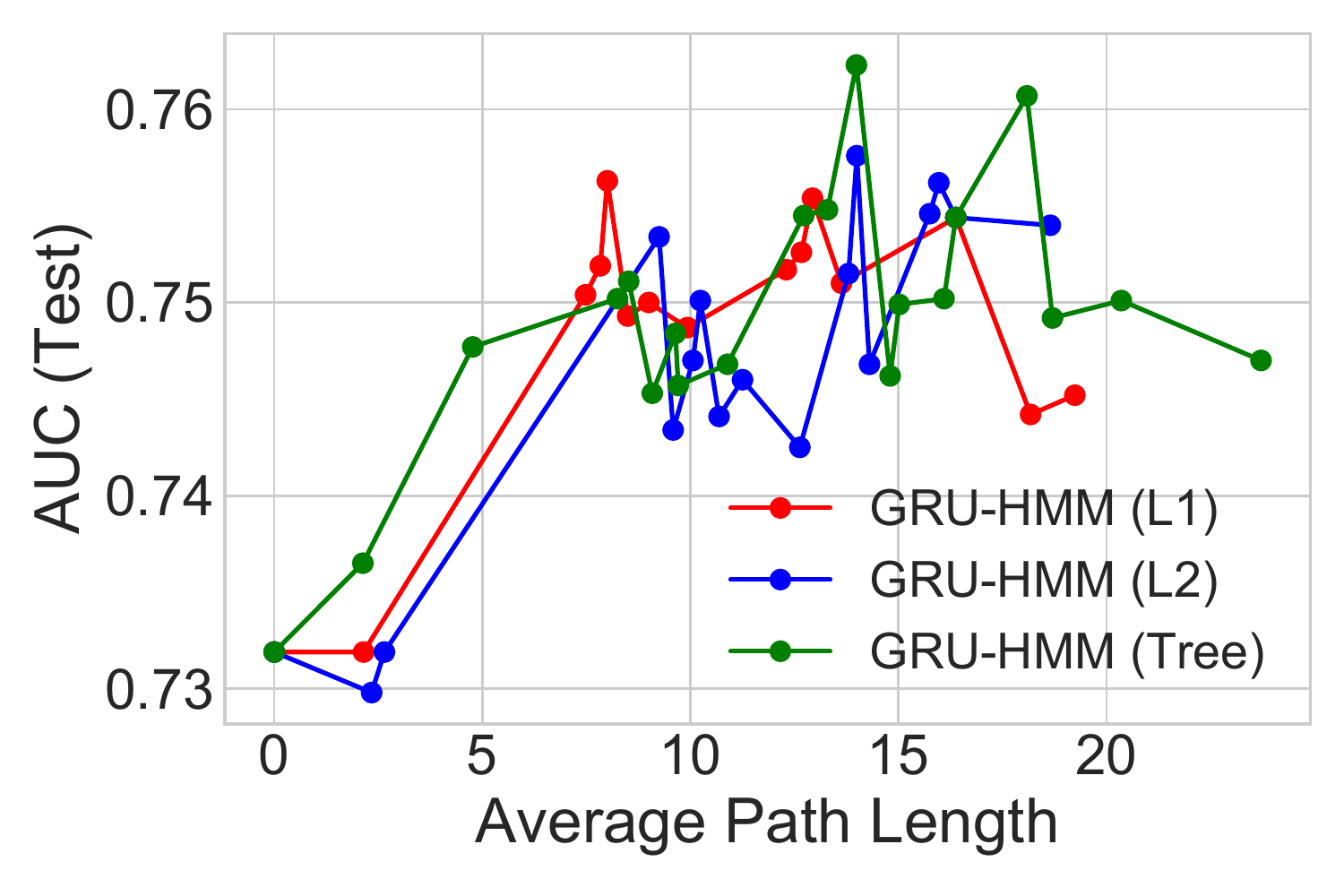}
        \label{fig:sepsis:gruhmmtree:1}
    \end{subfigure}
    \begin{subfigure}[b]{0.32\linewidth}
        \caption{Mechanical Ventilation}
        \includegraphics[width=\linewidth]{sepsis_gruhmmtree_2.pdf}
        \label{fig:sepsis:gruhmmtree:2}
    \end{subfigure}
    \begin{subfigure}[b]{0.32\linewidth}
        \caption{Median Vasopressor}
        \includegraphics[width=\linewidth]{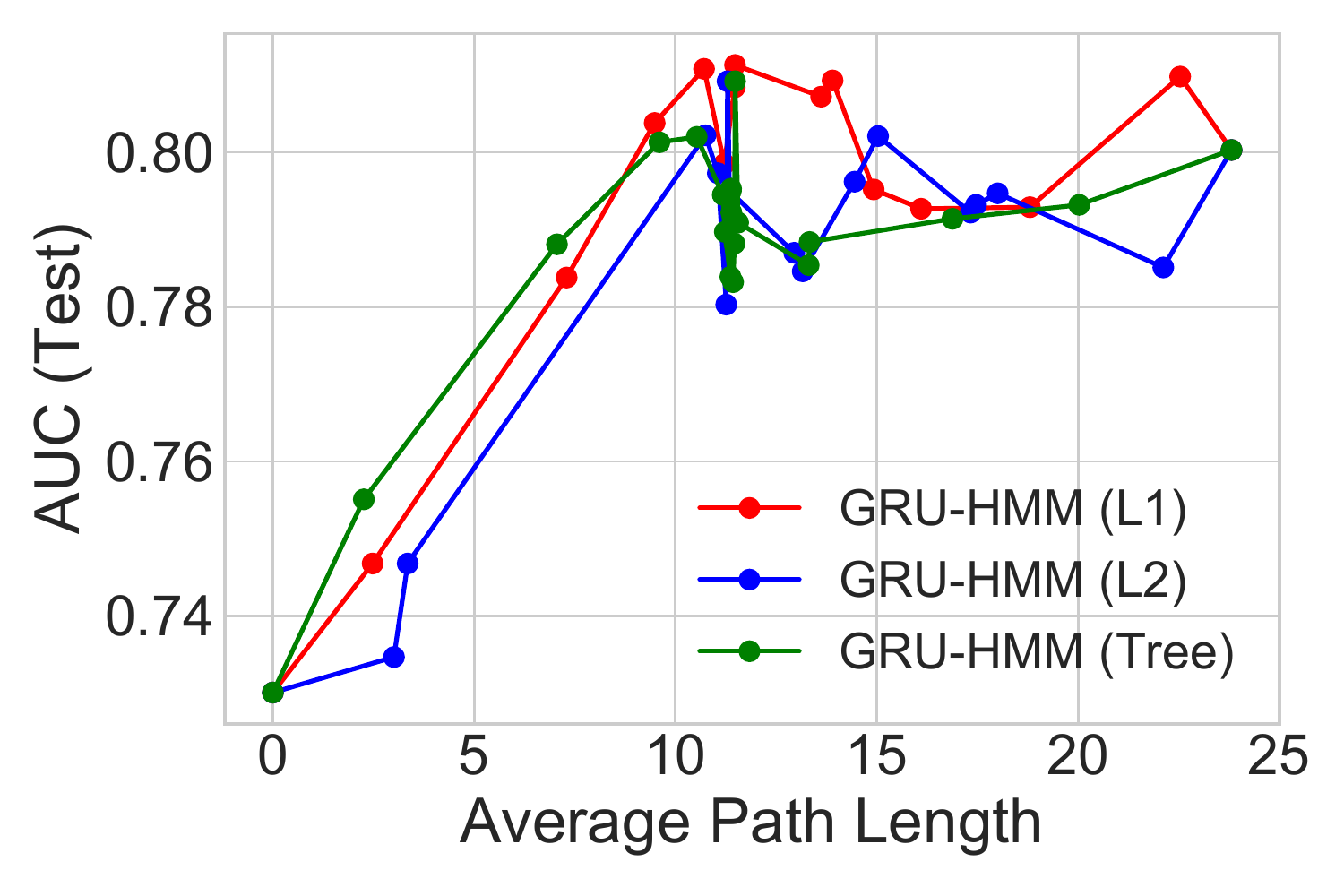}
        \label{fig:sepsis:gruhmmtree:3}
    \end{subfigure}
    \begin{subfigure}[b]{0.32\linewidth}
        \caption{Max Vasopressor}
        \includegraphics[width=\linewidth]{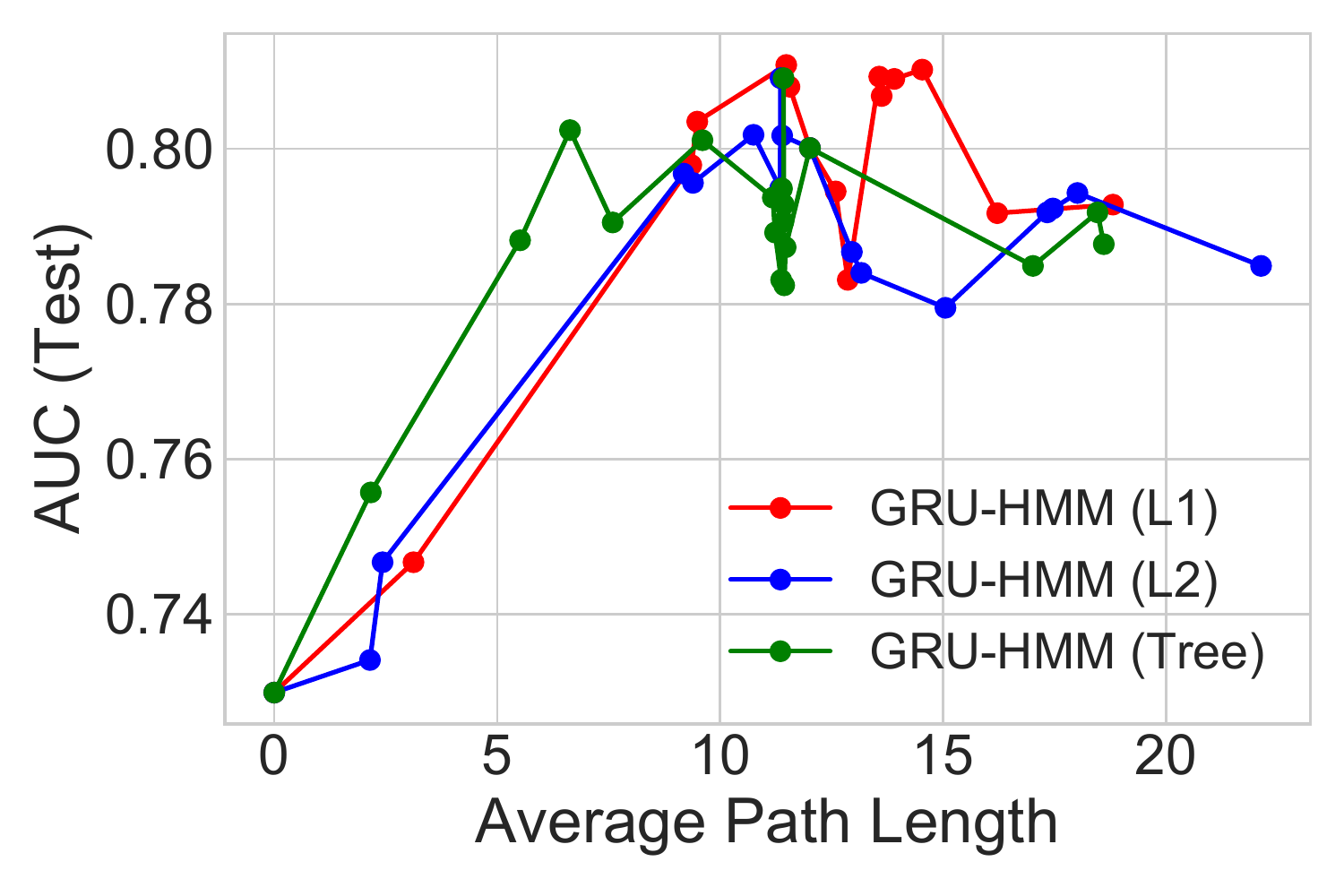}
        \label{fig:sepsis:gruhmmtree:4}
    \end{subfigure}
    \caption{Performance and complexity trade-offs using L1, L2, and Tree
    regularization on GRU-HMM performance on the Sepsis dataset.}
    \label{fig:sepsis-gruhmm-plots}
\end{figure*}

\begin{figure*}[!h]
    \centering
    \begin{subfigure}[b]{0.19\linewidth}
        \includegraphics[width=\linewidth]{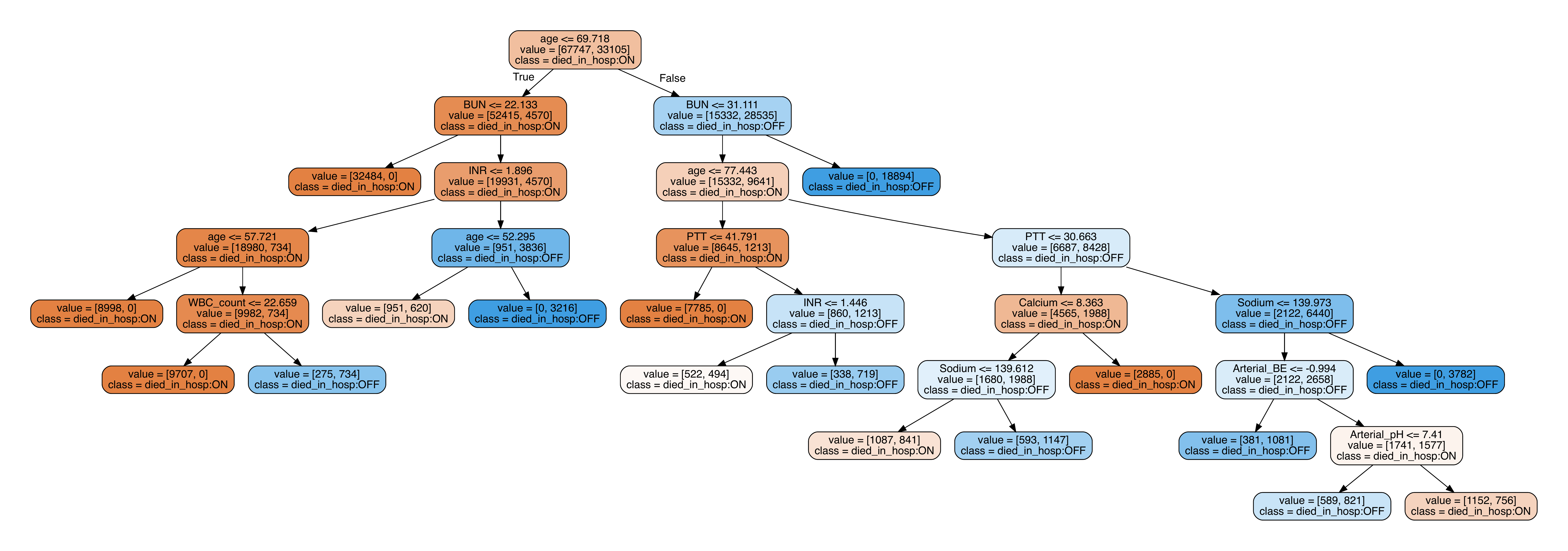}
        \label{fig:sepsis:gru:hmm:tree:dim:1}
        \caption{In-Hospital Mortality}
    \end{subfigure}
    \begin{subfigure}[b]{0.19\linewidth}
        \includegraphics[width=\linewidth]{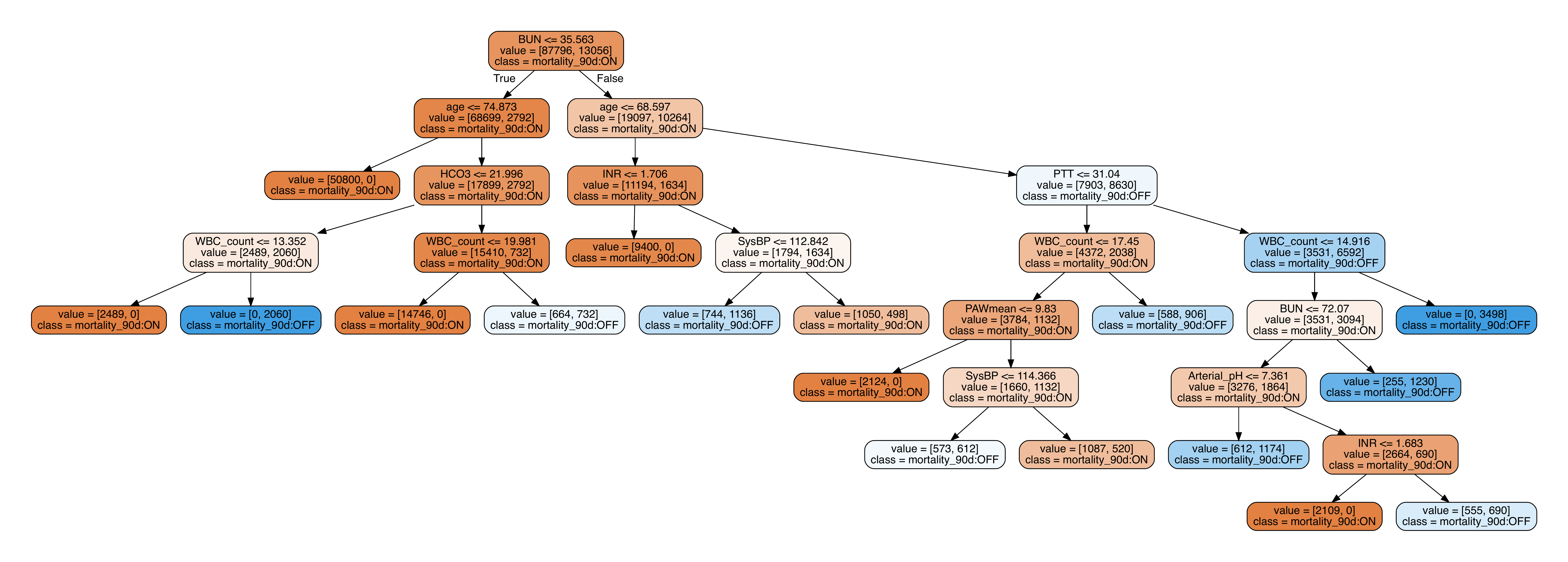}
        \label{fig:sepsis:gru:hmm:tree:dim:2}
        \caption{90-Day Mortality}
    \end{subfigure}
    \begin{subfigure}[b]{0.19\linewidth}
        \includegraphics[width=\linewidth]{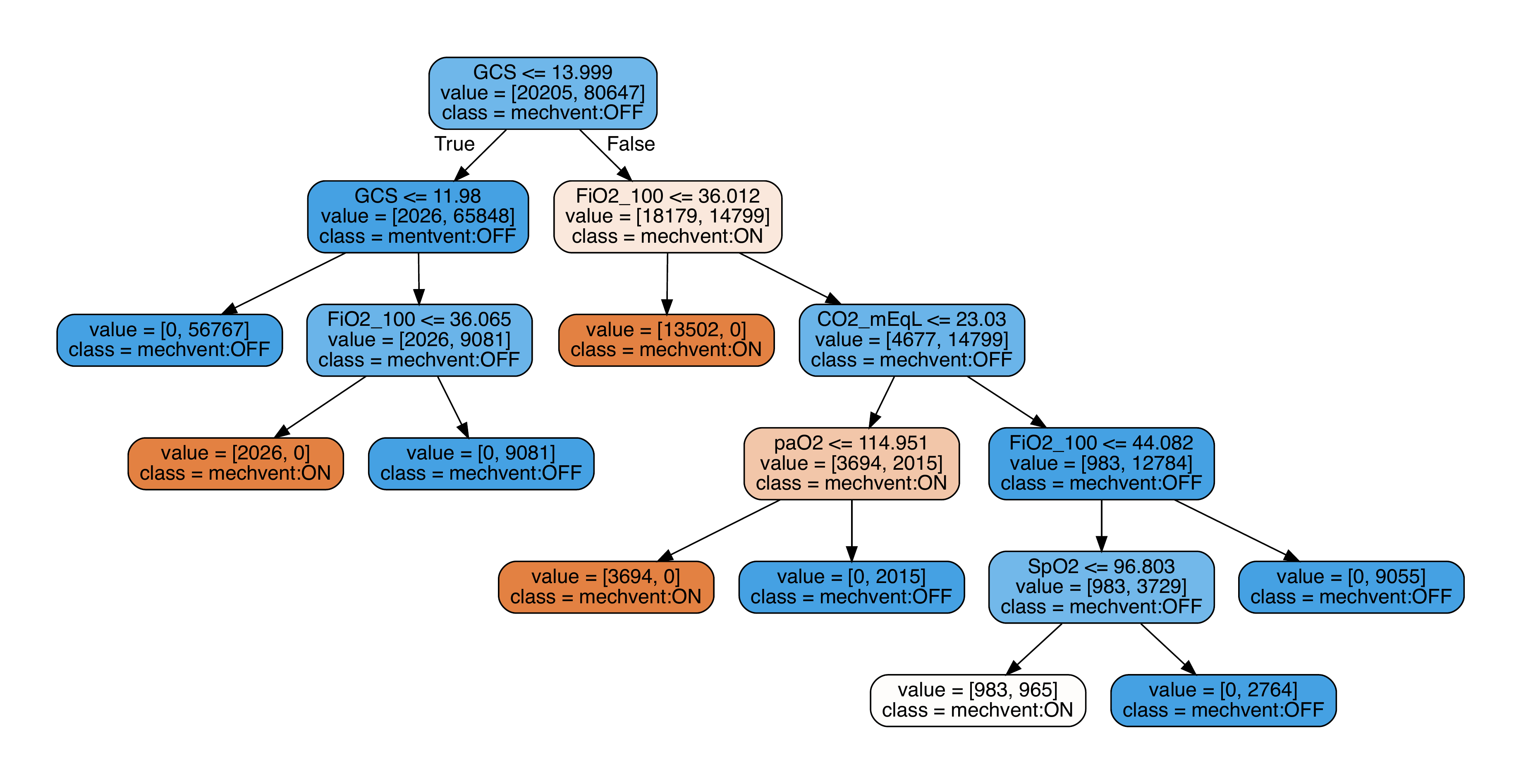}
        \label{fig:sepsis:gru:hmm:tree:dim:3}
        \caption{Mechanical Ventilation}
    \end{subfigure}
    \begin{subfigure}[b]{0.19\linewidth}
        \includegraphics[width=\linewidth]{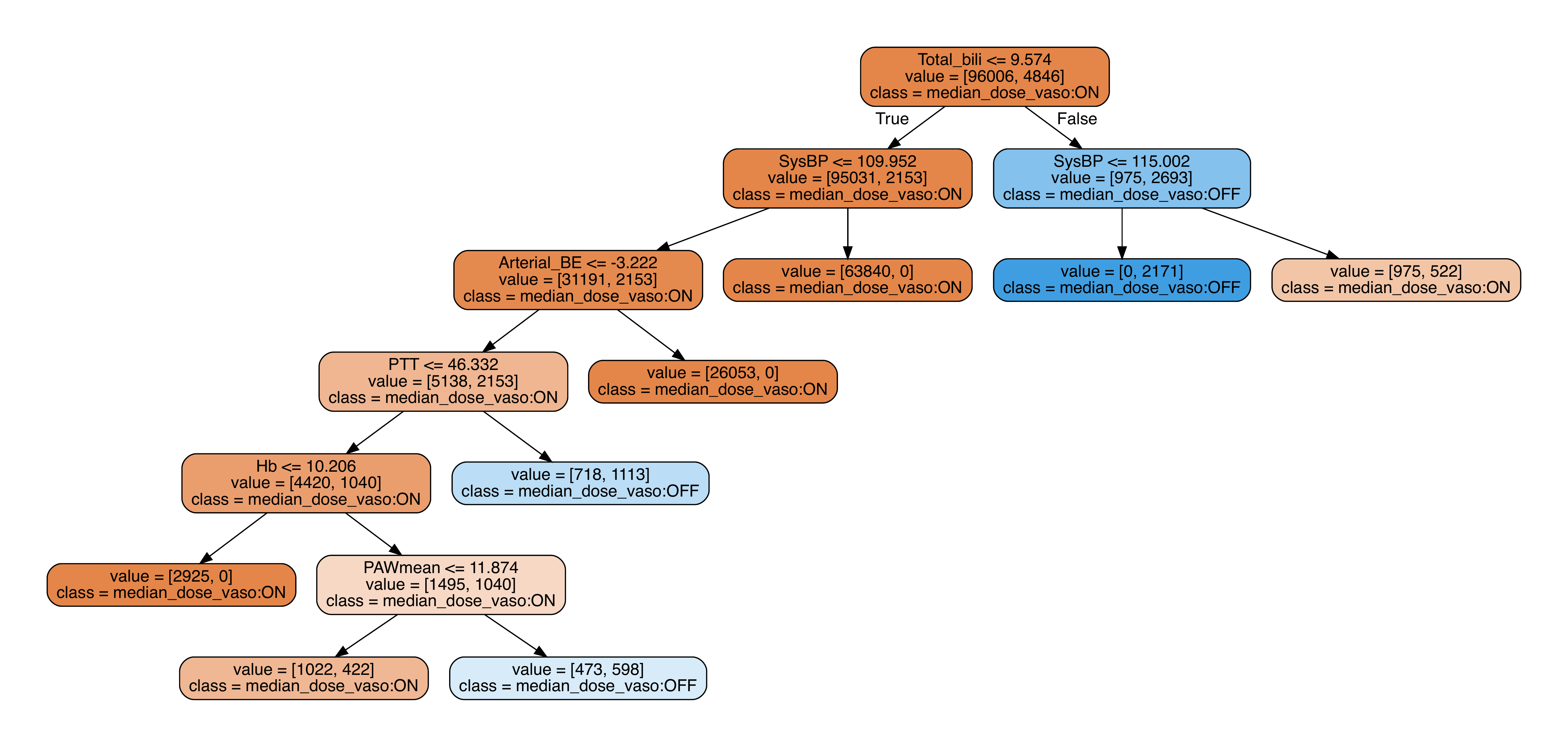}
        \label{fig:sepsis:gru:hmm:tree:dim:4}
        \caption{Median Vasopressor}
    \end{subfigure}
    \begin{subfigure}[b]{0.19\linewidth}
        \includegraphics[width=\linewidth]{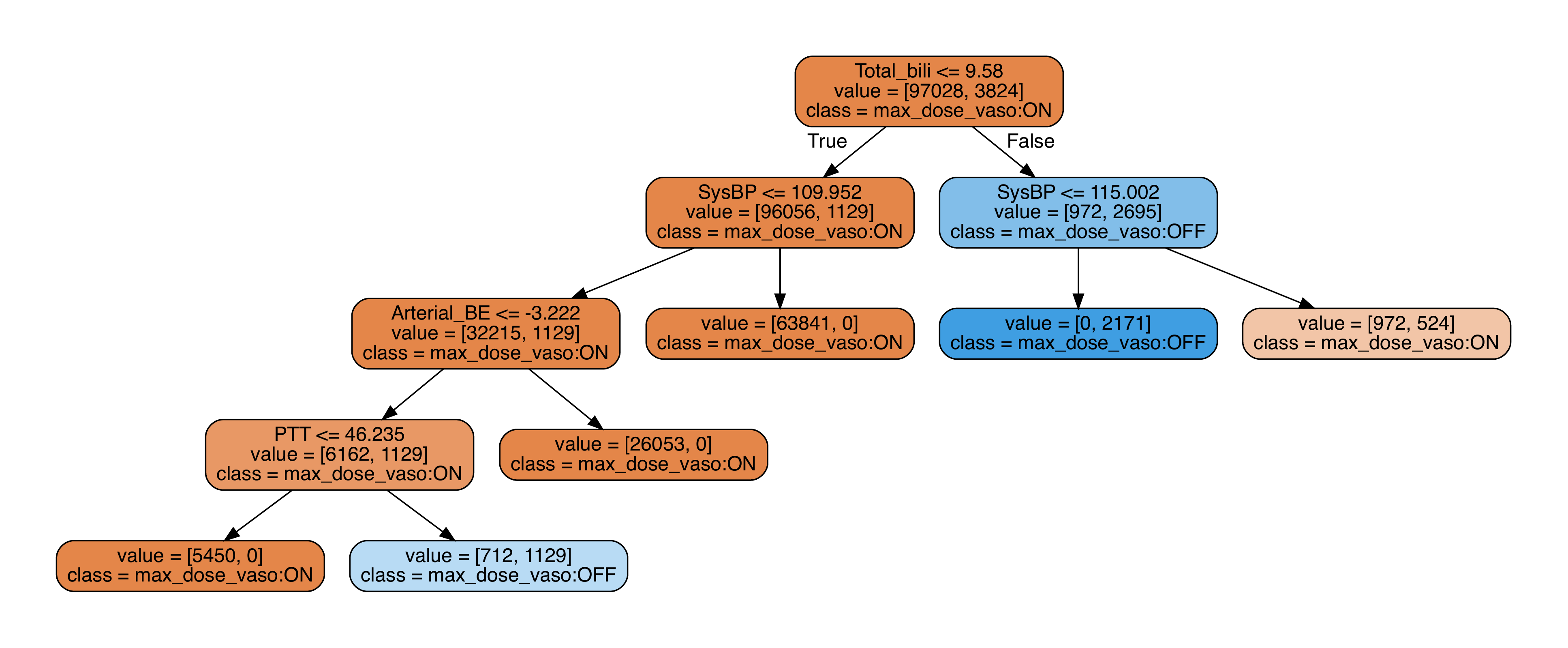}
        \label{fig:sepsis:gru:hmm:tree:dim:5}
        \caption{Max Vasopressor}
    \end{subfigure}
    \caption{Decision trees trained using Tree regularization ($\lambda=2000.0$) from GRU-HMM predictions on the Sepsis dataset.}
    \label{fig:sepsis-gruhmm-trees}
\end{figure*}

By nature of being a residual model, decision trees were trained only
on the GRU output node, leaving the HMM unconstrained. See Figure \ref{fig:gru-hmm-we-regularize:gruhmm} for a pictoral
representation. Similar to what we did for GRU models, figures
\ref{fig:2hmm:gruhmm:plot}, \ref{fig:sepsis-gruhmm-plots} compare
model performance as the $\lambda$ parameter for L1, L2, and Tree regularization
increase. We can see a similar albeit less pronounced effect where Tree
regularization dominates other methods in low node count regions. It is important
to notice the range of the AUC axis in these figures, where the worst the residual
model can performance is the HMM-only AUC. Figure \ref{fig:sepsis-gruhmm-trees}
show the regularized trees produced by the GRU-HMM. Although they share some
structure  with Figure \ref{table:sepsis-gru-tree}, there are important distinctions
that encourage us to conclude that the GRU in a residual models performs a
different role than when trained alone.

\newpage
\subsection{GRU-HMM: Sepsis Plots}

\subsection{GRU-HMM: Sepsis Tree Visualization}

\newpage
\subsection{GRU-HMM: HIV Plots/Tree Visualization}
\begin{figure*}[!h]
    \centering
    \begin{subfigure}[b]{0.42\linewidth}
        \includegraphics[width=\linewidth]{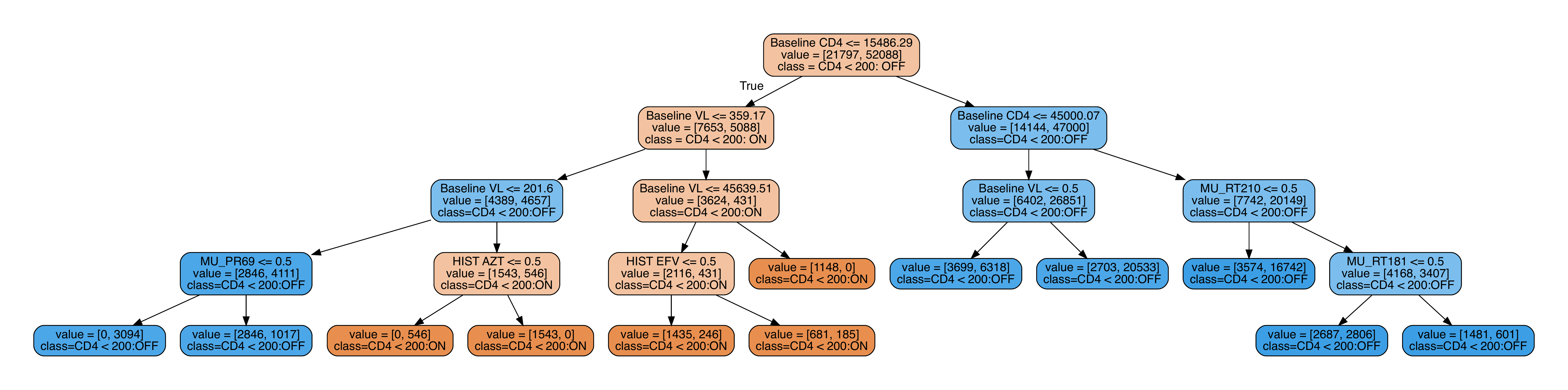}
        \label{fig:timit:gruhmmtree:tree}
        \caption{GRU-HMM: CD4$^{+}$ $\leq$ 200 cells/ml}
    \end{subfigure}
    \begin{subfigure}[b]{0.42\linewidth}
        \includegraphics[width=\linewidth]{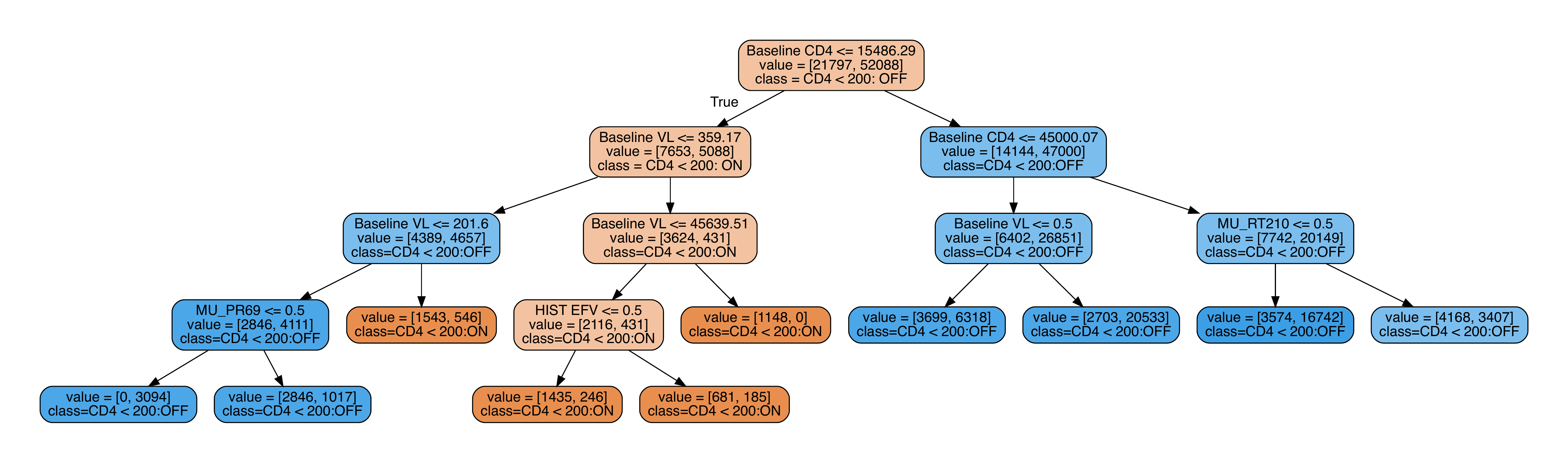}
        \label{fig:timit:gruhmmtree:tree:2}
        \caption{GRU-HMM: CD4$^{+}$ $\leq$ 200 cells/ml}
    \end{subfigure}
\caption{
\emph{HIV task:}
Study of different regularization techniques for GRU-HMM model with 75 GRU nodes and 25 HMM states, trained to predict whether CD4${+}$ $\leq$ 200 cells/ml.  
(a) Example decision tree for $\lambda = 1000.0$.
(b) Example decision tree for $\lambda = 3000.0$. 
The tree in (b) is slightly smaller than the tree in (a) as a result of the regularisation.
}
\label{fig:results:timit}
\end{figure*}
\subsection{GRU-HMM: TIMIT Plots/Tree Visualization}
\begin{figure*}[!h]
    \centering
    \begin{subfigure}[b]{0.32\linewidth}
        \includegraphics[width=\linewidth]{timit_gruhmmtree}
        \label{fig:timit:gruhmmtree}
        \caption{GRU-HMM: Stop vs Non-Stop}
    \end{subfigure}
    \begin{subfigure}[b]{0.32\linewidth}
        \includegraphics[width=\linewidth]{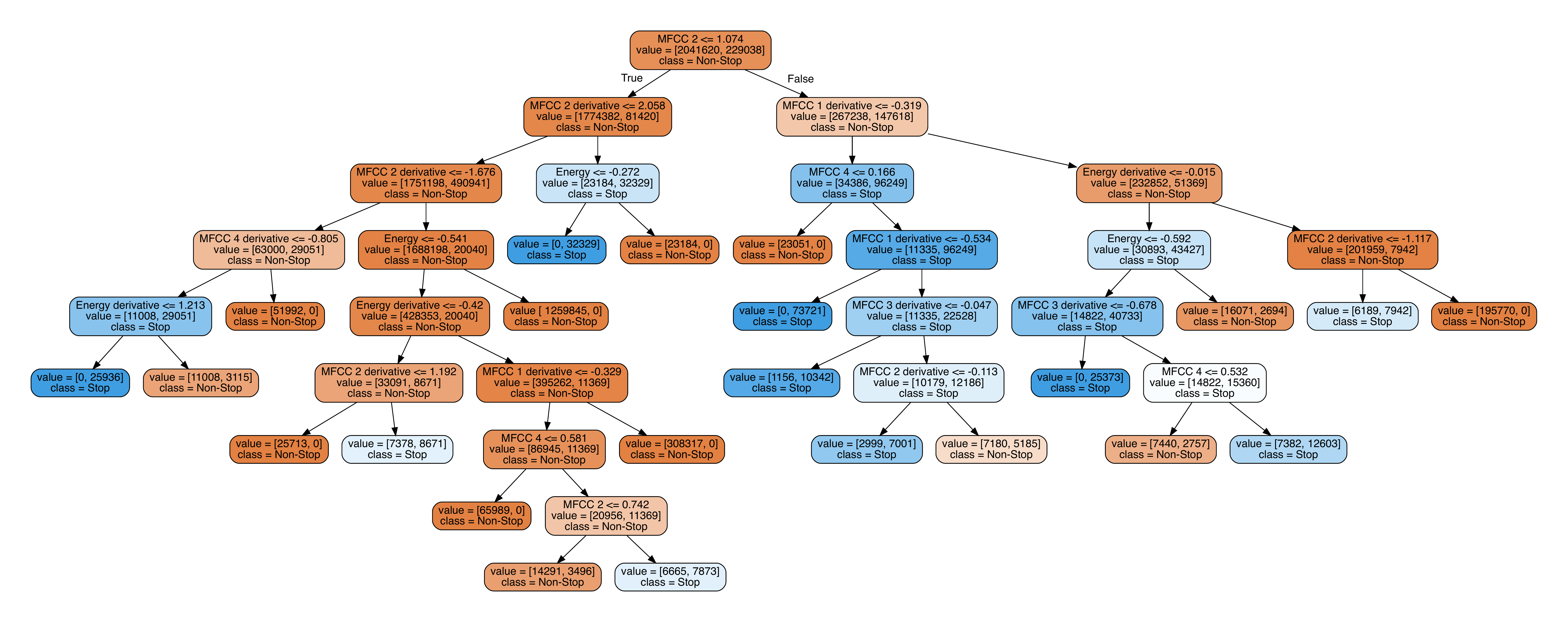}
        \label{fig:timit:gruhmmtree:tree}
        \caption{GRU-HMM: Stop vs Non-Stop}
    \end{subfigure}
    \begin{subfigure}[b]{0.32\linewidth}
        \includegraphics[width=\linewidth]{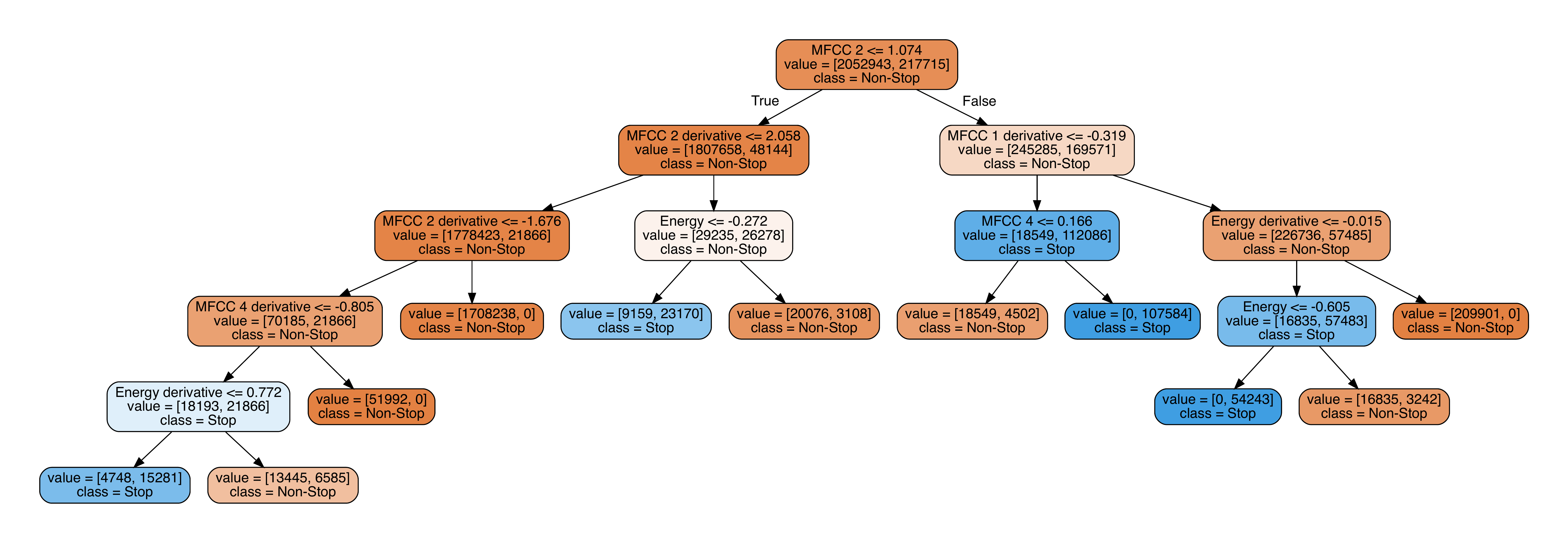}
        \label{fig:timit:gruhmmtree:tree:2}
        \caption{GRU-HMM: Stop vs Non-Stop}
    \end{subfigure}
\caption{
\emph{TIMIT task:}
Study of different regularization techniques for GRU-HMM model with 75 GRU nodes and 25 HMM states, trained to predict STOP phonemes.
(a) Tradeoff curves showing how AUC predictive power and decision-tree complexity evolve with increasing regularization strength under L1, L2, or Tree regularization.
(b) Example decision tree for $\lambda = 3000.0$.
(c) Example decision tree for $\lambda = 7000.0$. When comparing with figure \ref{timit:results:tree}, this tree is significantly smaller, suggesting that the GRU performs a different role in the residual model.
}
\label{fig:results:timit}
\end{figure*}

\newpage
\section{Runtime comparisons}

\paragraph{Training Time for Tree-Regularized Models.}
Table \ref{table:model-timings} shows the wall time for training one epoch of each of the models presented in this paper using each of the datasets. Please note that the wall times for GRU-TREE and GRU-HMM-TREE include the cost of surrogate training. If the retraining frequency is small, then the amortized cost should be small.

\begin{table*}[h!]
    \centering
    \resizebox{0.55\linewidth}{!}{%
    \begin{tabular}{l l l}
        Dataset & Model & Epoch Time (Sec.) \\ [0.5ex]
        \hline
        Signal-and-noise HMM & HMM & $16.66 \pm 2.53$ \\
        Signal-and-noise HMM & GRU & $30.48 \pm 1.92$ \\
        Signal-and-noise HMM & GRU-HMM & $50.40 \pm 5.56$ \\
        Signal-and-noise HMM & GRU-TREE & $43.83 \pm 3.84$ \\
        Signal-and-noise HMM & GRU-HMM-TREE & $73.24 \pm 7.86$ \\
        \hline
        SEPSIS & HMM & $589.80 \pm 24.11$ \\
        SEPSIS & GRU & $822.27 \pm 11.17$ \\
        SEPSIS & GRU-HMM & $1\,666.98 \pm 147.00$ \\
        SEPSIS & GRU-TREE & $2\,015.15 \pm 388.12$ \\
        SEPSIS & GRU-HMM-TREE & $2\,443.66 \pm 351.22$ \\
        \hline
        TIMIT & HMM & $1\,668.96 \pm 126.96$ \\
        TIMIT & GRU & $2\,116.83 \pm 438.83$ \\
        TIMIT & GRU-HMM & $3207.16 \pm 651.85$ \\
        TIMIT & GRU-TREE & $3\,977.01 \pm 812.11$ \\
        TIMIT & GRU-HMM-TREE & $4\,601.44 \pm 805.88$ \\
    \end{tabular}}
    \caption{Training time for recurrent models measured
    against all datasets used in this paper. Epoch time denotes the
    number of seconds it took for a single pass through all the
    training data. The epoch times for GRU-TREE and GRU-HMM-TREE
    include surrogate training expenses. If we retrain sparsely, then
    the cost of surrogate training is amortized and the epoch time for
    GRU and GRU-TREE, GRU-HMM and GRU-HMM-TREE are approximately the same. To
    measure epoch time, we used 10 HMM states, 10 GRU states, and 5 of each
    for GRU-HMM models. We trained the surrogate model for 5000 epochs.
    These tests were run on a single Intel Core i5 CPU.}
    \label{table:model-timings}
\end{table*}

\newpage
\section{Extended Stability Tests}

\begin{figure}[!h]
    \centering
    \begin{subfigure}[b]{0.24\linewidth}
        \centering
        \includegraphics[width=\linewidth]{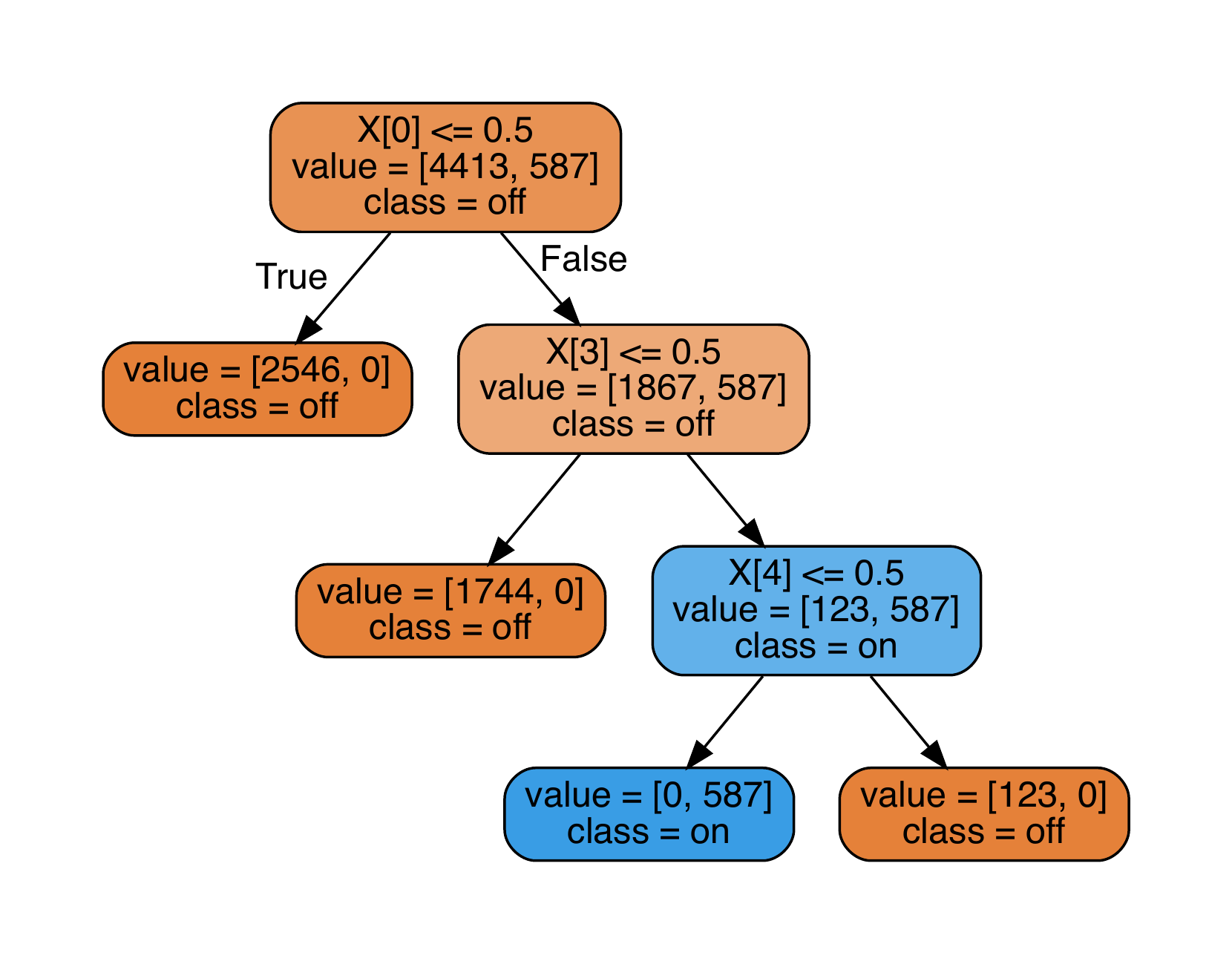}
        \caption{7/10 Runs}
        \label{}
    \end{subfigure}
    \begin{subfigure}[b]{0.24\linewidth}
        \centering
        \includegraphics[width=\linewidth]{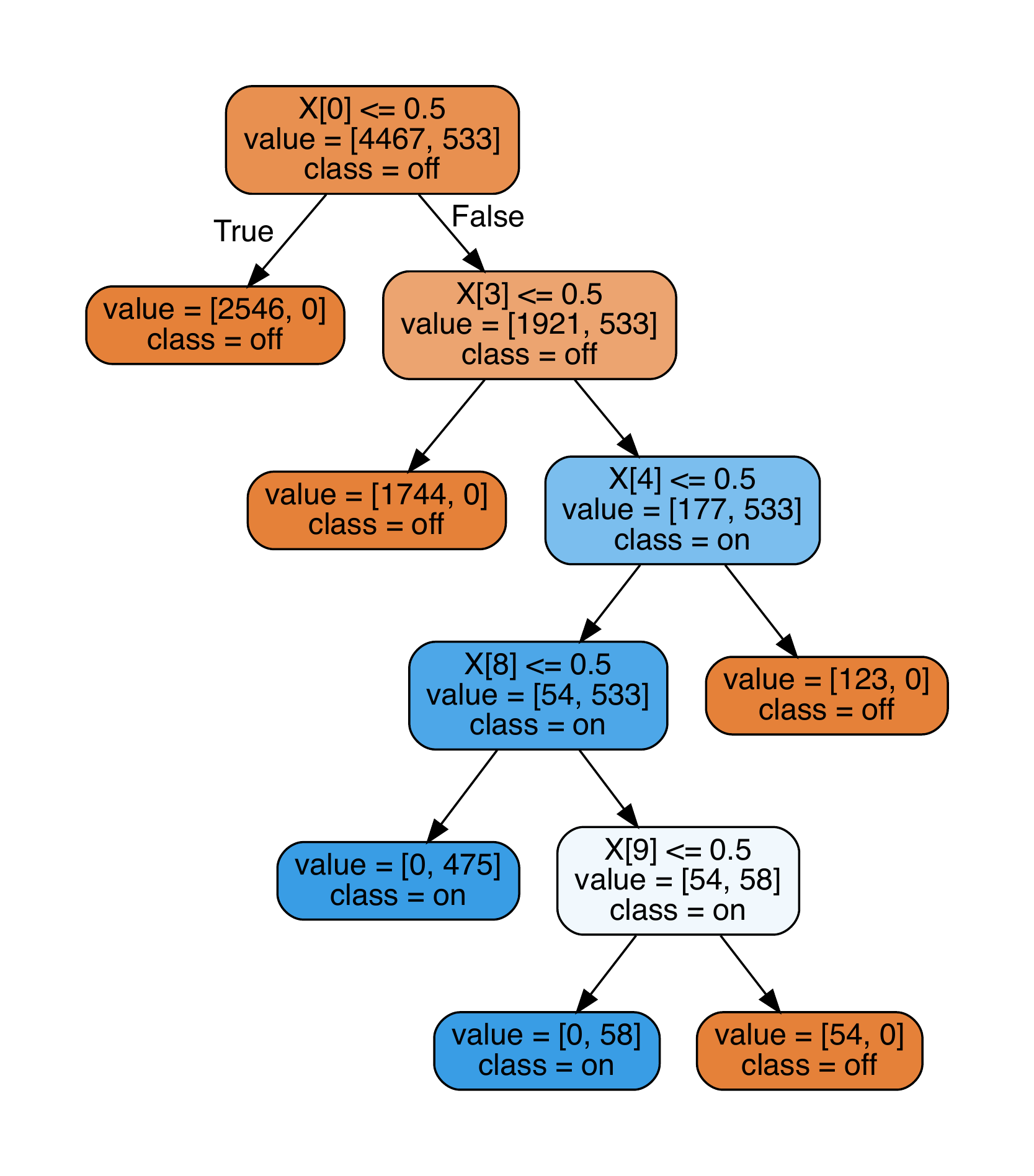}
        \caption{2/10 Runs}
        \label{}
    \end{subfigure}
    \begin{subfigure}[b]{0.24\linewidth}
        \centering
        \includegraphics[width=\linewidth]{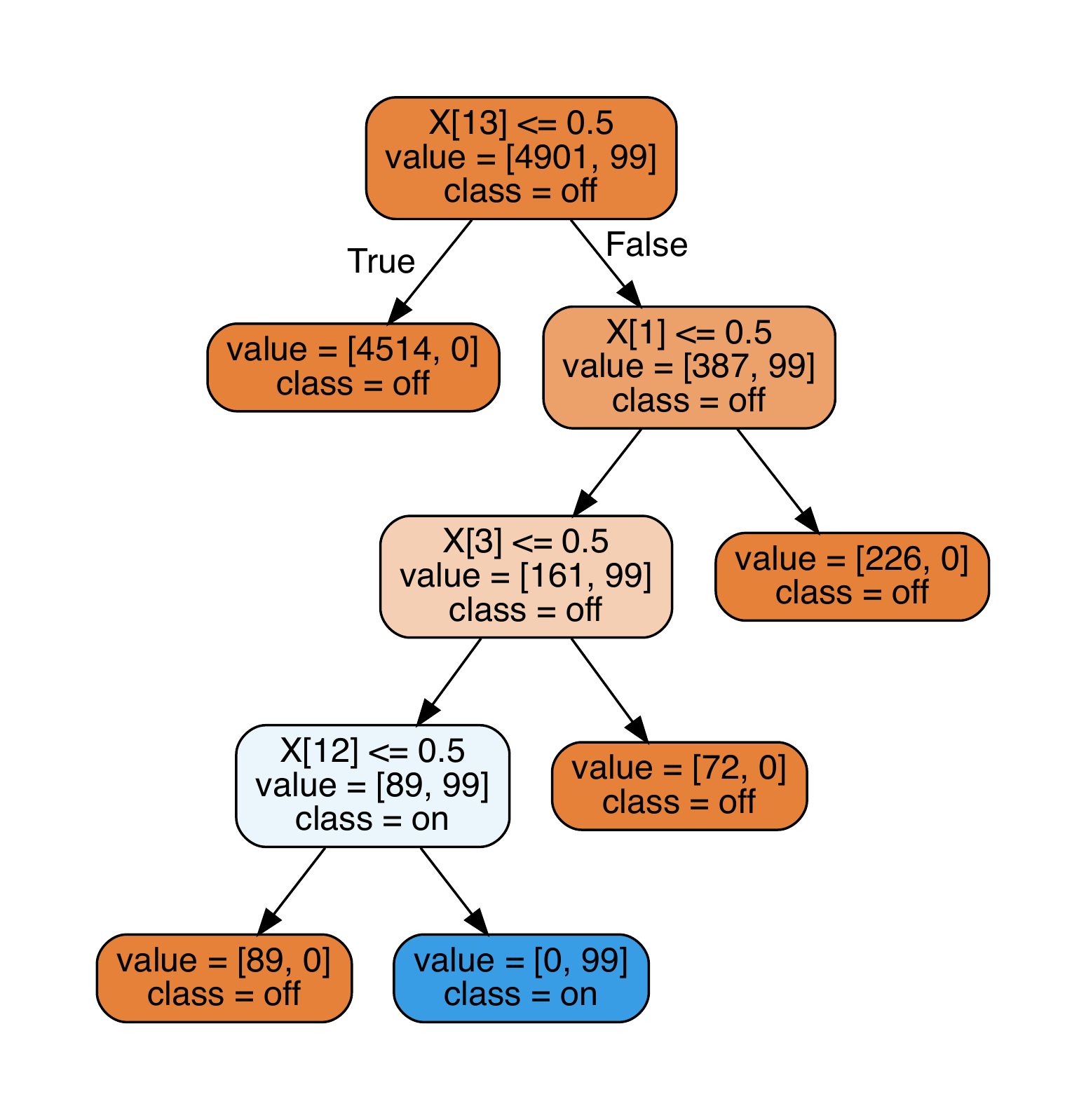}
        \caption{1/10 Runs}
        \label{}
    \end{subfigure}
    \caption{Decision trees from 10 independent runs on the signal-and-noise HMM dataset with $\lambda = 1000.0$. Seven of the ten runs resulted in a tree of the same structure. The other three trees are similar, often having additional subtrees but sharing the same splits and features.}
    \label{fig:results:stable:test}
\end{figure}

\begin{figure}[!h]
    \begin{subfigure}[b]{0.24\linewidth}
        \centering
        \includegraphics[width=\linewidth]{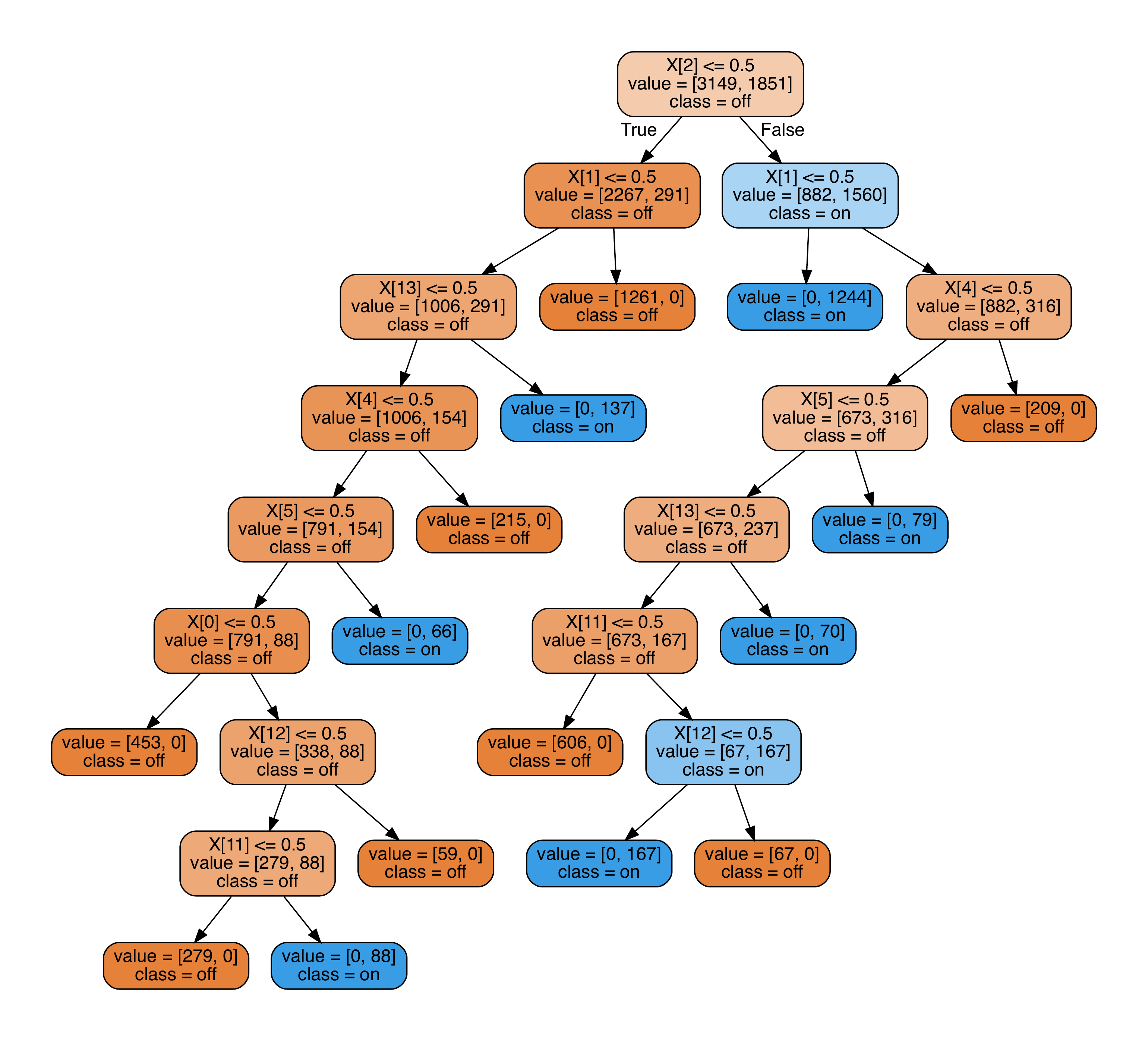}
        \caption{}
        \label{}
    \end{subfigure}
    \begin{subfigure}[b]{0.24\linewidth}
        \centering
        \includegraphics[width=\linewidth]{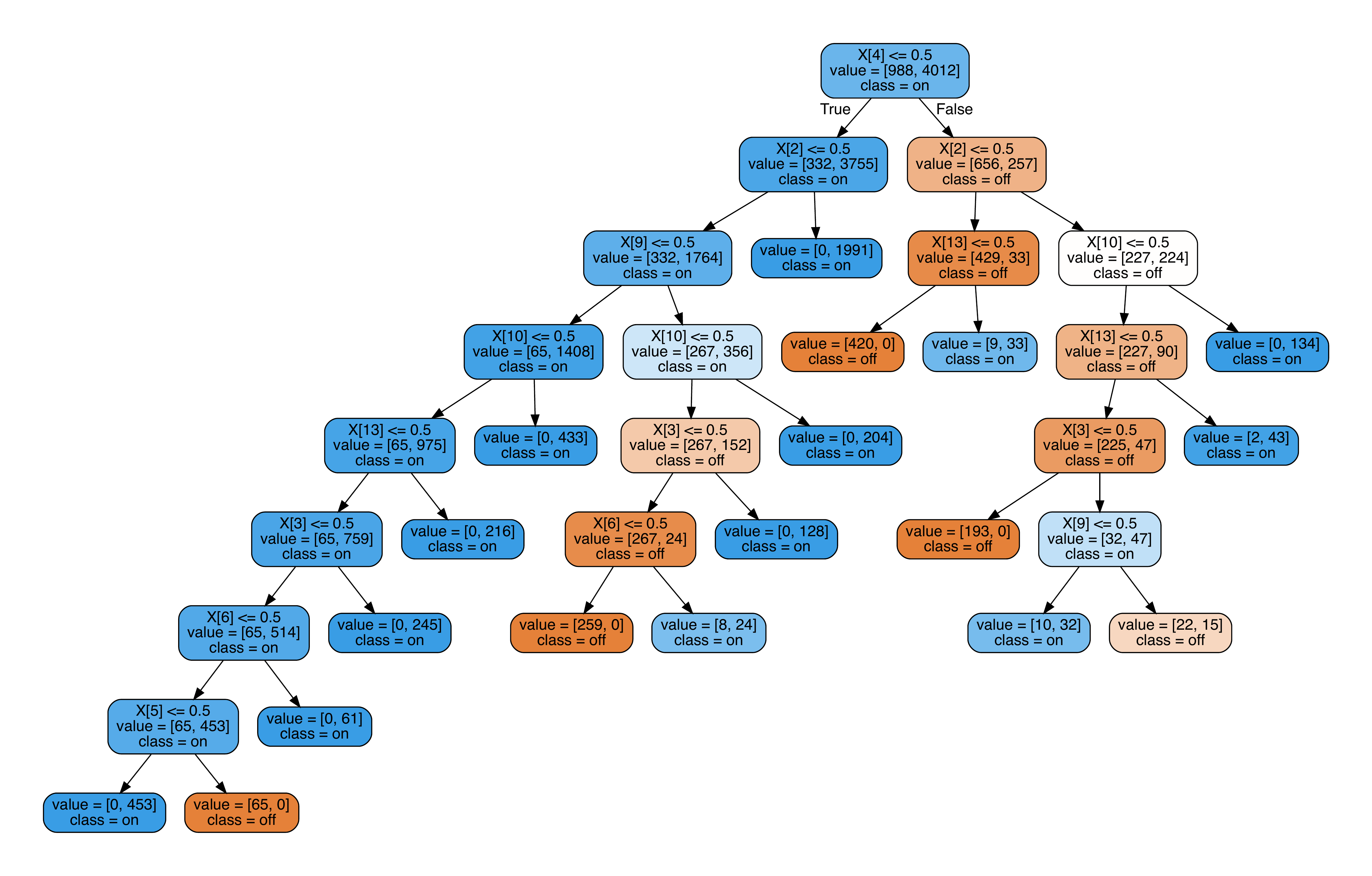}
        \caption{}
        \label{}
    \end{subfigure}
    \begin{subfigure}[b]{0.24\linewidth}
        \centering
        \includegraphics[width=\linewidth]{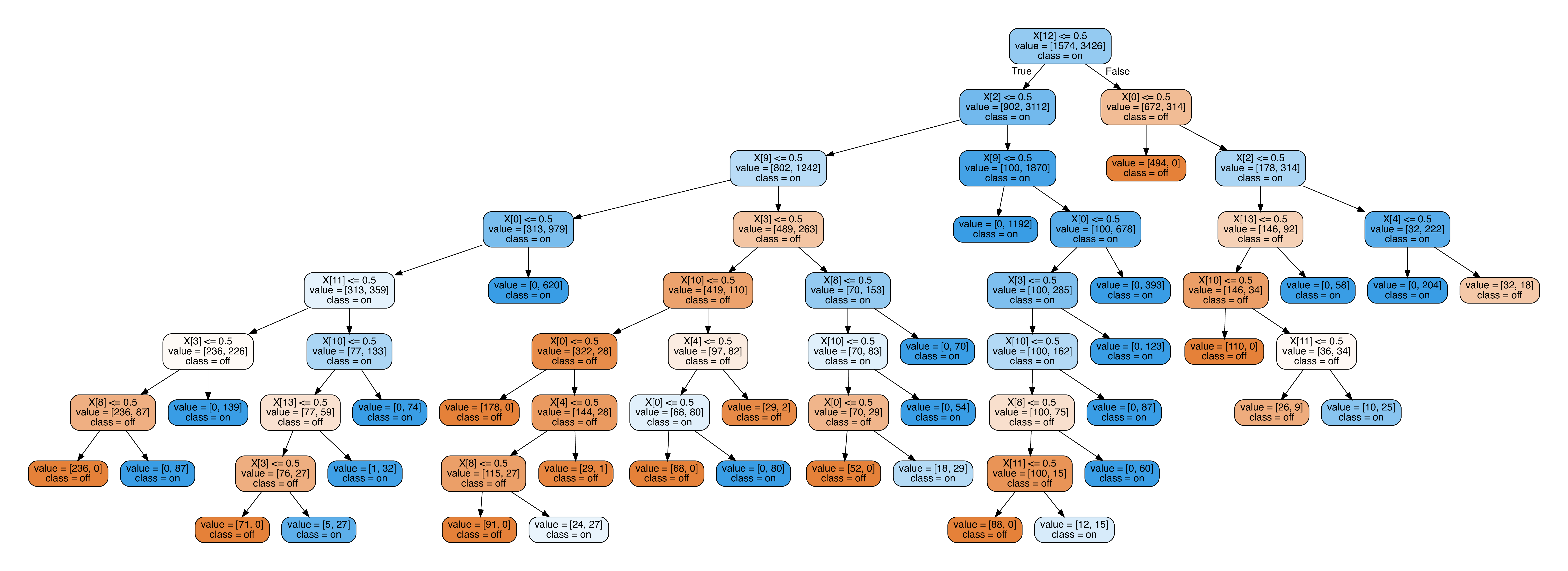}
        \caption{}
        \label{}
    \end{subfigure}
    \begin{subfigure}[b]{0.24\linewidth}
        \centering
        \includegraphics[width=\linewidth]{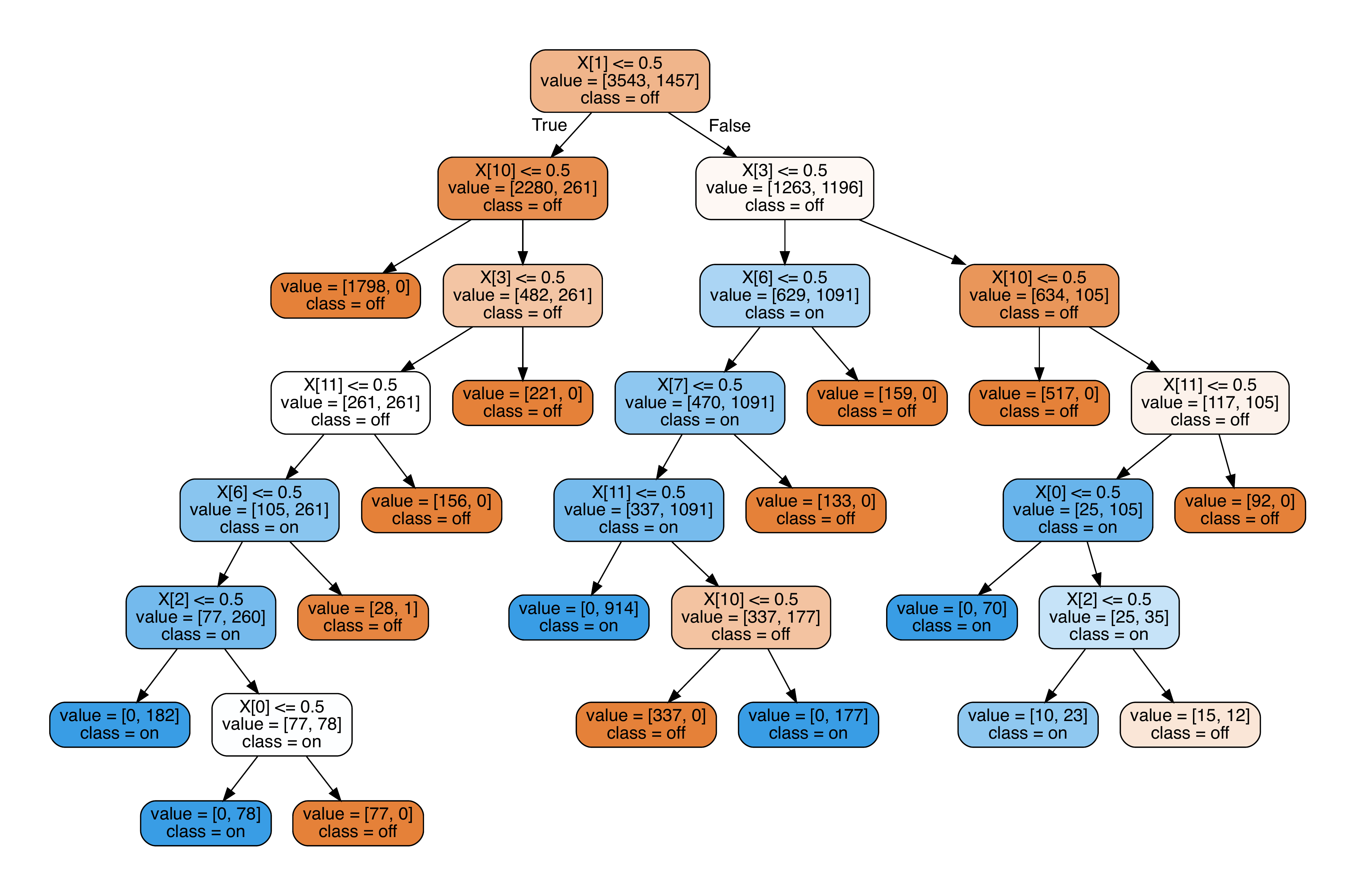}
        \caption{}
        \label{}
    \end{subfigure}
    \caption{Decision trees from 10 independent runs on the signal-and-noise HMM dataset with $\lambda = 0.01$.
    With low regularization, the variance in tree size and shape is high.}
    \label{fig:results:instable:test}
\end{figure}

In the paper, we noted that decision trees are stable over multiple run. Here, we show that
using the signal-and-noise HMM dataset, 10 independent runs with random initializations and 
$\lambda = 1000.0$ produce either the same or comparable trees. Additionally, we show that 
with weak regularization ($\lambda=0.01$), the variability of the learned decision trees is high. 
Figures \ref{fig:results:stable:test}, \ref{fig:results:instable:test} include examples of such trees on 
the signal-and-noise dataset. Similar results are found for real-world datasets.

\end{document}